\documentclass[review]{elsarticle}
\usepackage{lineno,hyperref}

\usepackage[utf8]{inputenc}
\usepackage[T1]{fontenc}
\usepackage{graphicx}
\usepackage{subcaption}
\usepackage[cmex10]{amsmath}
\usepackage{multirow}
\usepackage{array}

\usepackage{tcolorbox}
\usepackage{tikz}
\usepackage{tcolorbox}
\usepackage{xcolor}
\usepackage{color}
\usepackage[export]{adjustbox}
\graphicspath{ {./ImagesJPG2/} }

\begin{document}

\begin{frontmatter}

\title{Extracting sub-exposure images from a single capture through Fourier-based optical modulation}

\author[mymainaddress]{Shah Rez Khan}
\author[mysecondaryaddress]{Martin Feldman}
\author[mymainaddress]{Bahadir K. Gunturk\corref{mycorrespondingauthor}\fnref{myfootnote}}

\cortext[mycorrespondingauthor]{Corresponding author}
\ead{bkgunturk@medipol.edu.tr}
\fntext[myfootnote]{This work is supported by TUBITAK Grant 114C098.}

\address[mymainaddress]{Dept. of Electrical and Electronics Eng., Istanbul Medipol University, Istanbul, Turkey}
\address[mysecondaryaddress]{Div. of Electrical and Computer Eng., Louisiana State University, Baton Rouge, LA}

\begin{abstract}
Through pixel-wise optical coding of images during exposure time, it is possible to extract sub-exposure images from a single capture. Such a capability can be used for different purposes, including high-speed imaging, high-dynamic-range imaging and compressed sensing. In this paper, we demonstrate a sub-exposure image extraction method, where the exposure coding pattern is inspired from frequency division multiplexing idea of communication systems. The coding masks modulate sub-exposure images in such a way that they are placed in non-overlapping regions in Fourier domain. The sub-exposure image extraction process involves digital filtering of the captured signal with proper band-pass filters. The prototype imaging system incorporates a Liquid Crystal over Silicon (LCoS) based spatial light modulator synchronized with a camera for pixel-wise exposure coding.
\end{abstract}

\end{frontmatter}

\section{Introduction}

Coded aperture and coded exposure photography methods, which involve control of aperture shape and exposure pattern during exposure period, present new capabilities and advantages over traditional photography. In coded aperture photography, the aperture shape is designed to achieve certain goals. For example, the aperture shape can be designed to improve depth estimation accuracy as a part of depth-from-defocus technique \cite{levin2007image}, or to improve deblurring performance through adjusting the zero crossings of point spread function \cite{zhou2009good}. Coded aperture photography may involve capture of multiple images, where each image is captured with a different aperture shape, for instance, to acquire light field \cite{liang2008programmable}, or to improve depth estimation and deblurring performance \cite{zhou2011coded}. Using coded aperture, it is possible to do lensless imaging as well \cite{Sekikawa2014,Asif2015}.

In coded exposure photography, the exposure pattern is controlled during exposure period. The coding can be global, where all pixels are exposed together with a temporal pattern, or pixel-wise, where each pixel has its own exposure pattern. An example of global exposure coding is the flutter shutter technique \cite{raskar2006coded}, where the shutter is opened and closed according to a specific pattern during exposure period to enable better recovery from motion blur. The flutter shutter idea can also be used for high-speed imaging \cite{Holloway2012}. Pixel-wise exposure control presents more flexibility and wider range of applications compared to global exposure coding. An example of pixel-wise exposure control is presented in \cite{nayar2003adaptive}, where the goal is to spatially adapt the dynamic range of the captured image. Pixel-wise coded exposure imaging can also be used for focal stacking through moving the lens during the exposure period \cite{Lin2013}, and for high-dynamic-range video through per-pixel exposure offsets \cite{Portz2013}.

Pixel-wise exposure control is also used for high-speed imaging by extracting sub-exposure images from a single capture. In \cite{bub2010temporal}, pixels are exposed according to a regular non-overlapping pattern on the space-time exposure grid. Some spatial samples are skipped in one time period to take samples in another time period to improve temporal resolution. In other words, spatial resolution is traded off for temporal resolution. In \cite{gupta2010flexible}, there is also a non-overlapping space-time exposure sampling pattern; however, unlike the global spatial-temporal resolution trade-off approach of \cite{bub2010temporal}, the samples are integrated in various ways to spatially adapt spatial and temporal resolutions according to the local motion. For fast moving regions, fine temporal sampling is preferred; for slow moving regions, fine spatial sampling is preferred. Instead of a regular exposure pattern, random patterns can also be used \cite{reddy2011p2c2,hitomi2011video,liu2014efficient}. In \cite{reddy2011p2c2}, pixel-wise random sub-exposure masks are used during exposure period. The reconstruction algorithm utilizes the spatial correlation of natural images and the brightness constancy assumption in temporal domain to achieve high-speed imaging. In \cite{hitomi2011video,liu2014efficient}, the reconstruction algorithm is based on sparse dictionary learning. While learning-based approaches may yield outstanding performance, one drawback is that the dictionaries need to be re-trained each time a related parameter, such as target frame rate, is changed.

Alternative to arbitrary pixel-wise exposure patterns, it is also proposed to have row-wise control \cite{gu2010coded} and translated exposure mask \cite{llull2013coded,Koller2015}. Row-wise exposure pattern can be designed to achieve high-speed imaging, high-dynamic-range imaging, and adaptive auto exposure \cite{gu2010coded}. In \cite{llull2013coded}, binary transmission masks are translated during exposure period for exposure coding; sub-exposure images are then reconstructed using an alternating projections algorithm. The same coding scheme is also used in \cite{Koller2015}, but a different image reconstruction approach is taken.

\begin{figure*}
\centering
\includegraphics[width=2.8cm]{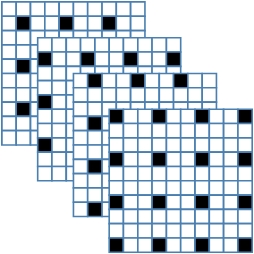}
\includegraphics[width=2.8cm]{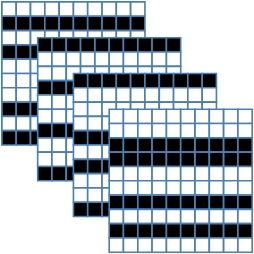}
\includegraphics[width=2.8cm]{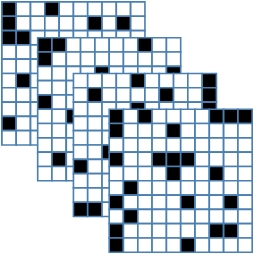}
\includegraphics[width=2.8cm]{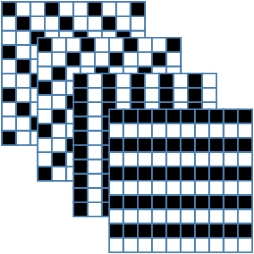}\\
(a) \hspace{2.5cm} (b) \hspace{2.5cm} (c) \hspace{2.5cm} (d)\\
\caption{Illustration of different exposure masks. (a) Non-overlapping uniform grid exposure \cite{gupta2010flexible}, (b) Coded rolling shutter \cite{gu2010coded}, (c) Pixel-wise random exposure \cite{hitomi2011video}, (d) Frequency division multiplexed imaging exposure \cite{gunturk2013frequency}.}
\label{fig:ExposureMaskTypes}
\end{figure*}

Pixel-wise exposure control can be implemented using regular image sensors with the help of additional optical elements. In \cite{nayar2003adaptive}, an LCD panel is placed in front of a camera to spatially control light attenuation. With such a system, pixel-wise exposure control is difficult since the LCD attenuator is optically defocused. In \cite{gao2007active}, an LCD panel is placed on the intermediate image plane, which allows better pixel-by-pixel exposure control. One disadvantage of using transmissive LCD panels is the low fill factor due to drive circuit elements between the liquid crystal elements. In \cite{nayar2006programmable}, a DMD reflector is placed on the intermediate image plane. DMD reflectors have high fill factor and high contrast ratio, thus they can produce sharper and higher dynamic range images compared to LCD panels. One drawback of the DMD based design is that the micromirrors on a DMD device reflect light at two small angles, thus the DMD plane and the sensor plane must be properly inclined, resulting in ``keystone'' perspective distortion. That is, a square DMD pixel is imaged as a trapezoid shape on the sensor plane. As a result, pixel-to-pixel mapping between the DMD and the sensor is difficult. In \cite{mannami2007high}, a reflective LCoS spatial light modulator (SLM) is used on the intermediate image plane. Because the drive circuits on an LCoS device is are on the back, high fill factor is possible as opposed to the transmissive LCD devices. Compared to a DMD, one-to-one pixel correspondence is easier with an LCoS SLM; however, the light efficiency is not as good as the DMD approach. In \cite{llull2013coded}, a lithographically patterned chrome-on-quartz binary transmission mask is placed on the intermediate image plane, and moved during exposure period with a piezoelectric stage for optical coding. This approach is limited in terms of the exposure pattern that can be applied.

In this paper, we demonstrate a sub-exposure image extraction idea. The idea, which is called frequency division multiplexed imaging (FDMI), was presented by Gunturk and Feldman as a conference paper \cite{gunturk2013frequency}. While the FDMI idea was demonstrated by merging two separate images with a patterned glass based and an LCD panel based modulation in \cite{gunturk2013frequency}, it was not demonstrated for sub-exposure image extraction. Here, we apply the FDMI idea to extract sub-exposure images using an optical setup incorporating an LCoS SLM synchronized with a camera for exposure coding.

In Section 2, we present the problem of extracting sub-exposure images through space-time exposure coding, and review the FDMI approach. In Section 3, we present the optical setup used in the experiments. In Section 4, we provide experimental results with several coded image captures. In Section 5, we conclude the paper with some future research directions.

\section{Extracting Sub-Exposure Images from a Single Capture}

There are various exposure coding schemes designed for extracting sub-exposure images from a single capture. First, we would like to present a formulation of the coding process, and then review the FDMI idea.

\subsection{Space-Time Exposure Coding}

Space-time exposure coding of an image can be formulated using a spatio-temporal video signal $I(x,y,t)$, where $(x,y)$ are the spatial coordinates and $t$ is the time coordinate. This signal is modulated during an exposure period $T$ with a mask $m(x,y,t)$ to generate an image:
\begin{equation}
I (x,y) = \int_{0}^{T} m(x,y,t) I(x,y,t) dt.
\label{eqn:imaging}
\end{equation}
The mask $m(x,y,t)$ can be divided in time into a set of constant sub-exposure masks: $m_1(x,y)$ for $t \in (t_0,t_1)$, $m_2(x,y)$ for $t \in (t_1,t_2)$, ..., $m_N(x,y)$ for $t \in (t_{N-1},t_{N})$, where $N$ is the number of masks, and $t_0$ and $t_N$ are the start and end times of the exposure period. Incorporating the sub-exposure masks $m_i(x,y)$ into equation (\ref{eqn:imaging}), the captured image becomes
\begin{equation}
I (x,y) = \sum_{i=1}^{N} m_i(x,y) \int_{t_{i-1}}^{t_{i}} I(x,y,t) dt = \sum_{i=1}^{N} m_i(x,y) I_i (x,y),
\end{equation}
where we define the sub-exposure image $I_i (x,y) = \int_{t_{i-1}}^{t_{i}} I(x,y,t) dt$. The above equation states that the sub-exposure images $I_i(x,y)$ are modulated with the masks $m_i(x,y)$ and added up to form the recorded image $I (x,y)$. The goal of sub-exposure image extraction is to estimate the images $I_i(x,y)$ given the masks and the recorded image.

As we have already mentioned in Section 1, the reconstruction process might be based on different techniques, varying from simple interpolation \cite{bub2010temporal} to dictionary learning \cite{liu2014efficient}. The masks $m_i(x,y)$ can be chosen in different ways as well. Some of the masks, including non-overlapping uniform grid \cite{bub2010temporal}, coded rolling shutter \cite{gu2010coded}, pixel-wise random exposure \cite{hitomi2011video}, and frequency division multiplexed imaging exposure \cite{gunturk2013frequency}, are illustrated in Figure \ref{fig:ExposureMaskTypes}.

\begin{figure*}
\centering
\includegraphics[width=12cm]{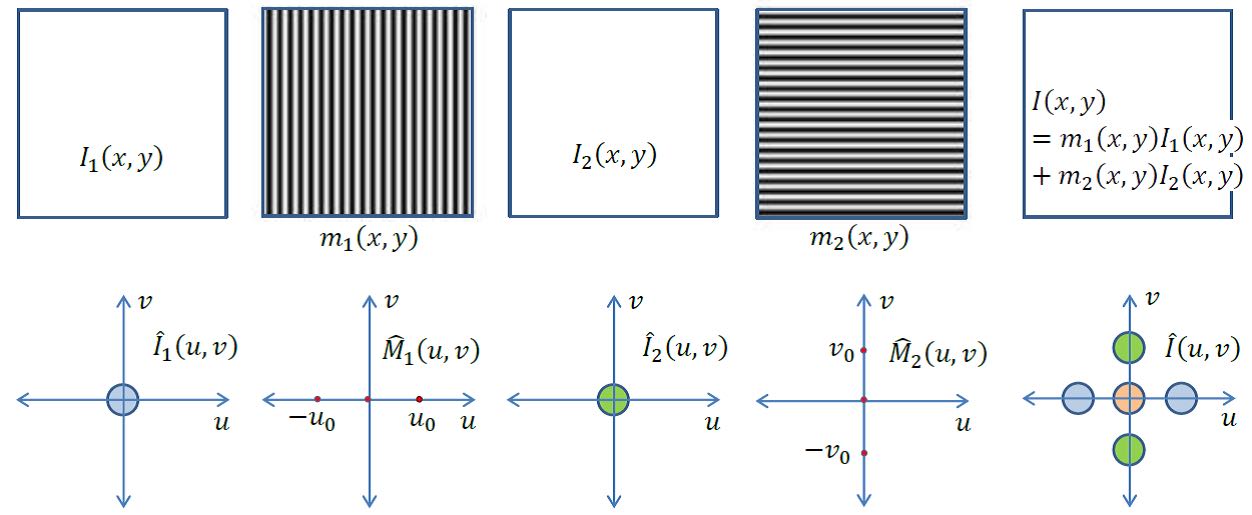}
\caption{Illustration of the FDMI idea with two images \cite{gunturk2013frequency}.}
\label{fig: Gunturk Illustration}
\end{figure*}

\subsection{Frequency Division Multiplexed Imaging}

The frequency division multiplexed imaging (FDMI) idea \cite{gunturk2013frequency} is inspired from frequency division multiplexing method in communication systems, where the communication channel is divided into non-overlapping sidebands, each of which carry independent signals that are properly modulated. In case of FDMI, sub-exposure images are modulated such that they are placed in different regions in Fourier domain. By ensuring that the Fourier components of different sub-exposure images do not overlap, each sub-exposure image can be extracted from the captured signal through band-pass filtering. The FDMI idea is illustrated for two images in Figure \ref{fig: Gunturk Illustration}. Two band-limited sub-exposure images $I_1(x, y)$ and $I_2(x, y)$ are modulated with horizontal and vertical sinusoidal masks $m_1(x, y)$ and $m_2(x, y)$ during exposure period. The masks can be chosen as raised cosines: $m_1(x, y) = a + b cos(2 \pi u_0x)$ and $m_2(x, y) = a + b cos(2 \pi v_0y)$, where $a$ and $b$ are positive constants with the condition $a \geq b$ so that the masks are non-negative, that is, optically realizable, and $u_0$ and $v_0$ are the spatial frequencies of the masks. The imaging system captures sum of the modulated images: $I(x, y) = m_1(x, y)I_1(x, y) + m_2(x, y)I_2(x, y).$

The imaging process from the Fourier domain perspective is also illustrated in Figure \ref{fig: Gunturk Illustration}. $\hat{I}_1(u, v)$ and $\hat{I}_2(u, v)$ are the Fourier transforms of the sub-exposure images $I_1(x, y)$ and $I_2(x, y)$, which are assumed to be band-limited. In Figure \ref{fig: Gunturk Illustration}, the Fourier transforms $\hat{I}_1(u, v)$ and $\hat{I}_2(u, v)$ are shown as circular regions. The Fourier transforms of the masks are impulses: $\hat{M}_1(u, v) = a \delta (u, v) + (b/2) (\delta (u \text{ - } u_0, v) + \delta (u + u_0, v))$ and $\hat{M}_2(u, v) = a \delta (u, v) + \\ (b/2) (\delta (u , v \text{ - } v_0) + \delta (u, v + v_0))$. As a result, the Fourier transform $\hat{I} (u, v)$ of the recorded image $I(x, y)$ includes $\hat{I}_1(u, v)$ and $\hat{I}_2(u, v)$ in its sidebands, and $\hat{I}_1(u, v) + \hat{I}_2(u, v)$ in the baseband. From the sidebands, the individual sub-exposure images can be recovered with proper band-pass filtering. From the baseband, the full-exposure image $I_1(x, y) + I_2(x, y)$ can be recovered with a low-pass filter.

It is possible to use other periodic signals instead of a cosine wave. For example, when a square wave is used, the baseband is modulated to all harmonics of the main frequency, where the weights of the harmonics decrease with a sinc function. Again, by applying band-pass filters on the first harmonics, sub-exposure images can be recovered.

\begin{figure*}
\centering
\includegraphics[width =10cm]{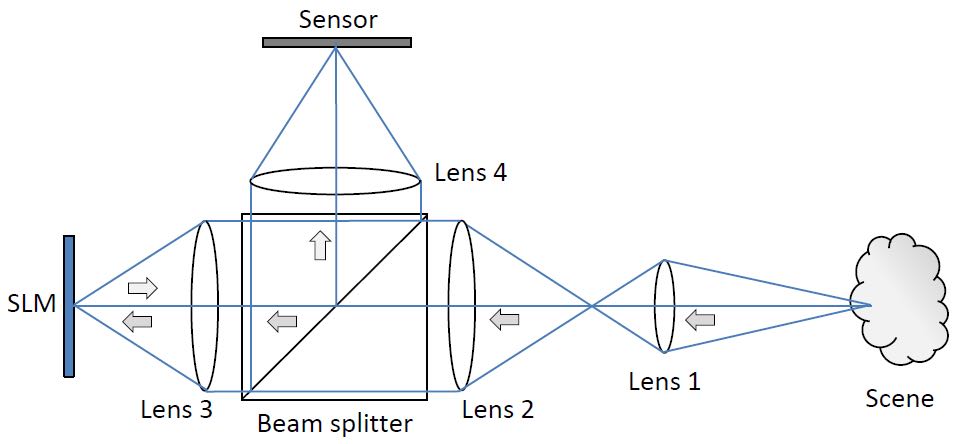}\\
(a)\\
\vspace{0.08cm}
\includegraphics[width=12cm]{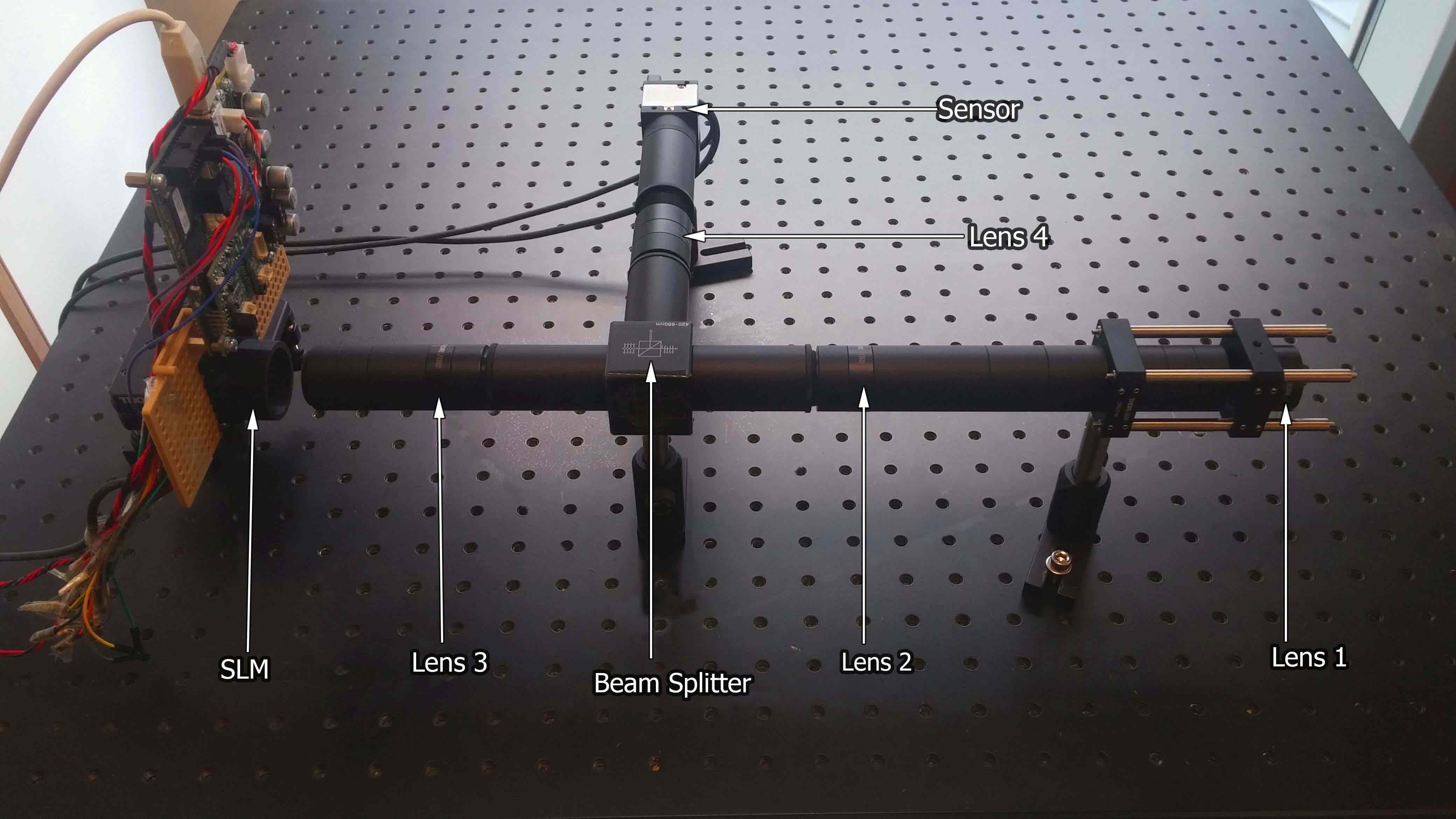}\\
(b)\\
\caption{Pixel-wise exposure control camera prototype. (a) Graphical illustration, (b) Actual prototype.}
\label{fig:prototype}
\end{figure*}

\section{Prototype Design}

The prototype system is based on the pixel-wise exposure control design that is adopted in several papers \cite{mannami2007high,hitomi2011video,liu2014efficient}. As shown in Figure \ref{fig:prototype}, the system consists of an objective lens, three relay lenses, one polarizing beam splitter, an LCoS SLM, and a camera. Relay lenses are 100mm aspherical doublets; the objective lens is a 75mm aspherical doublet; the SLM is a Forth Dimension Display SXGA-3DM with 1280x1024 pixel resolution and a pixel pitch of 13.62$\mu m$; and the camera is a Thorlabs DCU-224M monochromatic camera with 1280x1024 pixel resolution and a pixel pitch of 4.65$\mu m$. The objective lens forms an image on an intermediate image plane, which is relayed onto the SLM. The SLM controls the exposure pattern of each pixel through changing the polarization states. The image reflected from the SLM is recorded by the camera. The camera and the SLM are synchronized using the trigger signal from the SLM.

\section{Experimental Results}

% Experiment 1: Simulation
We conducted several experiments to demonstrate sub-exposure image extraction with FDMI. As the first experiment, we performed a simulation to discuss about the mask design and limitations of the FDMI approach. An image is rotated and translated to create twelve images. Each image is low-pass filtered to obtain band-limited signals. The images are then modulated, each with a different mask, to place them in different regions of the Fourier domain. The masks are designed such that the sidebands at the fundamental frequencies and the harmonics do not overlap in the Fourier domain. We first decide on where to place the sideband in the Fourier domain; form the corresponding 2D raised cosine signal (mask) as discussed in Section 2; and discretize (sample) the signal to match size of the input images. Since sampling is involved during the mask creation, we should avoid aliasing. The highest frequency sinusoidal mask (in, for instance, horizontal direction) becomes a square wave (binary pattern) with a period of two pixels.  

\begin{figure}
\centering\includegraphics[width=0.10\textwidth]{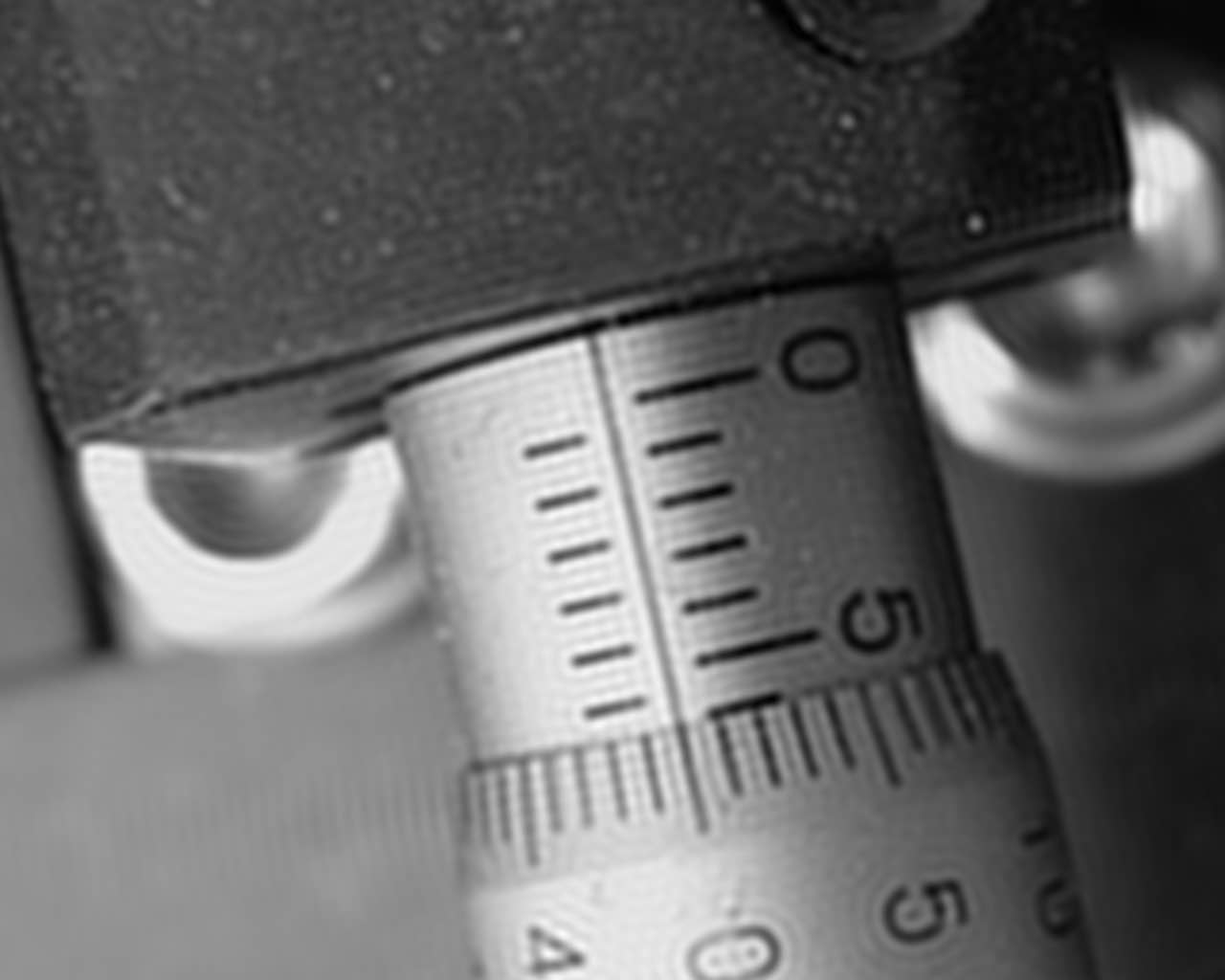}
\centering\includegraphics[width=0.10\textwidth]{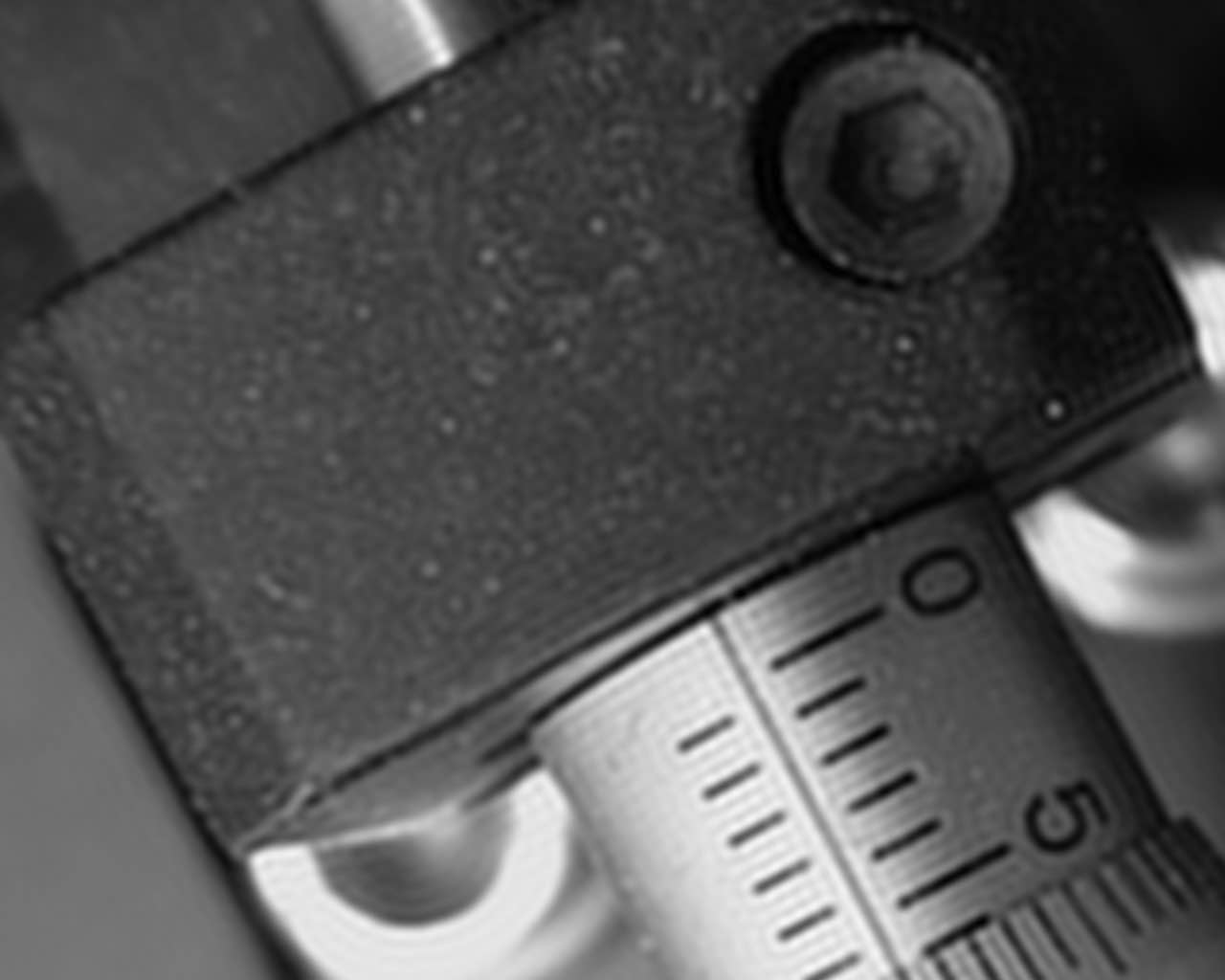}
\centering\includegraphics[width=0.10\textwidth]{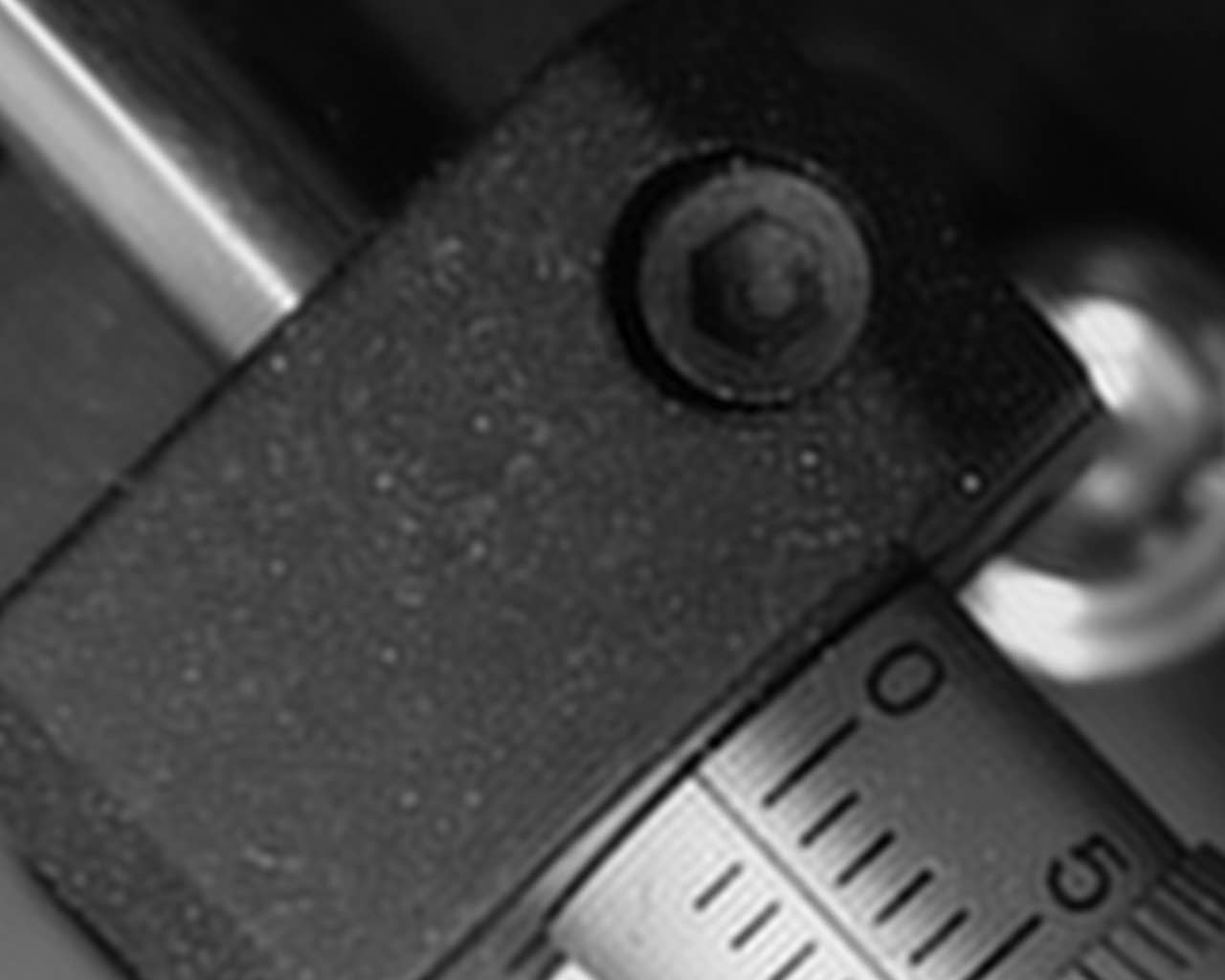}
\centering\includegraphics[width=0.10\textwidth]{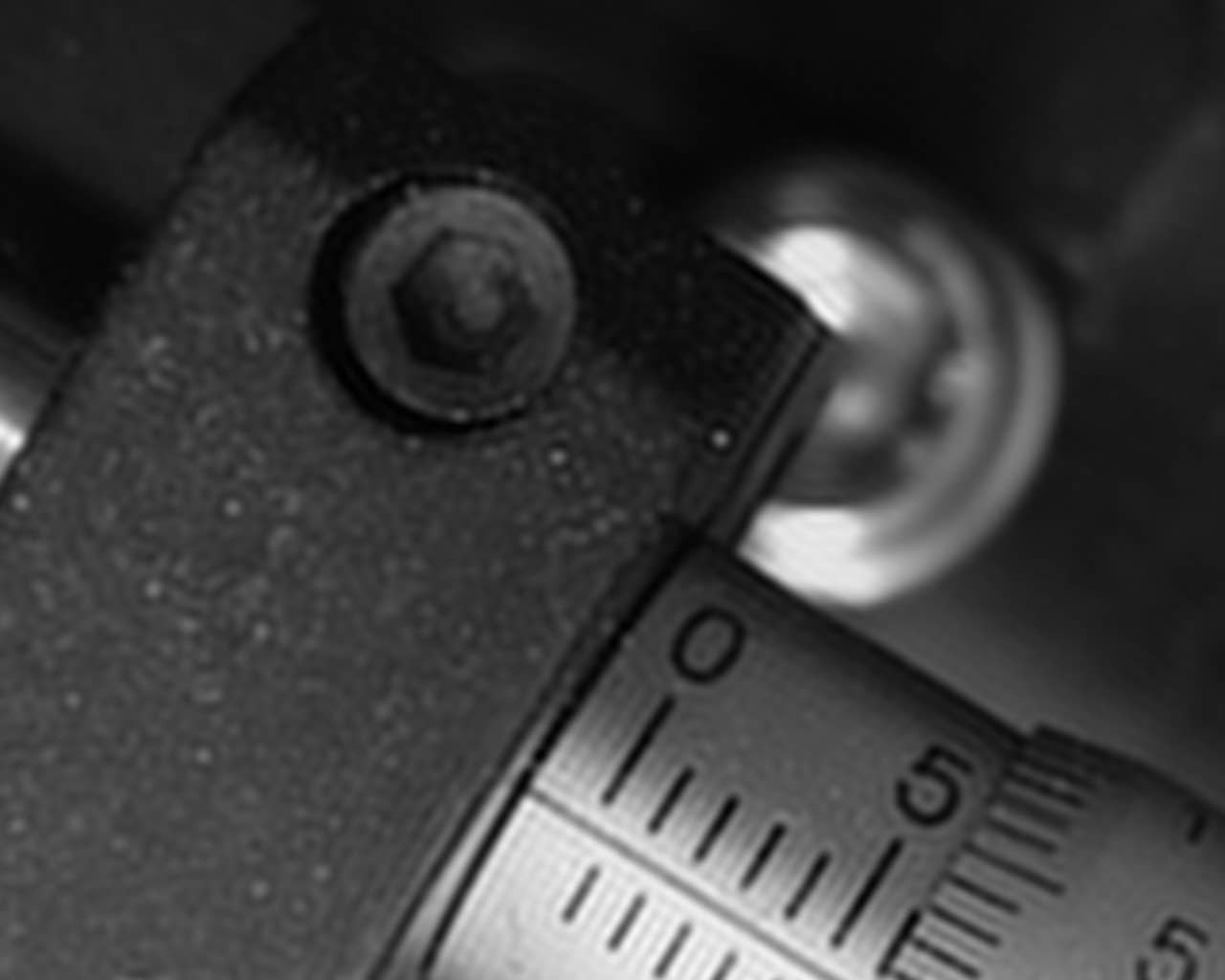}
\hspace{0.05\textwidth}
\centering\includegraphics[trim= 987 990 280 24, clip=true, width=0.10\textwidth]{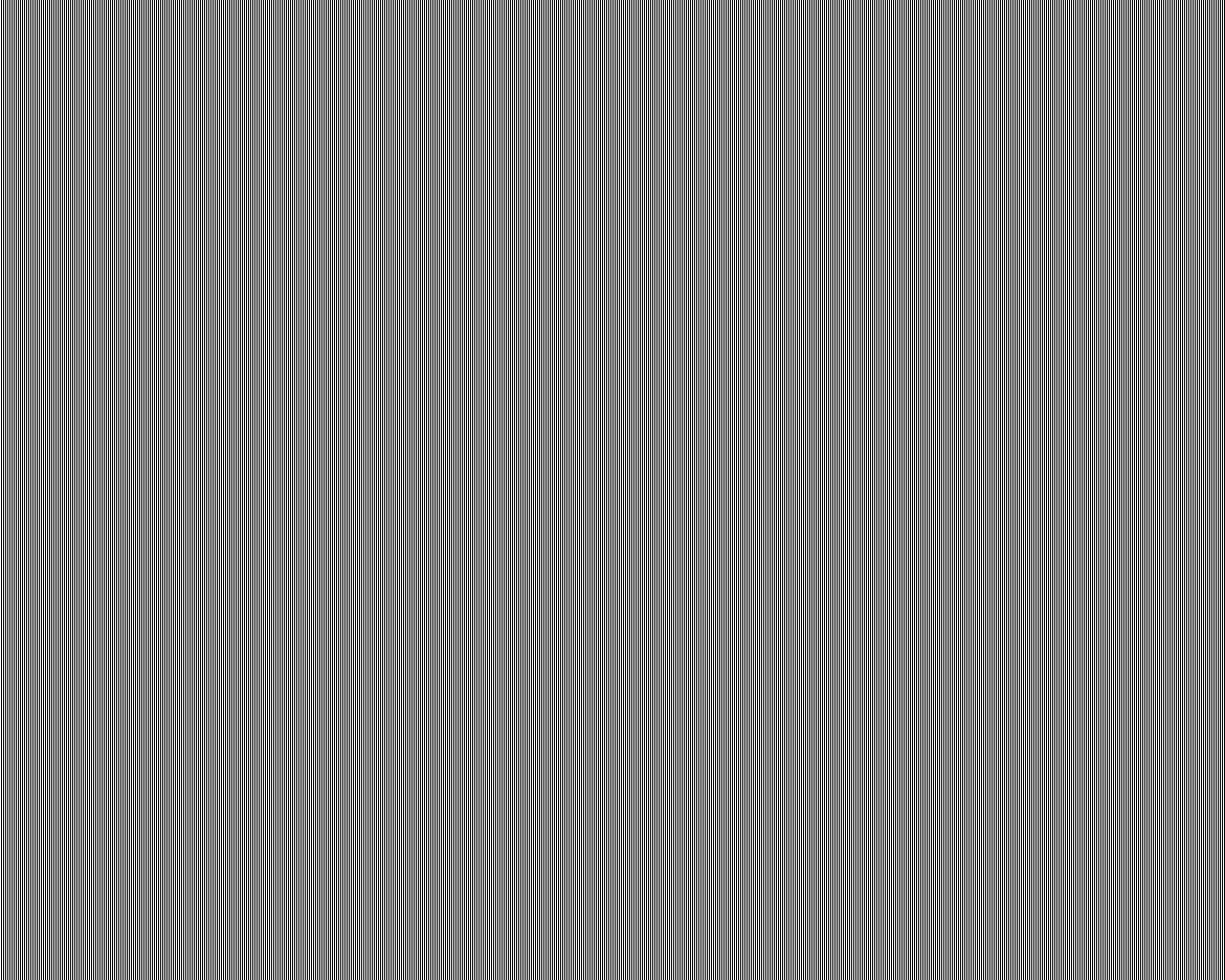}
\centering\includegraphics[trim= 987 990 280 24, clip=true, width=0.10\textwidth]{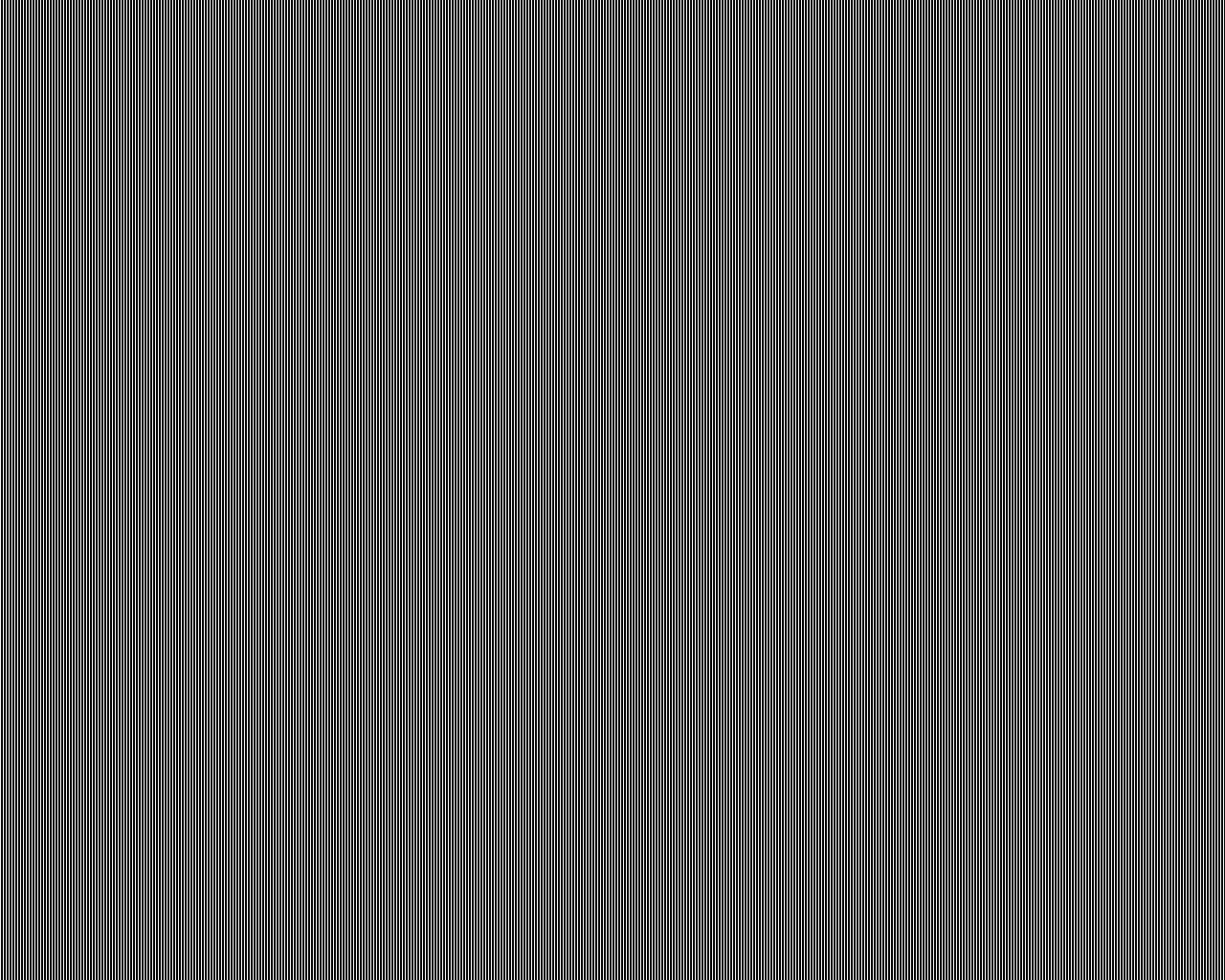}
\centering\includegraphics[trim= 987 990 280 24, clip=true, width=0.10\textwidth]{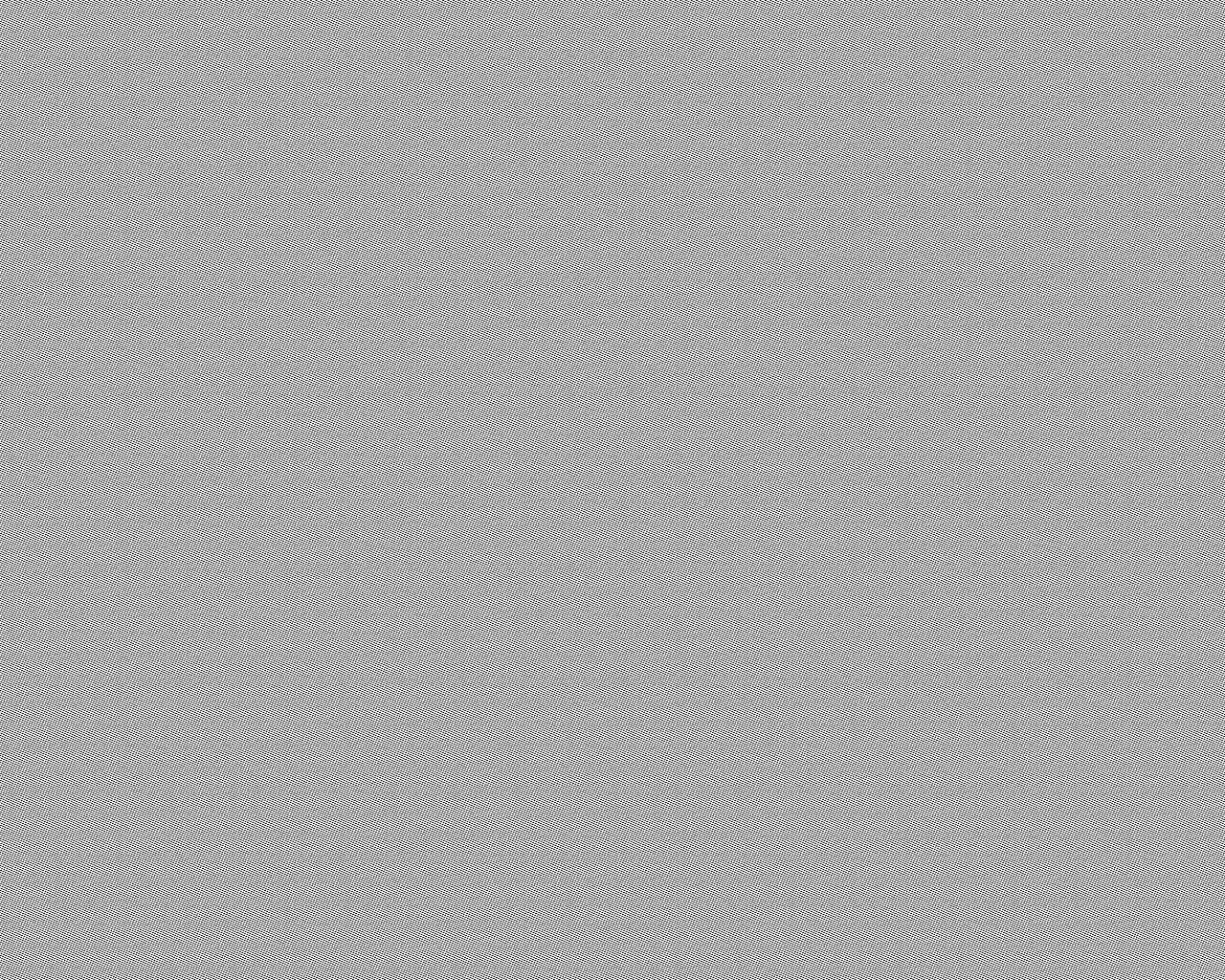}
\centering\includegraphics[trim= 987 990 280 24, clip=true, width=0.10\textwidth]{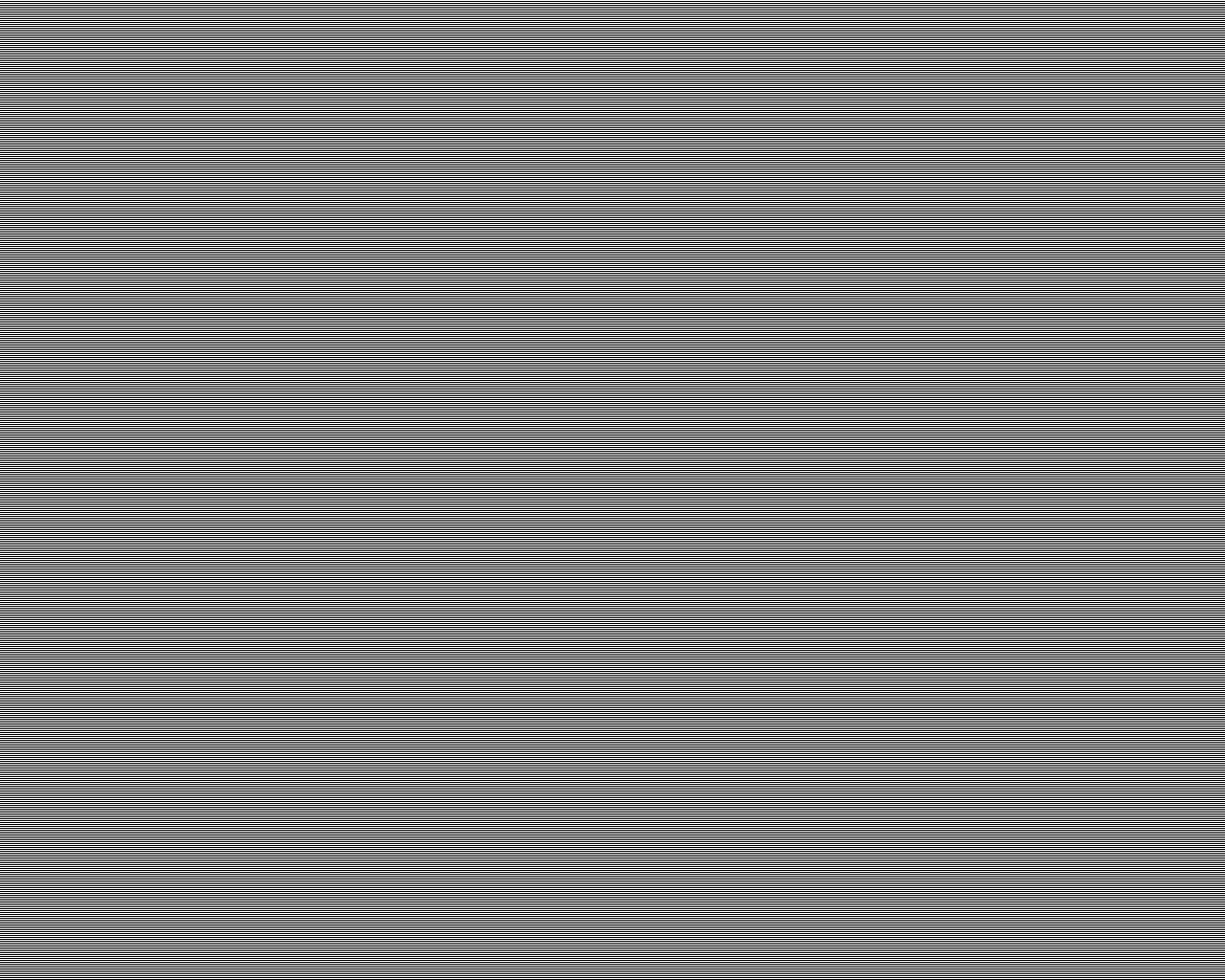}\\
\vspace{0.08cm}

\centering\includegraphics[width=0.10\textwidth]{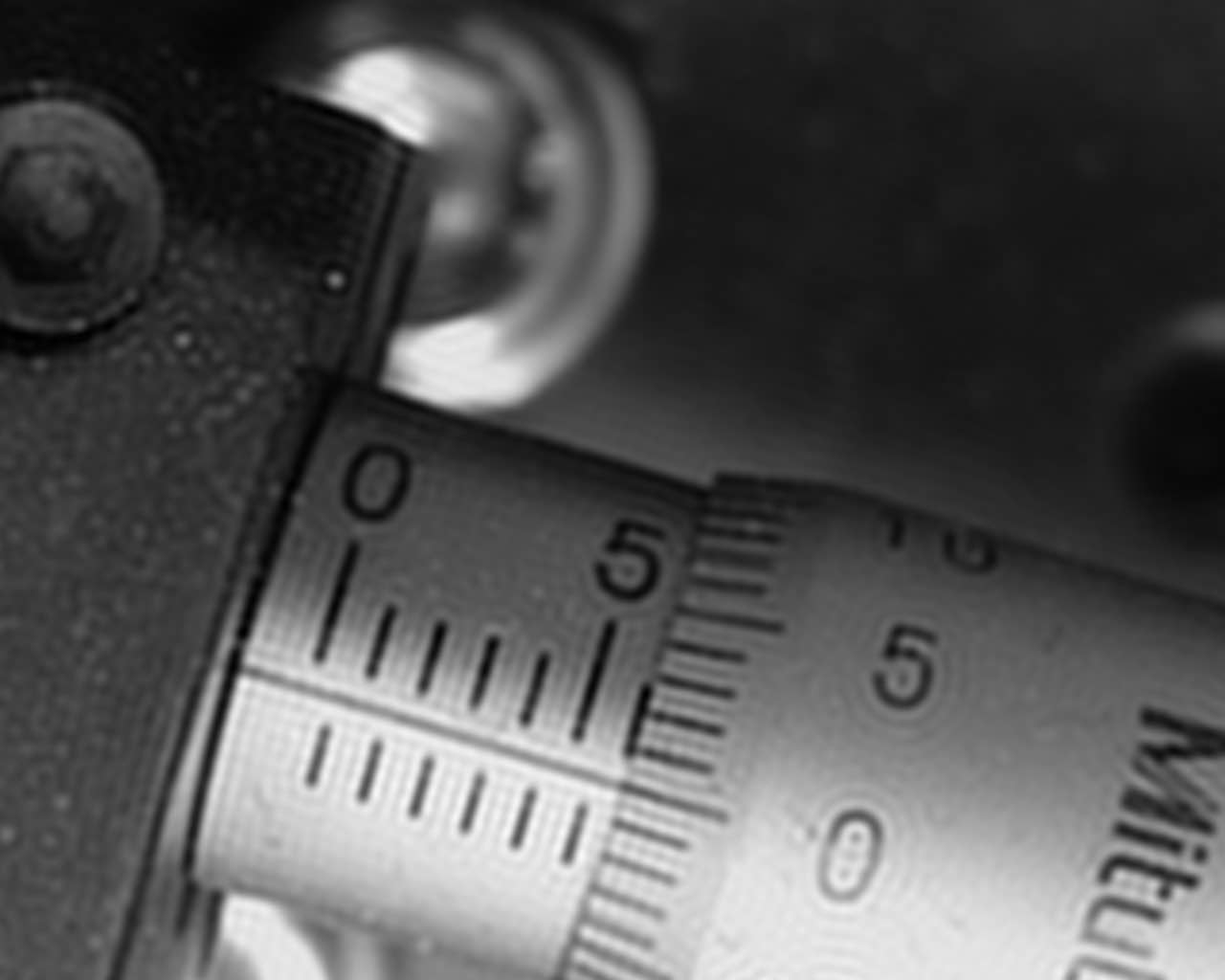}
\centering\includegraphics[width=0.10\textwidth]{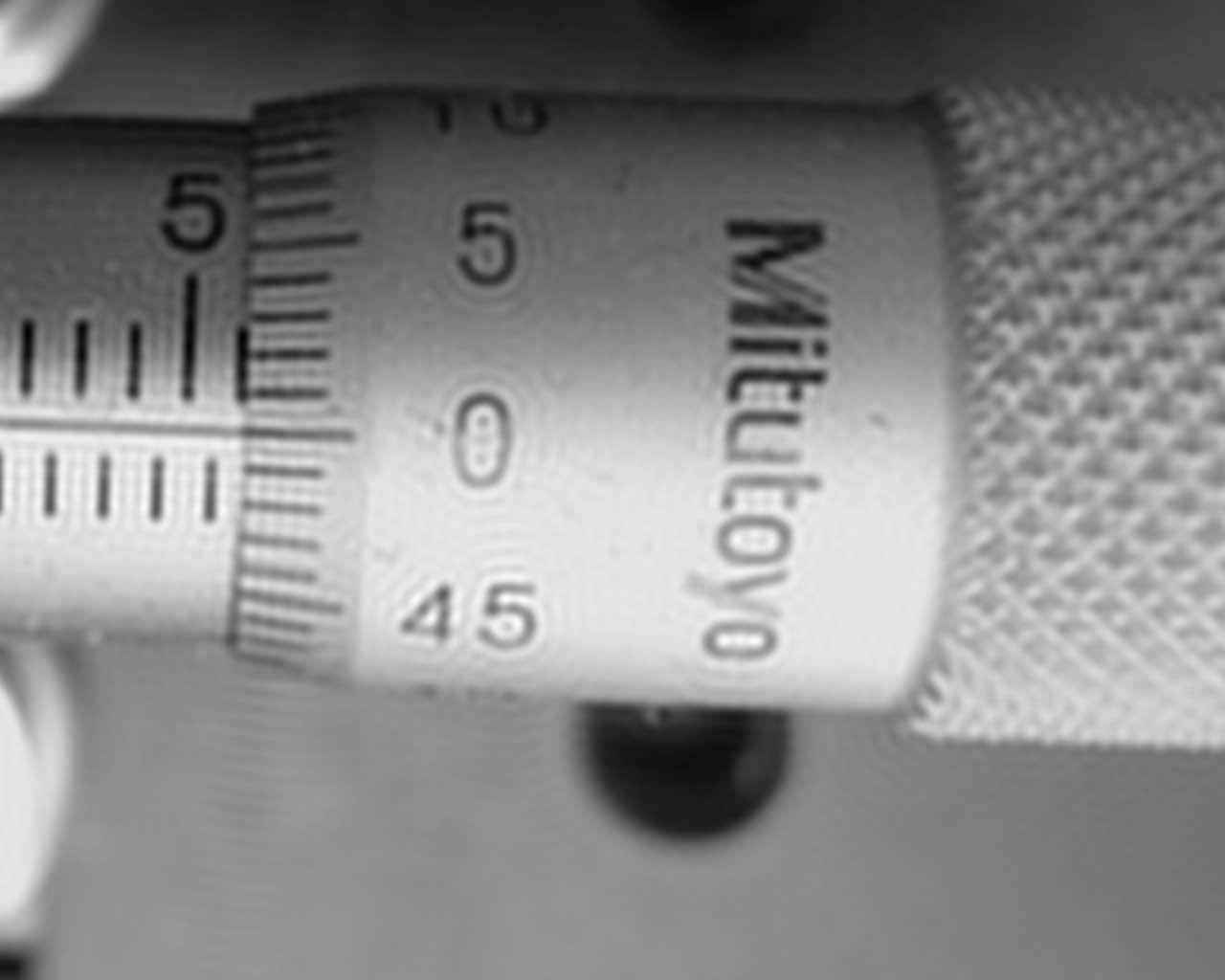}
\centering\includegraphics[width=0.10\textwidth]{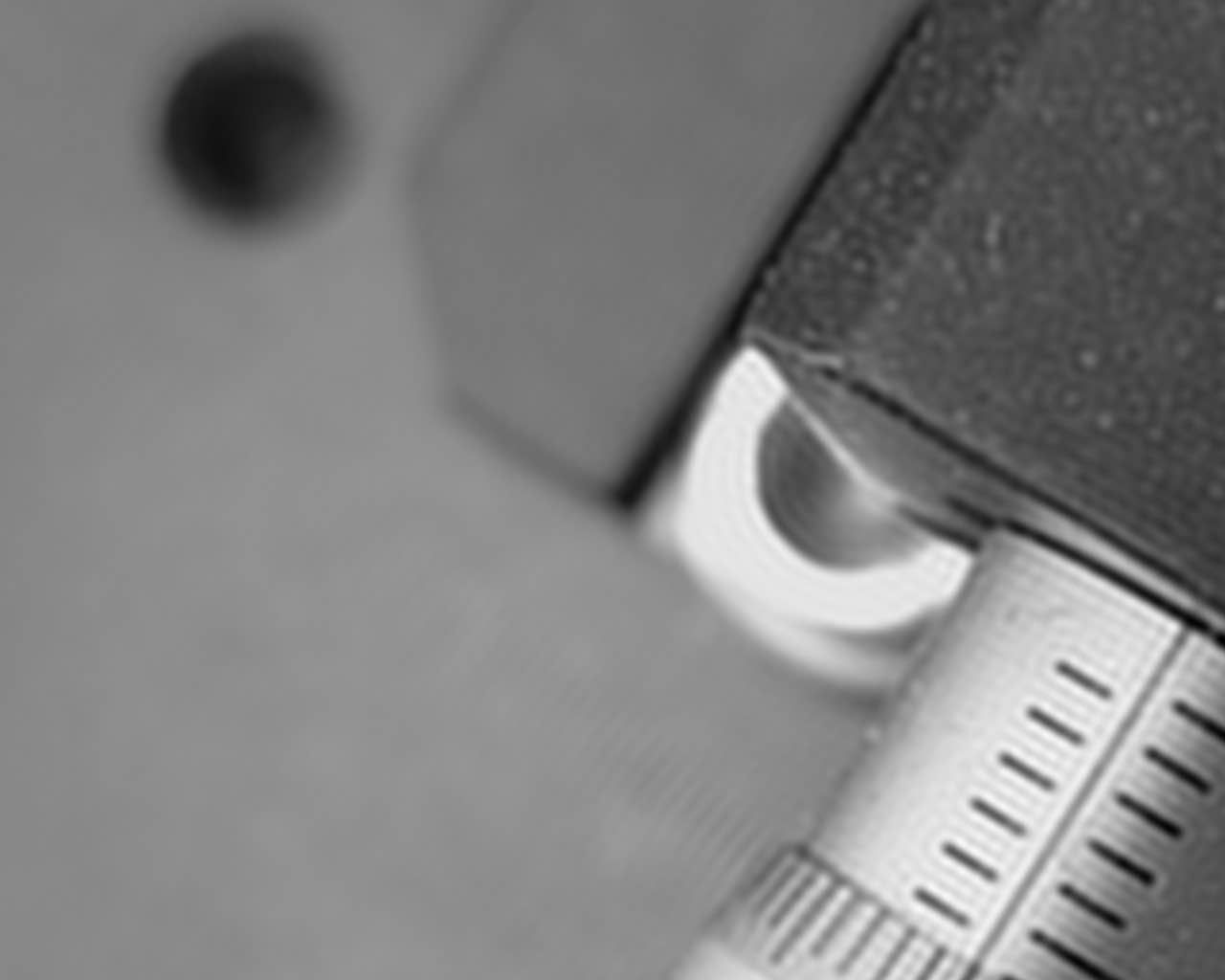}
\centering\includegraphics[width=0.10\textwidth]{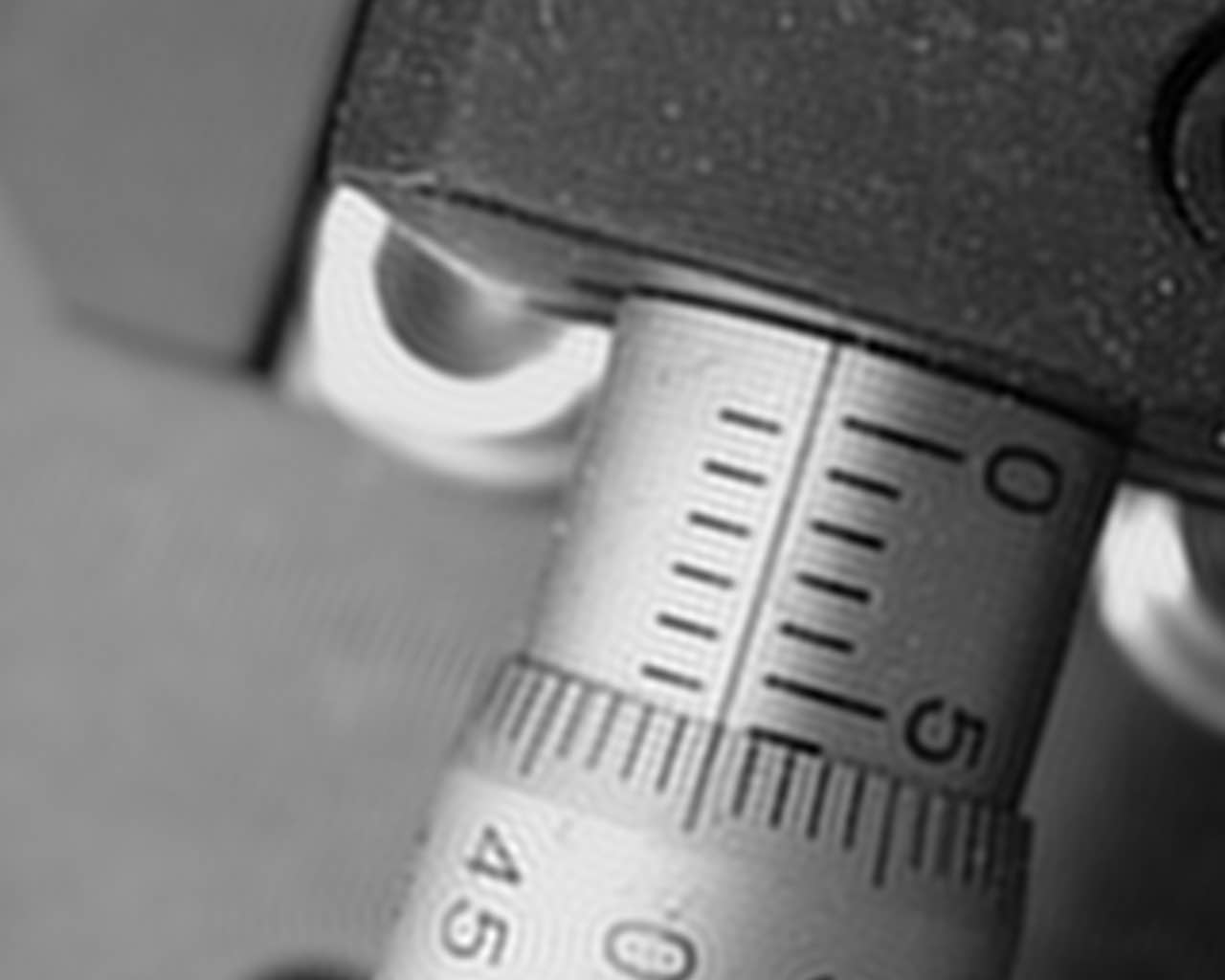}
\hspace{0.05\textwidth}
\centering\includegraphics[trim= 987 990 280 24, clip=true, width=0.10\textwidth]{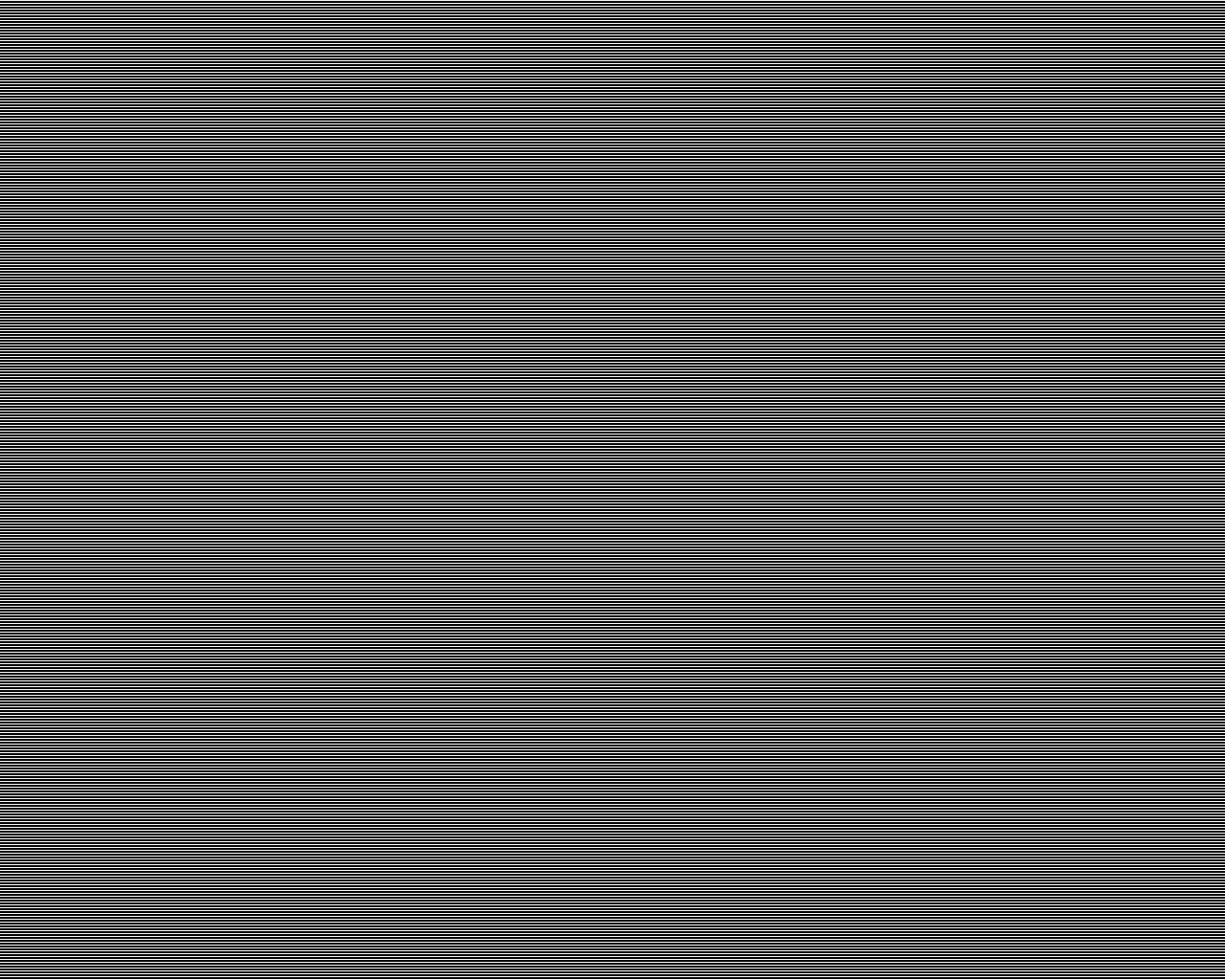}
\centering\includegraphics[trim= 987 990 280 24, clip=true, width=0.10\textwidth]{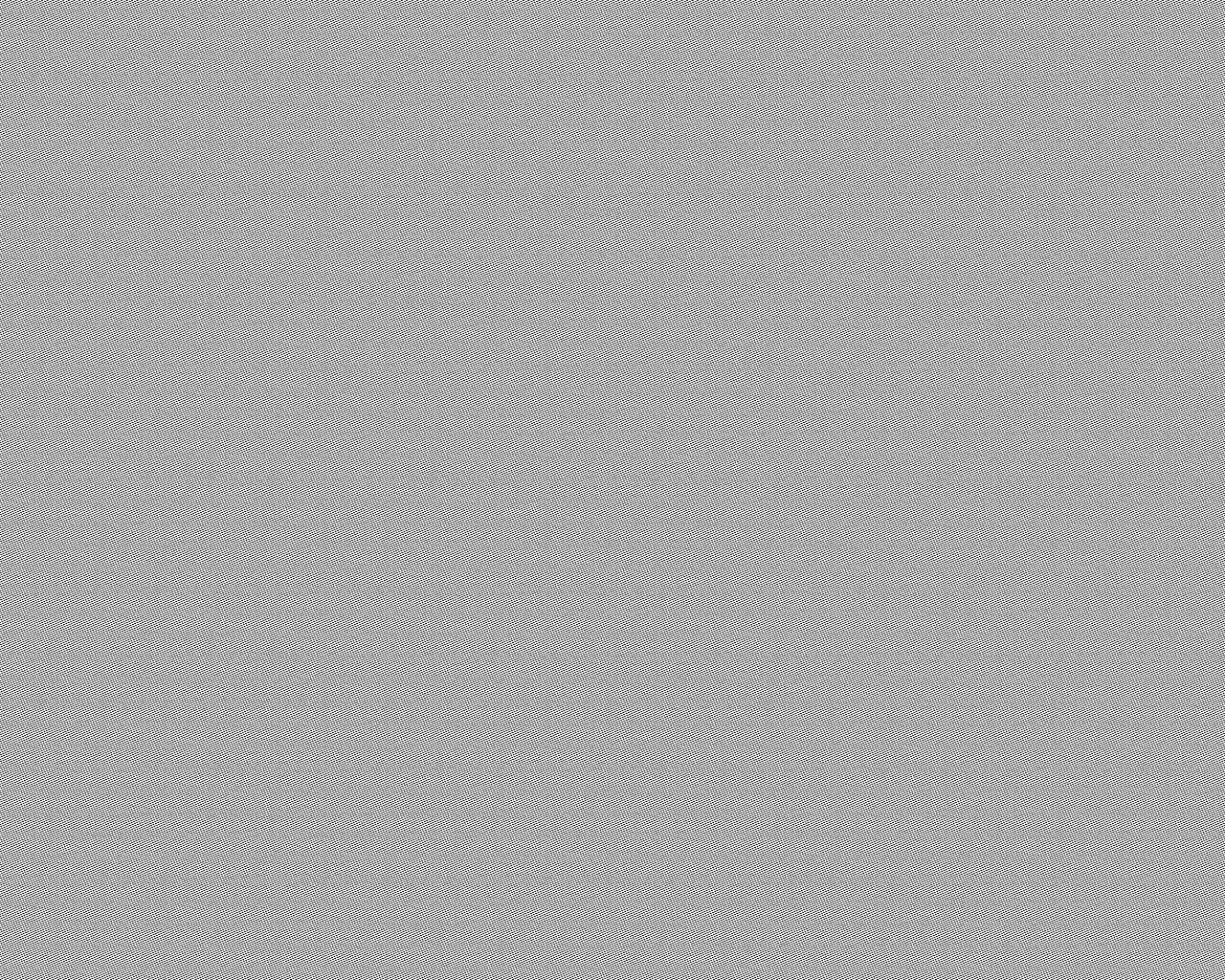}
\centering\includegraphics[trim= 987 990 280 24, clip=true, width=0.10\textwidth]{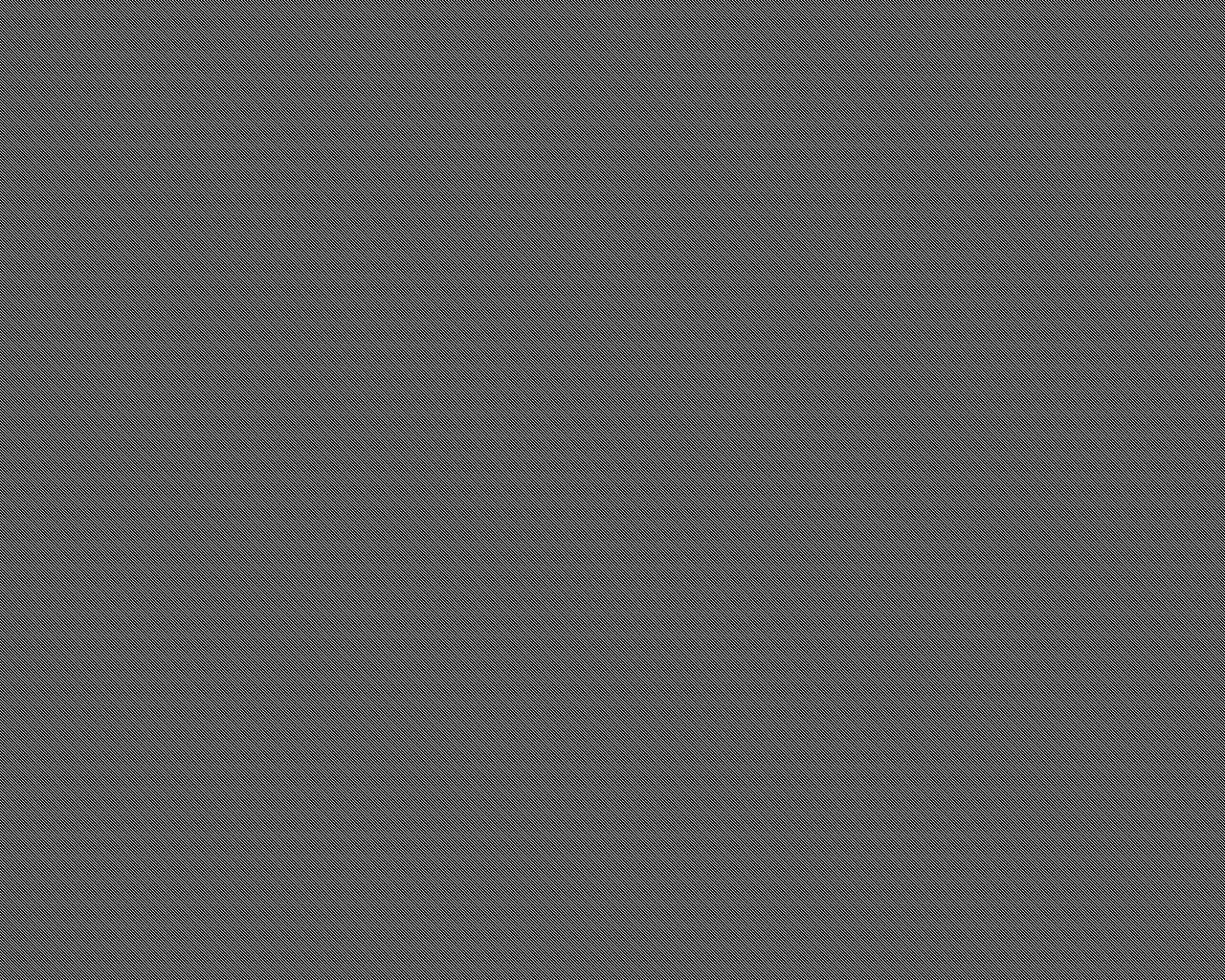}
\centering\includegraphics[trim= 987 990 280 24, clip=true, width=0.10\textwidth]{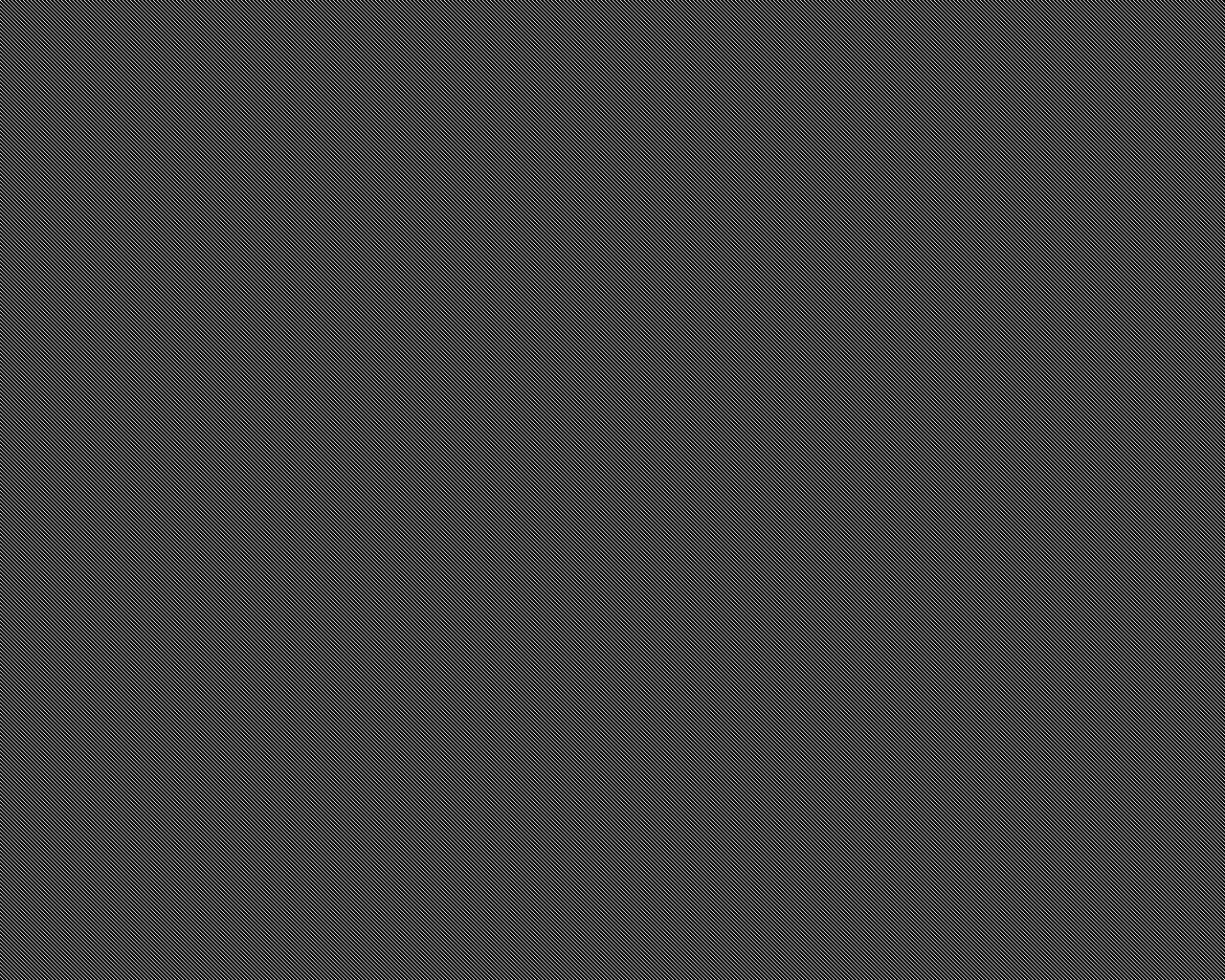}\\
\vspace{0.08cm}

\centering\includegraphics[width=0.10\textwidth]{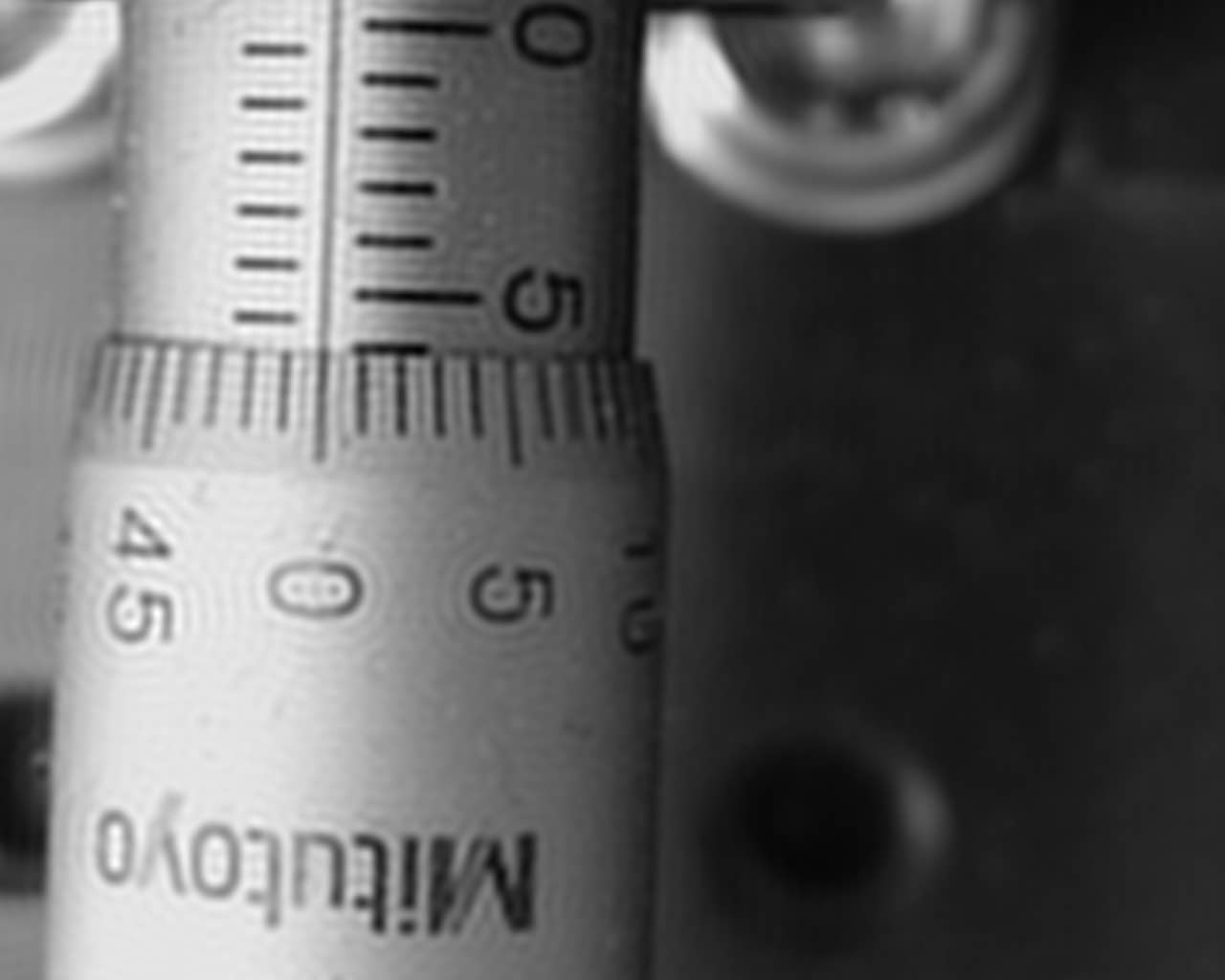}
\centering\includegraphics[width=0.10\textwidth]{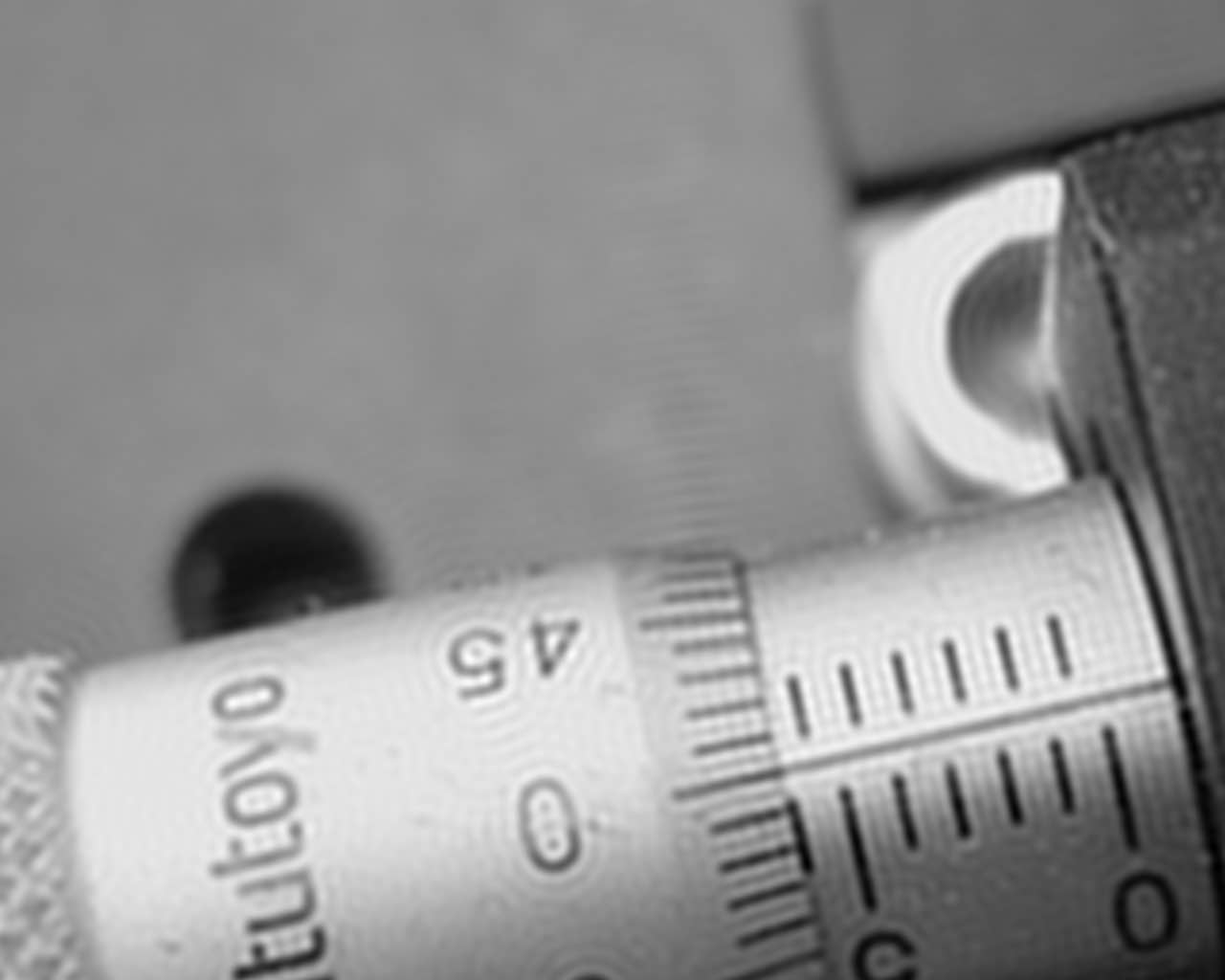}
\centering\includegraphics[width=0.10\textwidth]{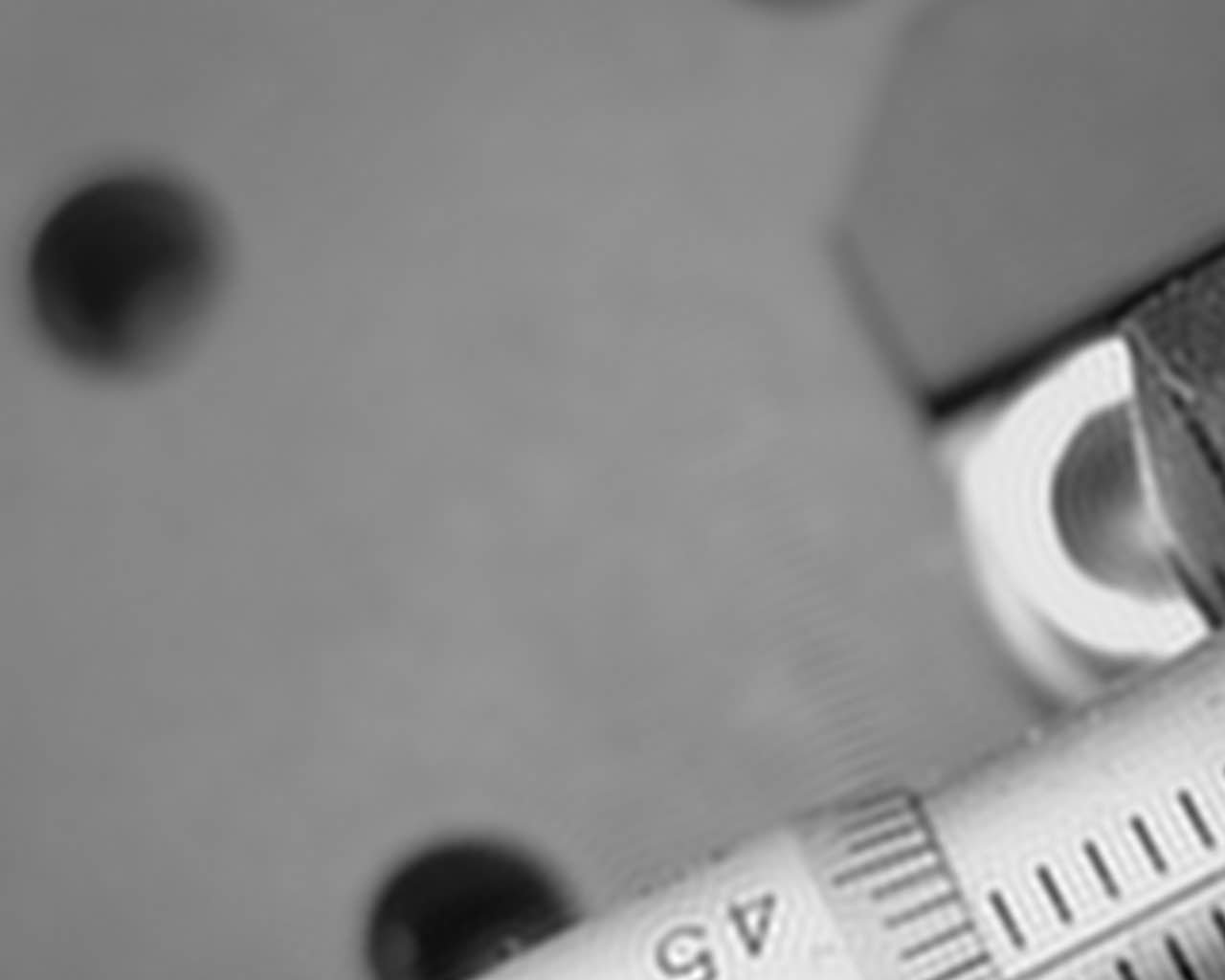}
\centering\includegraphics[width=0.10\textwidth]{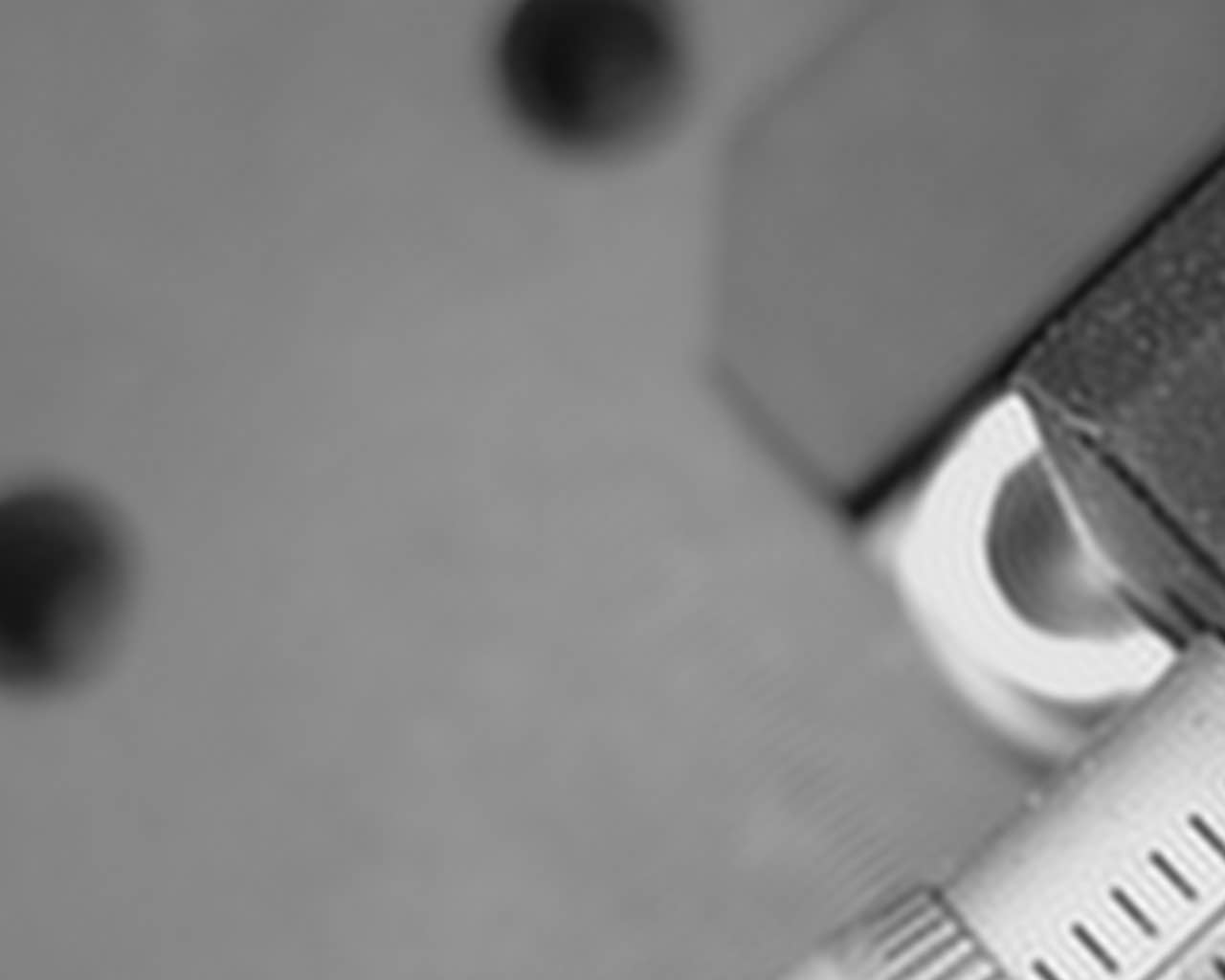}
\hspace{0.05\textwidth}
\centering\includegraphics[trim= 987 990 280 24, clip=true, width=0.10\textwidth]{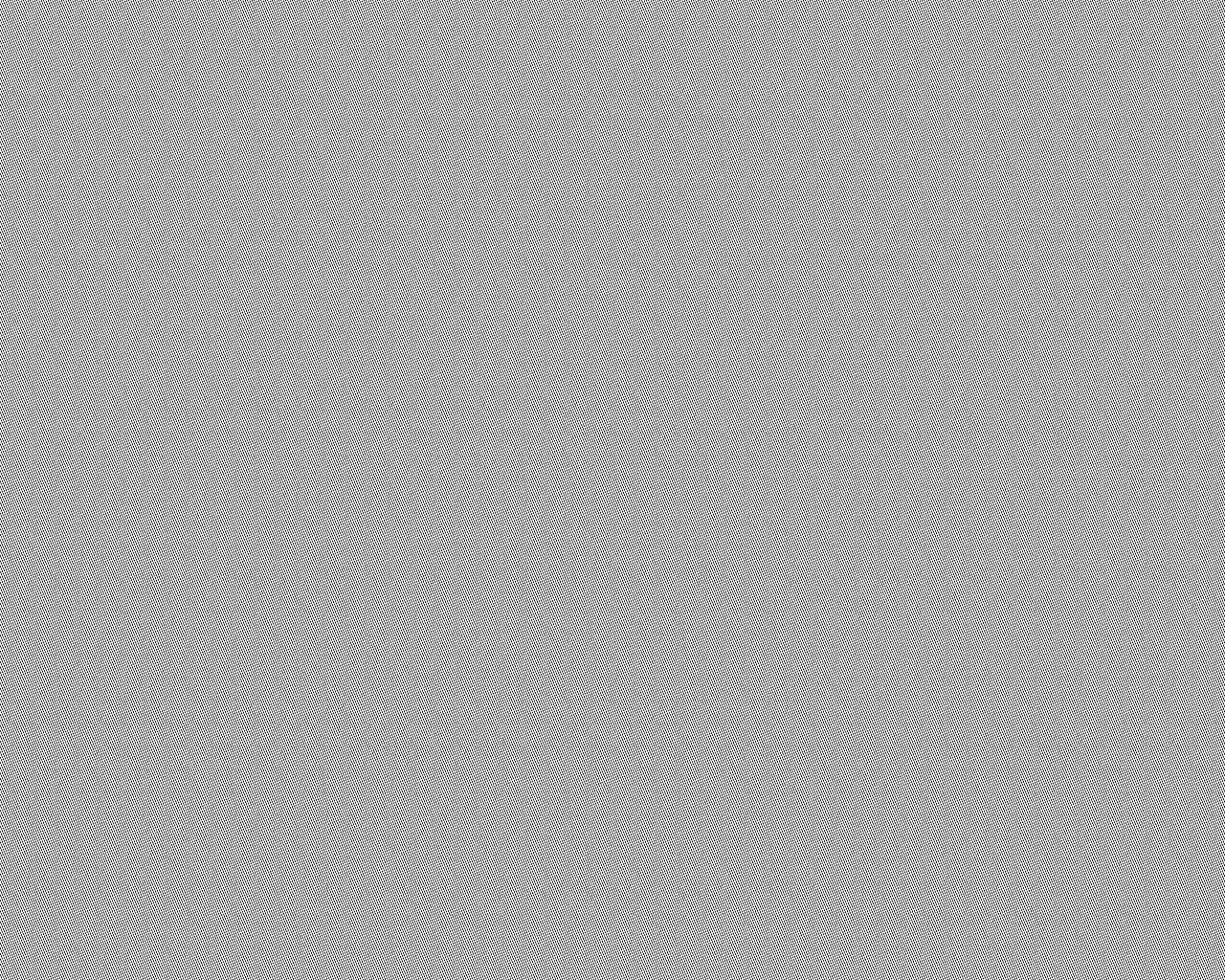}
\centering\includegraphics[trim= 987 990 280 24, clip=true, width=0.10\textwidth]{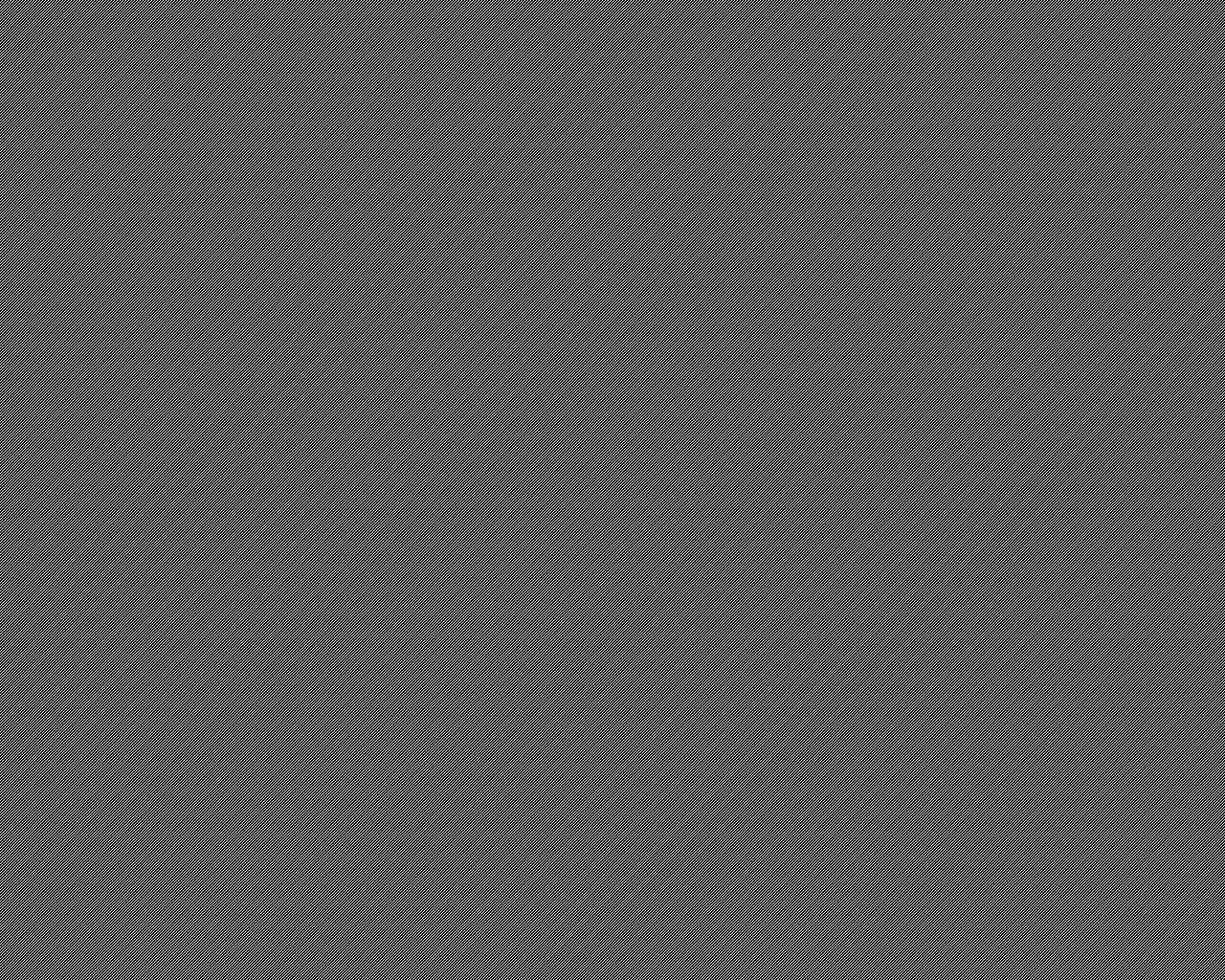}
\centering\includegraphics[trim= 987 990 280 24, clip=true, width=0.10\textwidth]{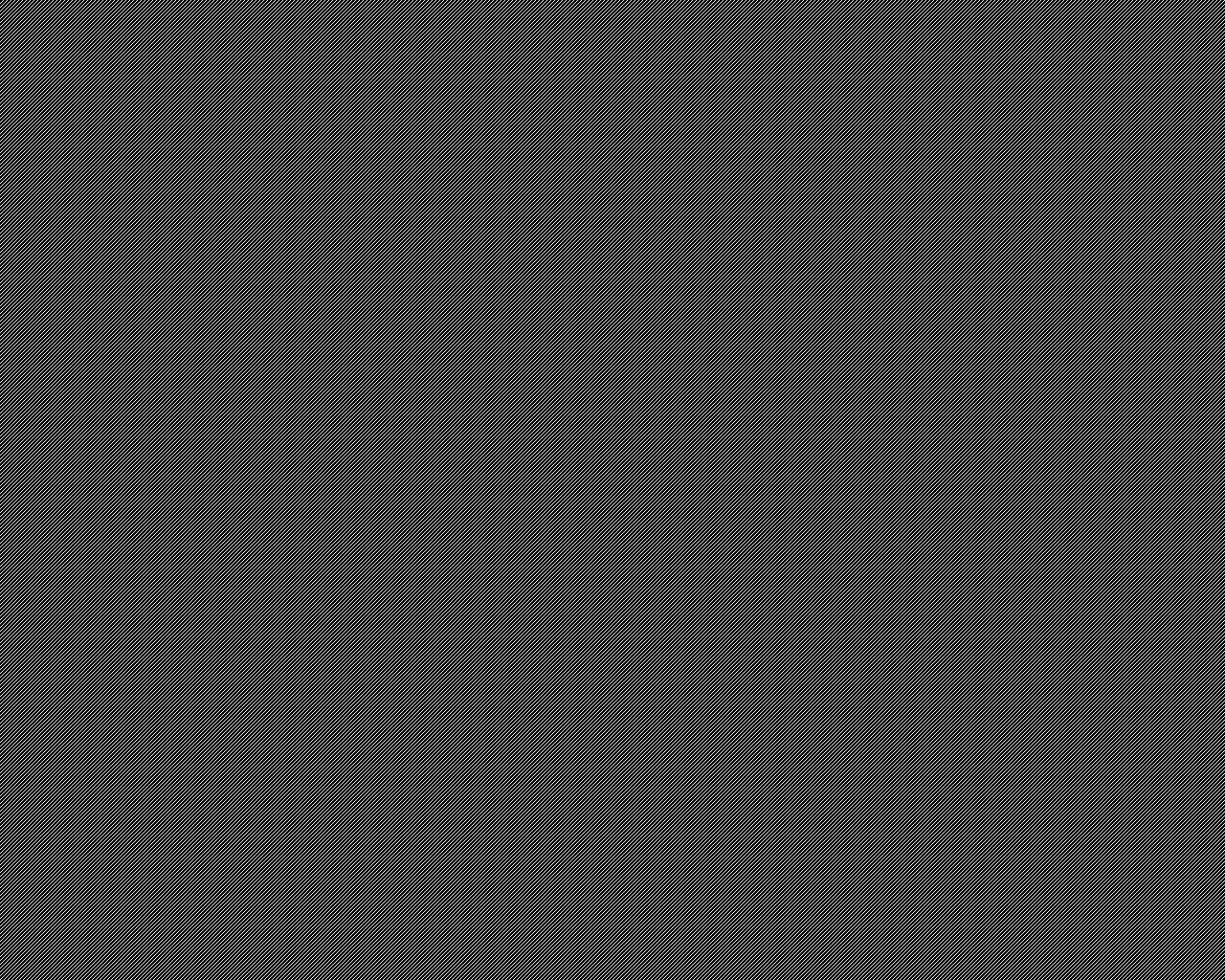}
\centering\includegraphics[trim= 987 990 280 24, clip=true, width=0.10\textwidth]{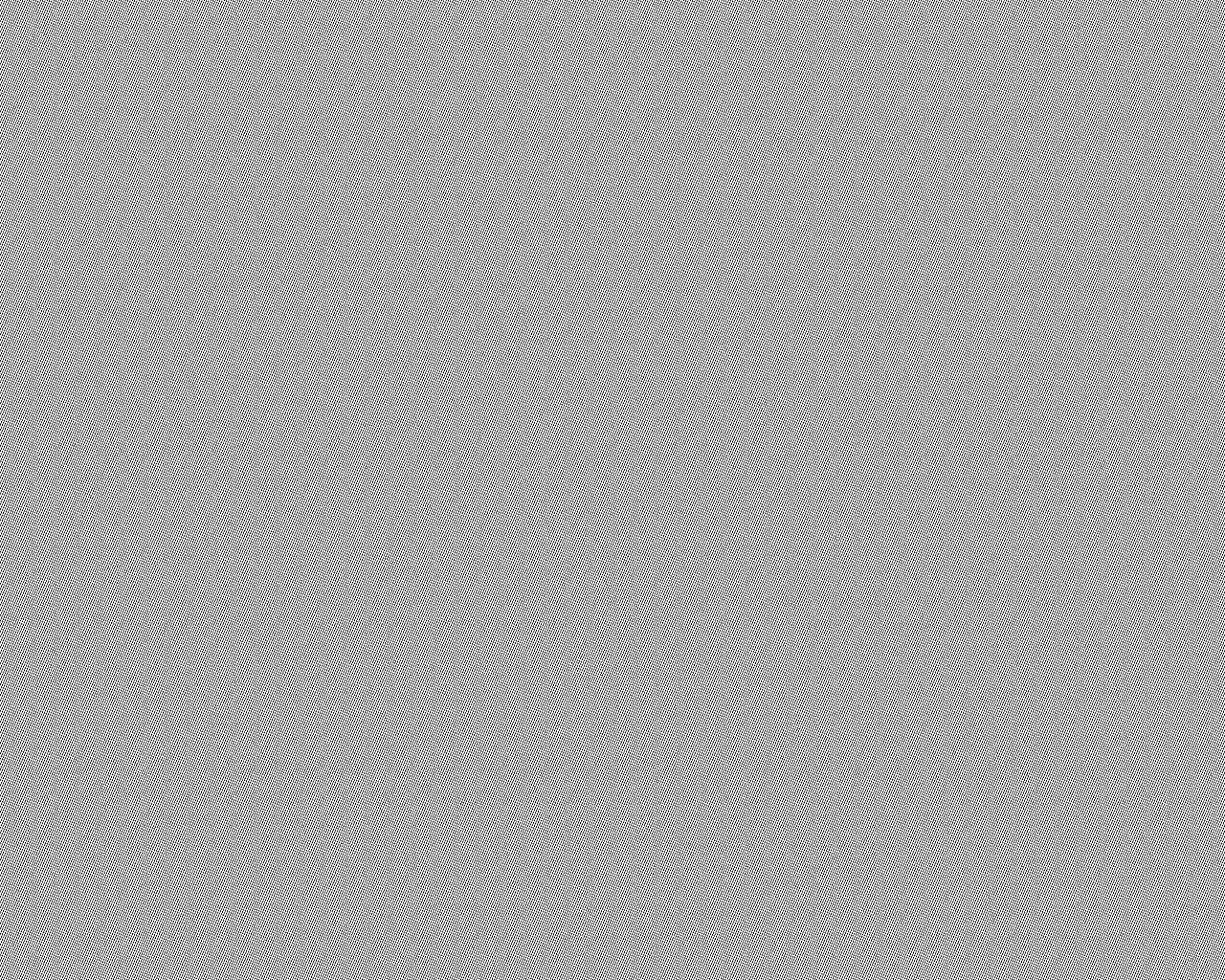}\\

(a) \hspace{0.45\textwidth} (b) \\

\vspace{0.2cm}
\hspace{-0.05\textwidth}
\centering\includegraphics[width=0.425\textwidth]{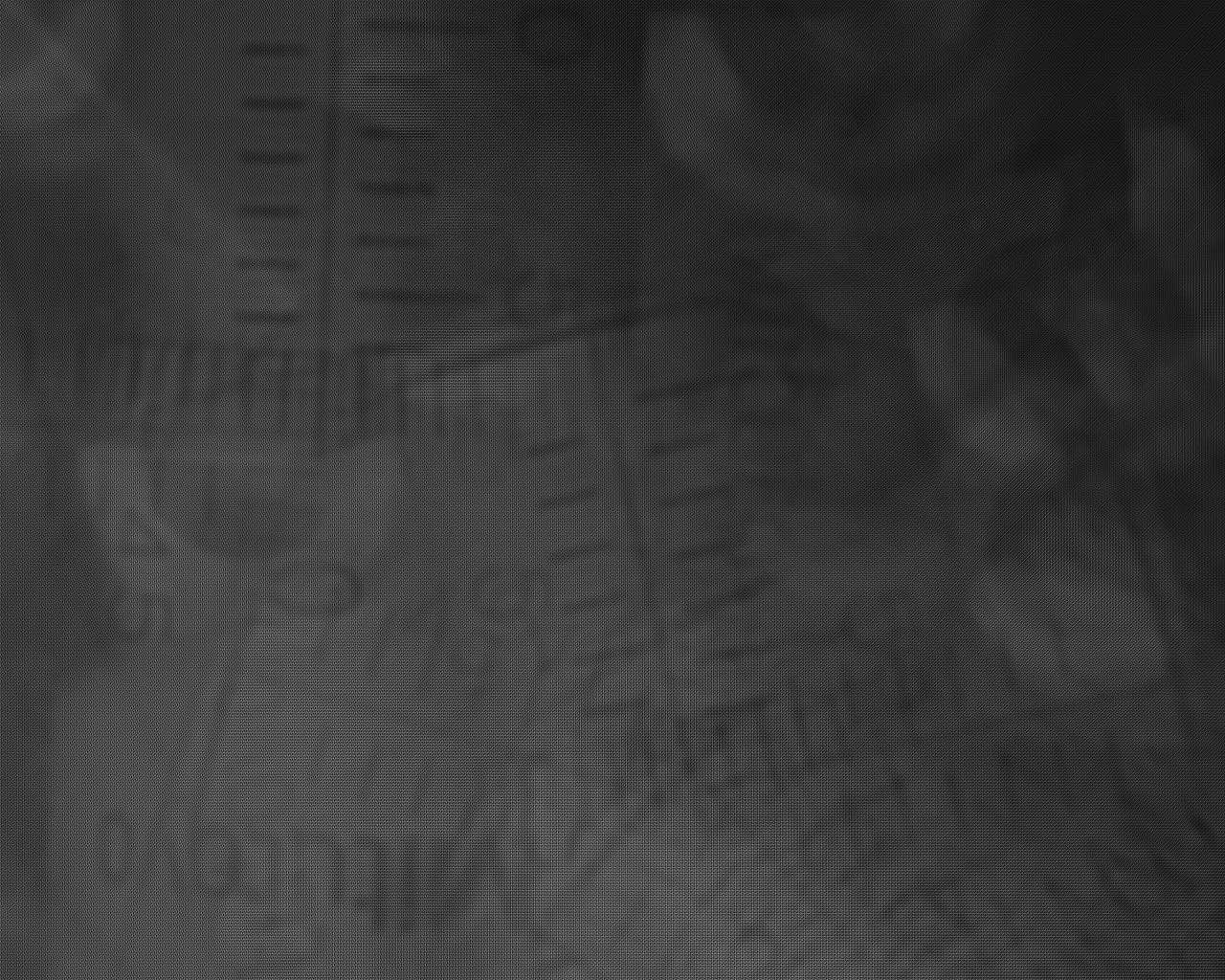}
\hspace{0.11\textwidth}
\centering\includegraphics[width=0.34\textwidth]{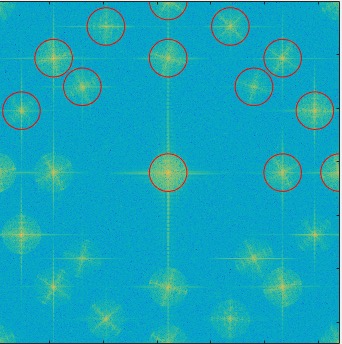} \\

(c) \hspace{0.45\textwidth} (d) \\

\vspace{0.2cm}

\centering\includegraphics[width=0.205\textwidth]{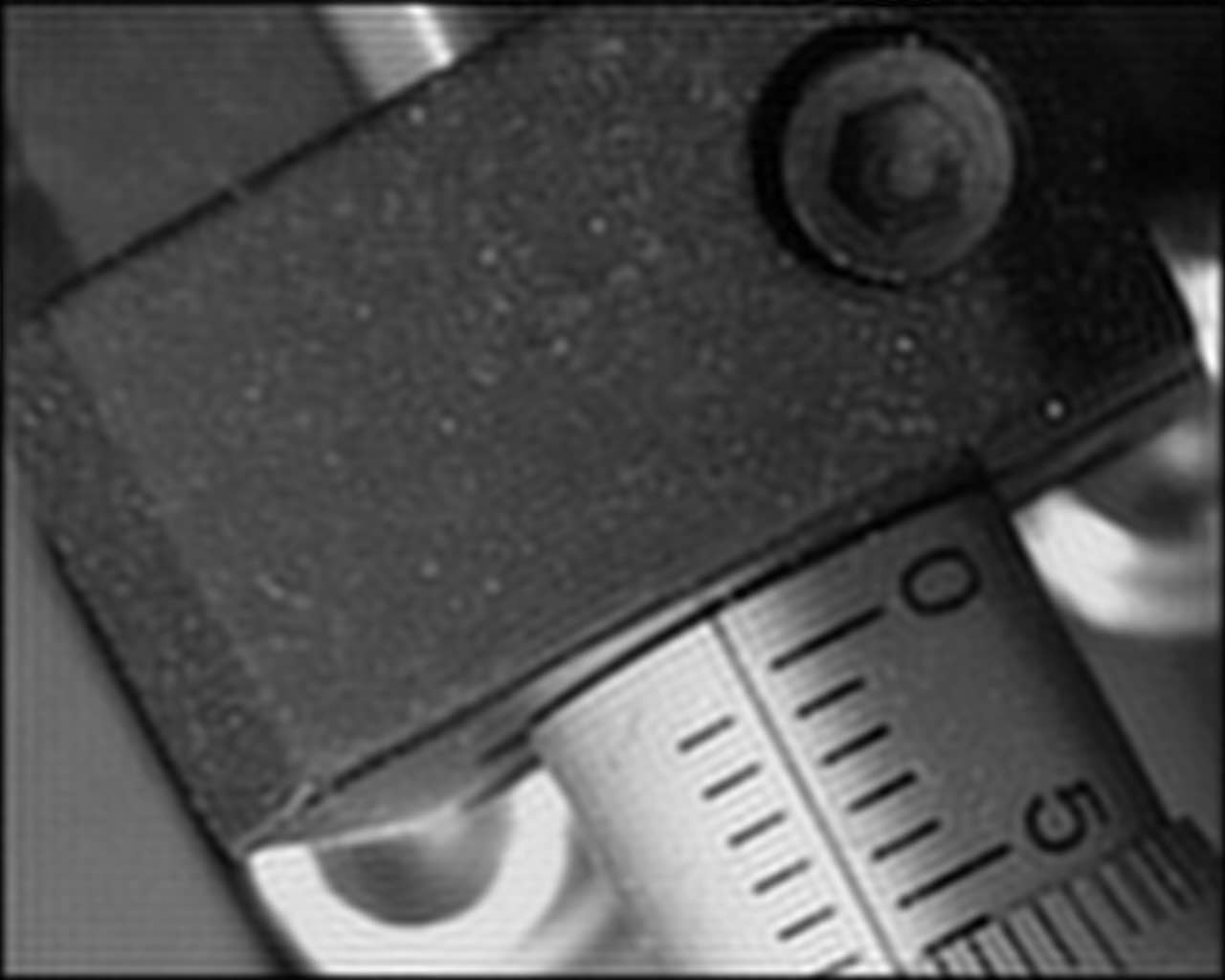}
\centering\includegraphics[width=0.205\textwidth]{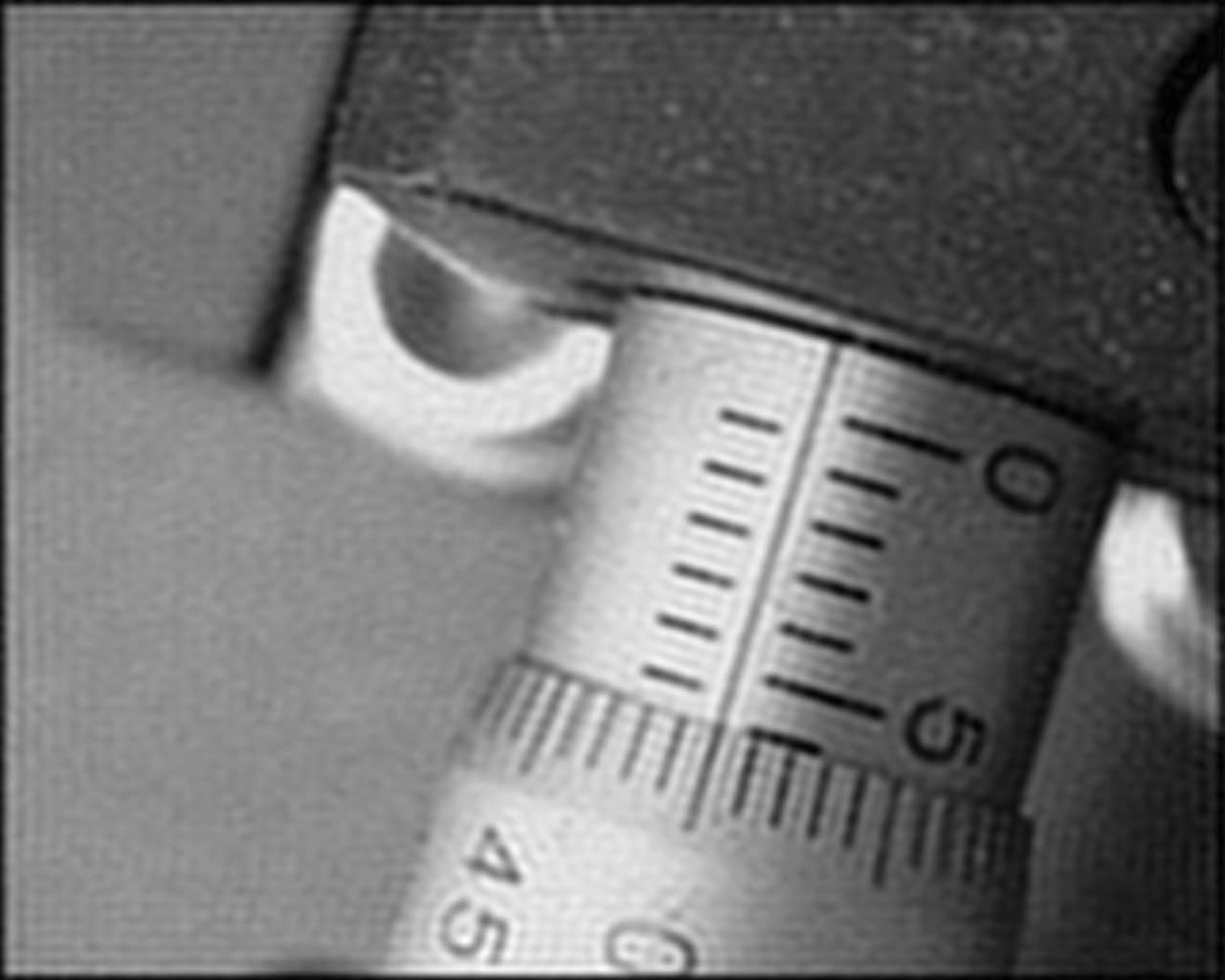}
%\hspace{0.05\textwidth}
\centering\includegraphics[width=0.205\textwidth]{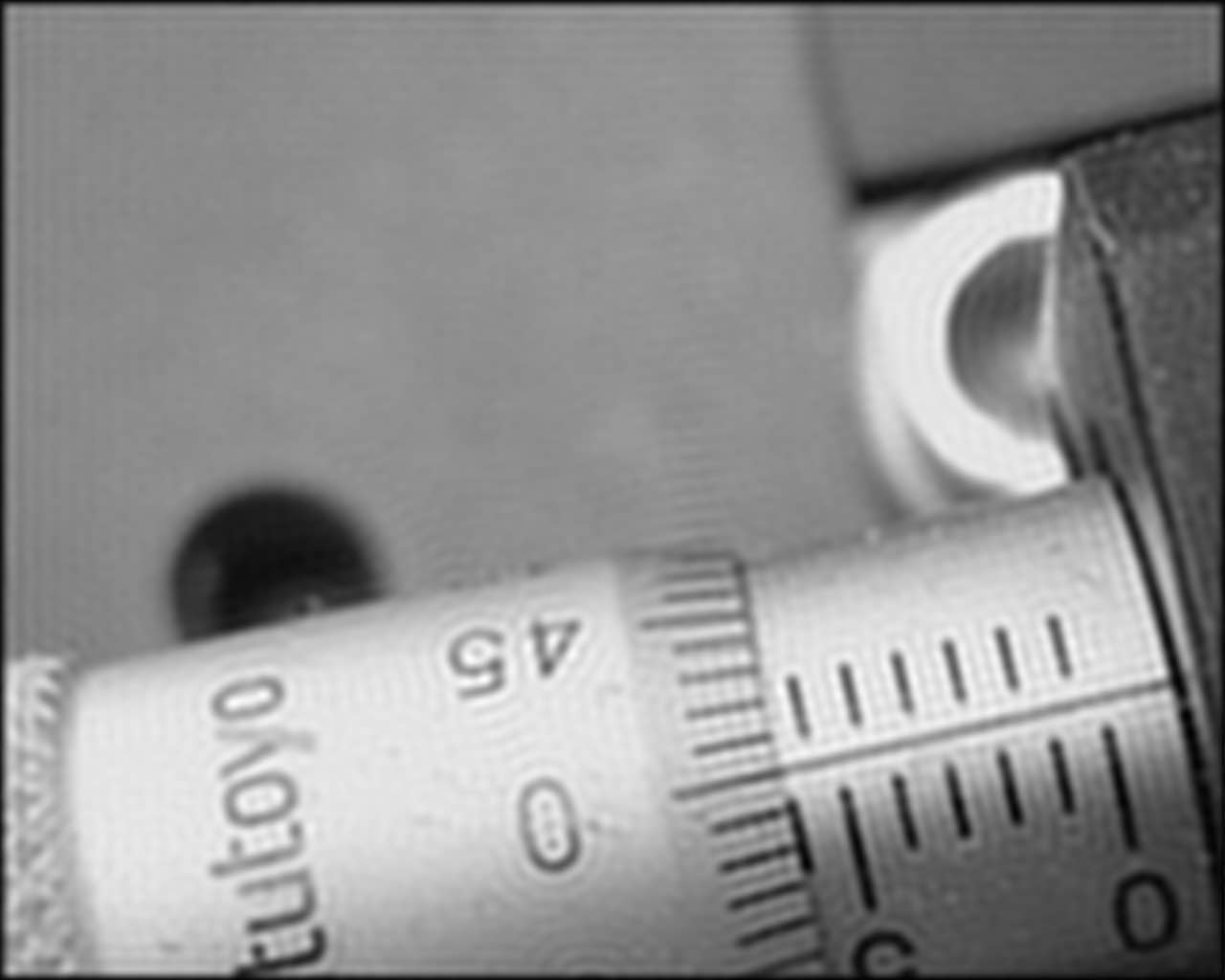}
\centering\includegraphics[width=0.205\textwidth]{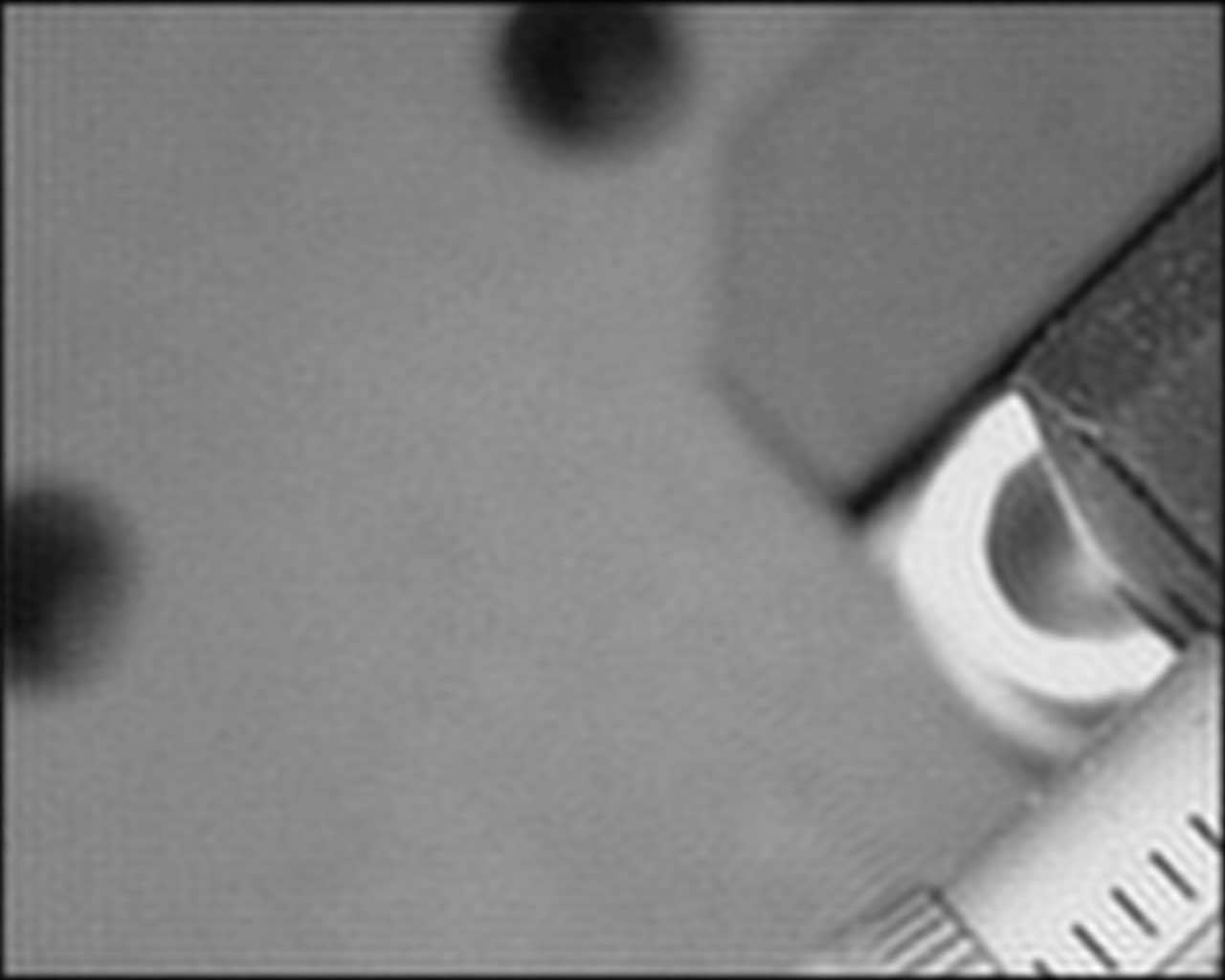}\\
(e) \\
%\vspace{0.1cm}

\caption{Simulation to demonstrate sub-exposure image extraction. (a) Original images, each with size $1280 \times 1024$ pixels. (b) Zoomed-in regions ($13 \times 10$ pixels) from the masks applied to the original images. (c) Simulated image capture. (d) Fourier transform (magnitude) of the captured image, with marked sidebands used to extract the sub-exposure images. (e) A subset of extracted sub-exposure images. (All 12 images can be successfully extracted; only four of them are shown here.)  }
\label{fig:Sim}
\end{figure}

The original images and zoomed-in regions from the masks are shown in Figure \ref{fig:Sim}(a) and Figure \ref{fig:Sim}(b), respectively. The highest frequency masks in horizontal and vertical directions are square waves.  The modulated image, obtained by multiplying each image with the corresponding mask and then averaging, is shown in Figure \ref{fig:Sim}(c). Its Fourier transform is shown in Figure \ref{fig:Sim}(d), where the placement of different sub-exposure images can be seen. The baseband and the sidebands from which the sub-exposure images are extracted are also marked in the figure. From these sidebands, we extract all 12 sub-exposure images. Four of the extracted sub-exposure images are shown in Figure \ref{fig:Sim}(e).

\begin{figure*}
\centering\includegraphics[width=0.48\textwidth]{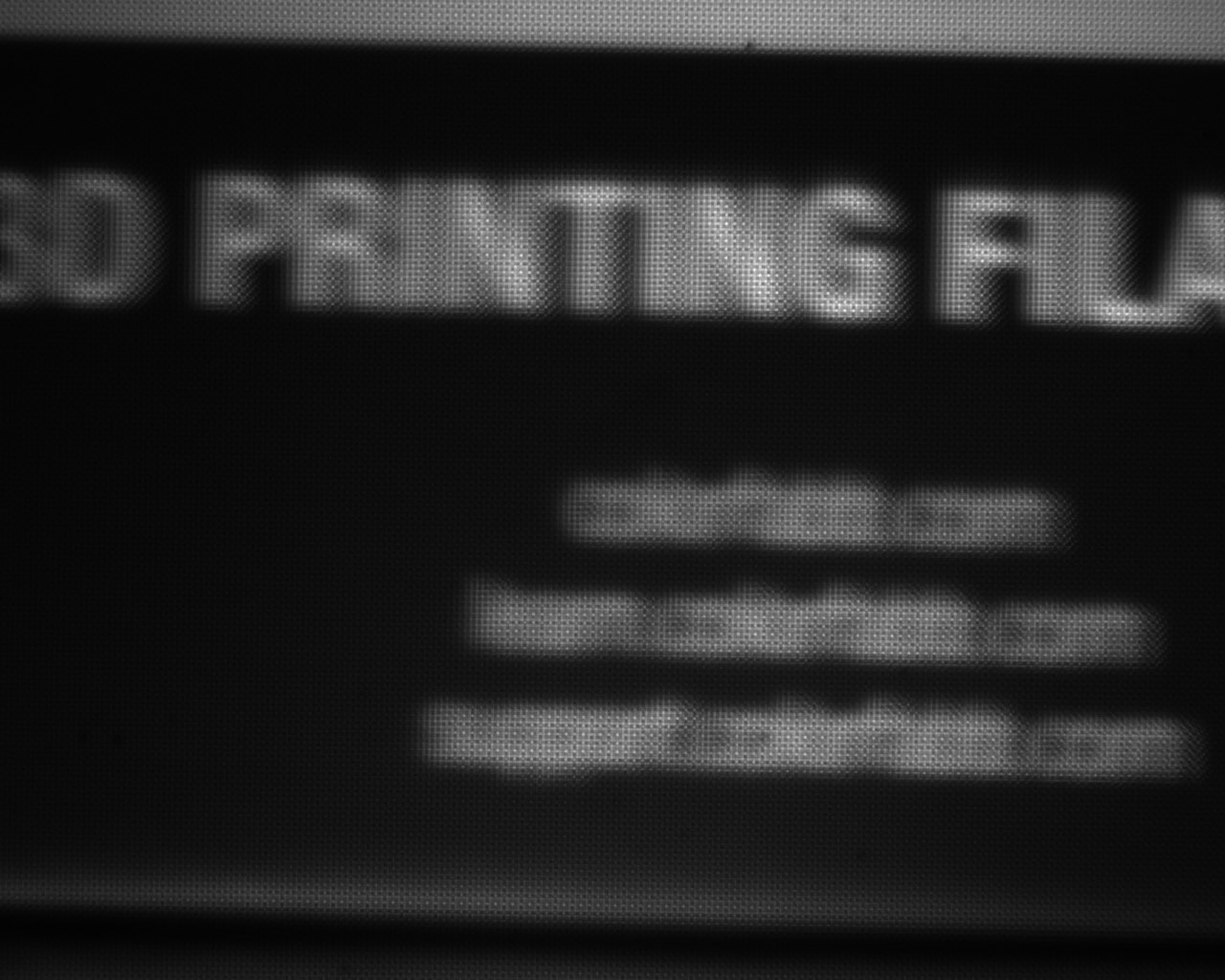}
\centering\includegraphics[width=0.48\textwidth]{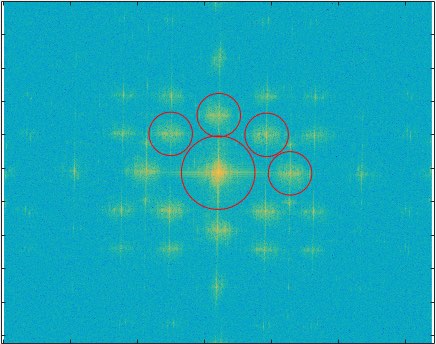}\\
(a) \hspace{6cm} (b)\\
\vspace{0.08cm}
\centering\includegraphics[width=0.48\textwidth]{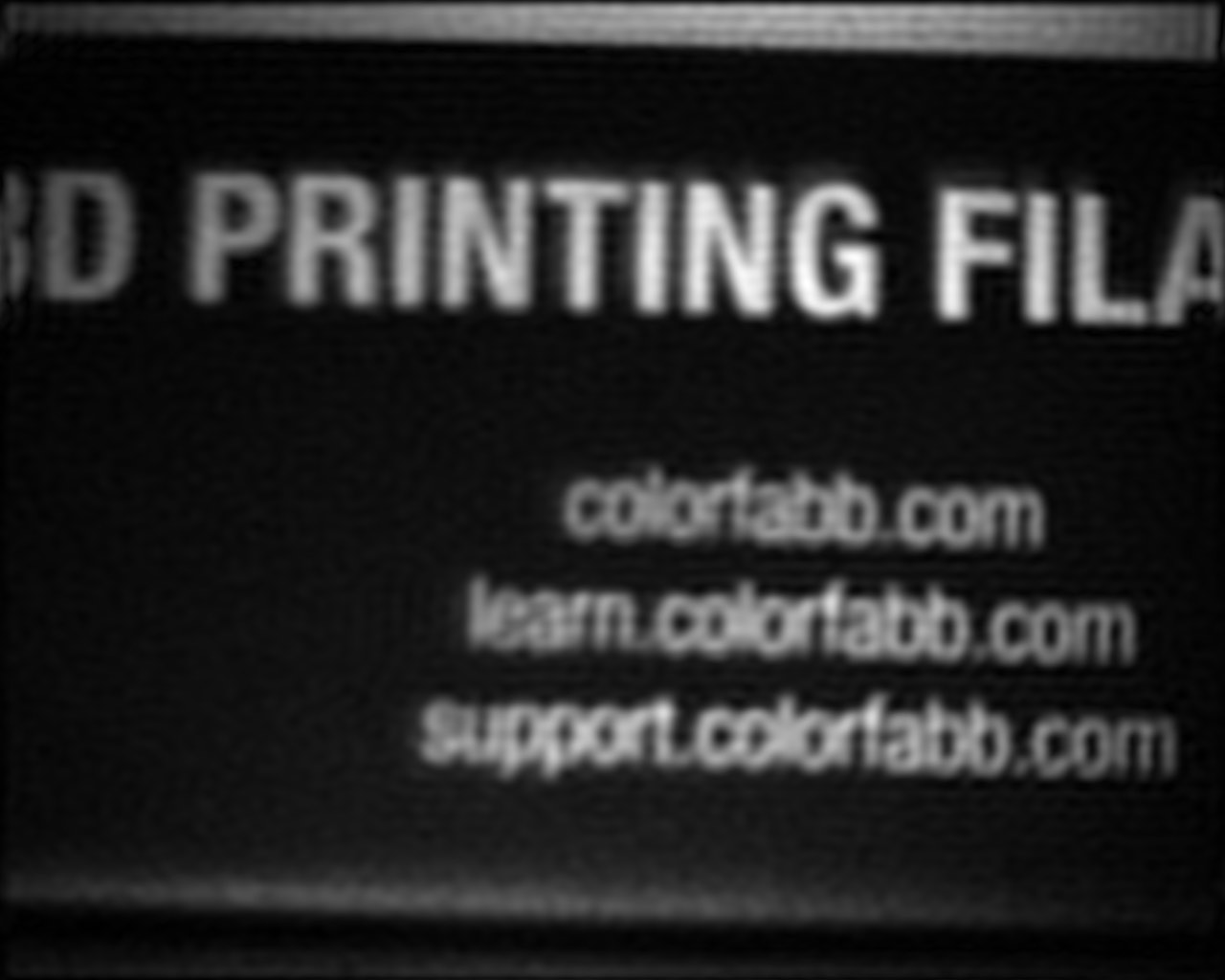}
\centering\includegraphics[width=0.48\textwidth]{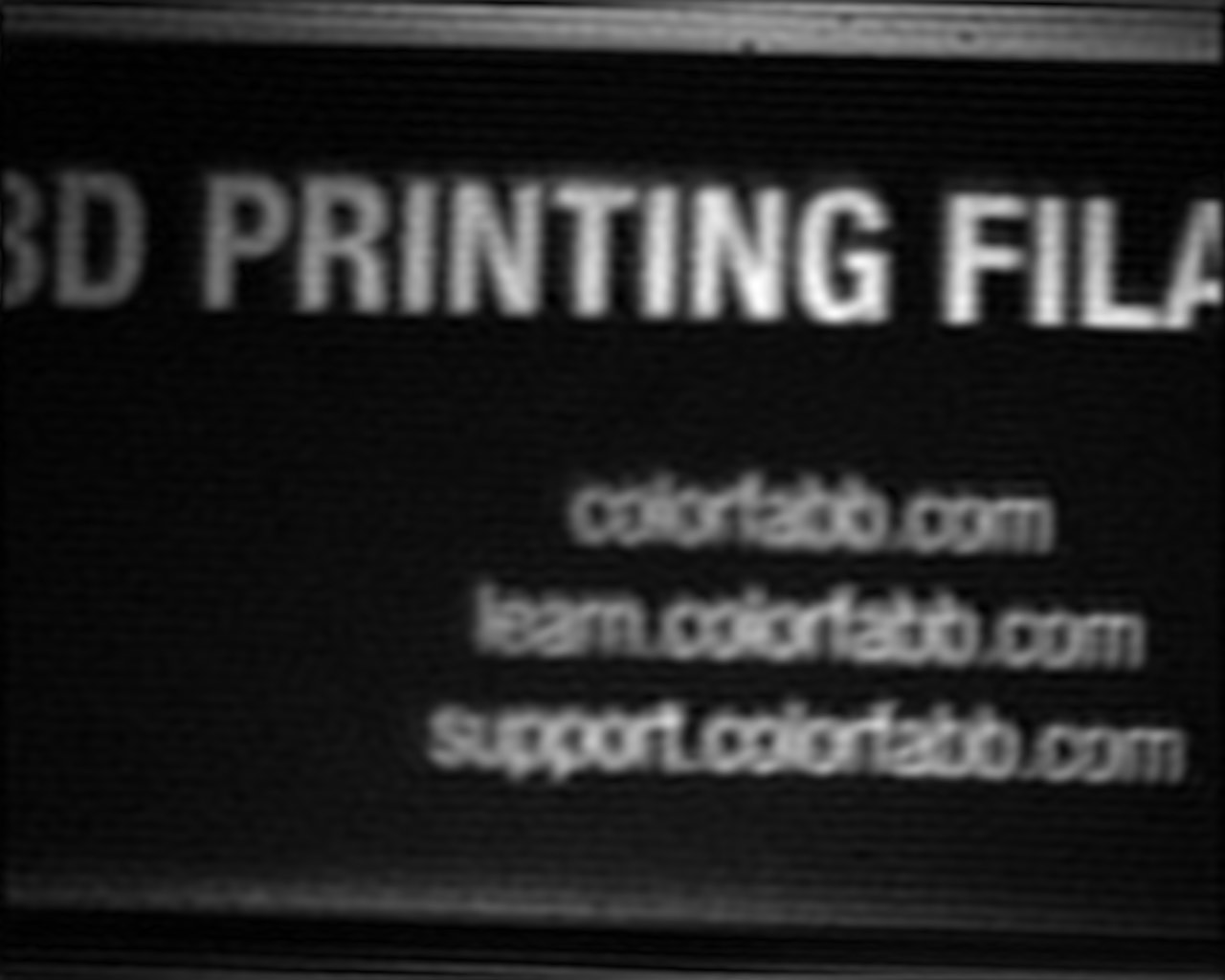}\\
(c) \hspace{6cm} (d)\\
\vspace{0.08cm}
\centering\includegraphics[width=0.48\textwidth]{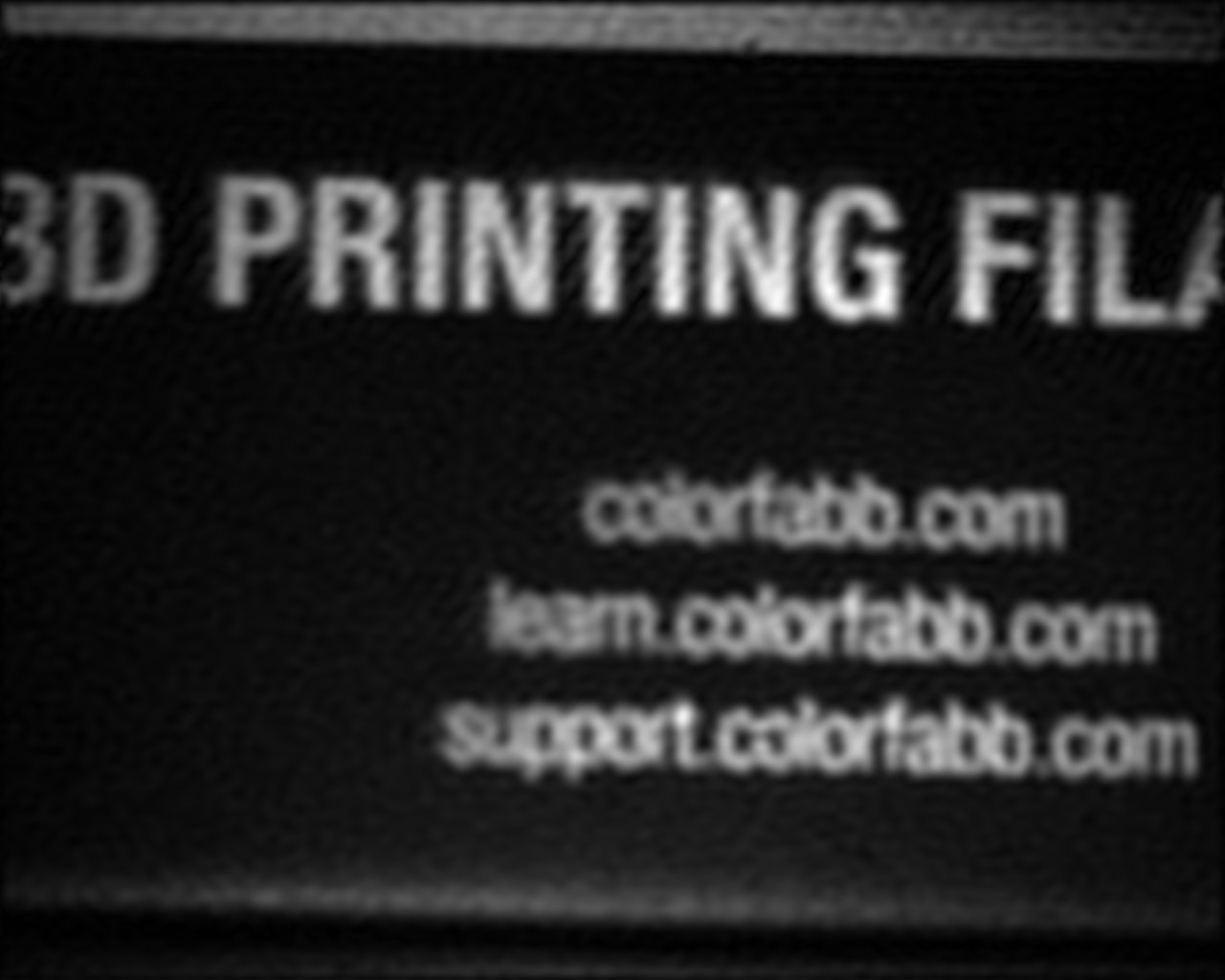}
\centering\includegraphics[width=0.48\textwidth]{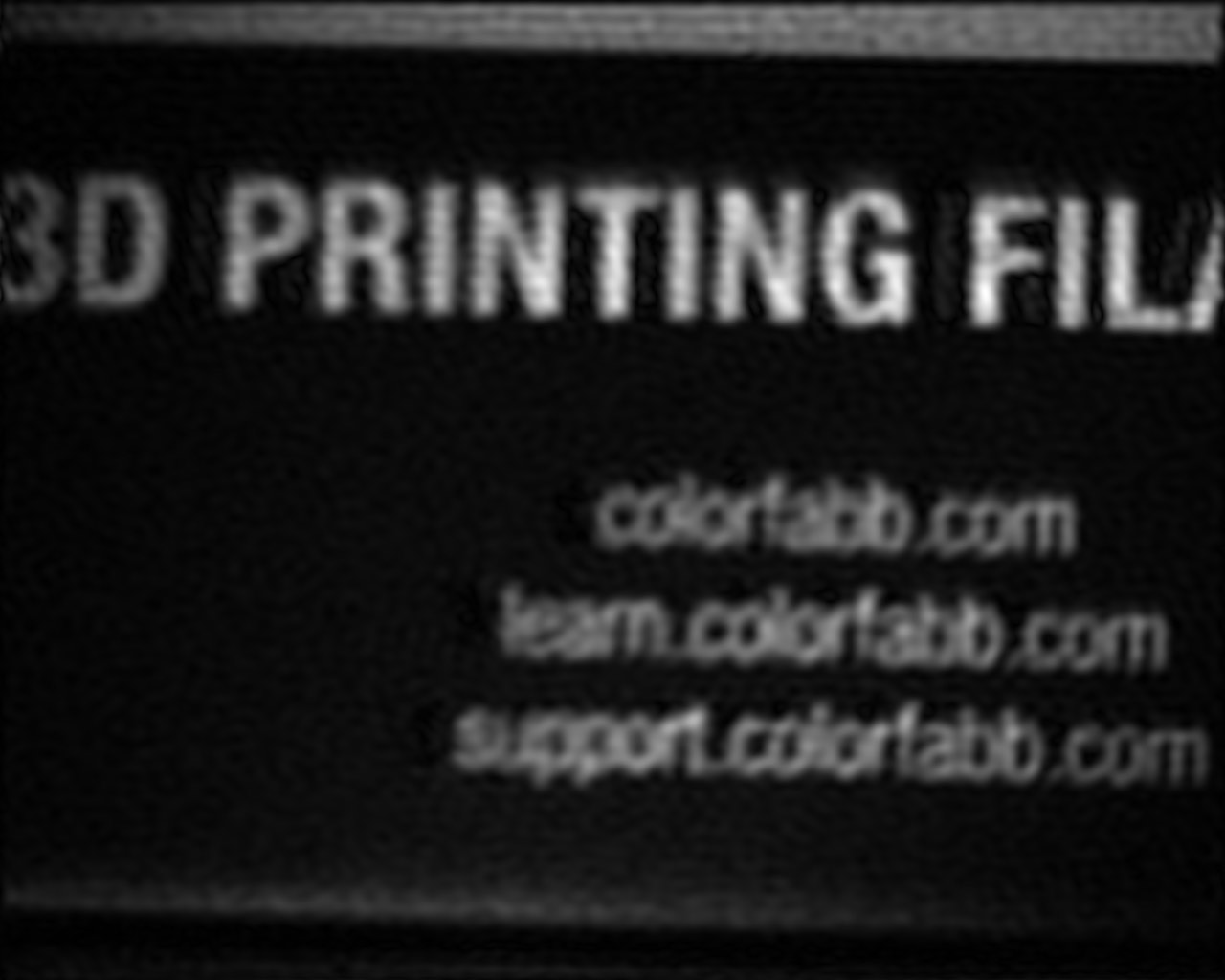}\\
(e) \hspace{6cm} (f)\\
\vspace{0.08cm}
\caption{Sub-exposure image extraction with FDMI. (a) Recorded exposure-coded image, (b) Fourier transform (magnitude) of the recorded image, (c) Sub-exposure image recovered from the horizontal sideband, (d) Sub-exposure image recovered from the vertical sideband, (e) Sub-exposure image recovered from the right diagonal sideband, (f) Sub-exposure image recovered from the left diagonal sideband.}
\label{fig: exp2cFourier}
\end{figure*}

The second experiment is realized with the prototype system. As shown in Figure \ref{fig: exp2cFourier}, a planar surface with a text printed on is moved in front of the camera. There are four sub-exposure masks: horizontal, vertical, right diagonal and left diagonal. (These masks are illustrated in Figure \ref{fig:ExposureMaskTypes}(d).) The exposure period for each mask is 10 milliseconds. The captured image is shown in Figure \ref{fig: exp2cFourier}(a). Its Fourier transform is shown Figure \ref{fig: exp2cFourier}(b). The recovered sub-exposure images are shown in Figures \ref{fig: exp2cFourier}(c)-(d). It is seen that the blurry text in the full-exposure capture becomes readable in the sub-exposure images.

In the experiment, the SLM pattern in horizontal and vertical directions has the highest possible spatial frequency; that is, one-pixel-on (full reflection) one-pixel-off (no reflection) SLM pattern. In other words, we cannot push the sidebands further away from the center in Figure \ref{fig: exp2cFourier}(b). (The reason why we have an extended Fourier region is that one SLM pixel (13.62$\mu m$) corresponds to about three sensor pixels (each with size 4.65$\mu m$). If we had one-to-one correspondence, the horizontal and vertical sidebands would appear all the way to the end of the Fourier regions because we have one-on one-off binary pattern for them.)  To encode and extract more sub-exposure images, the sub-exposure images should be passed through an optical low-pass filter to further reduce the frequency extent.      

The spatial resolution of the optical system is controlled by several factors, including the lens quality, SLM and sensor dimensions. Since the optical modulation is done by the SLM pixels, SLM pixel size is the main limiting factor of the spatial resolution on the sensing side. On the computational side, the FDMI technique requires band-limited signals, which is also a limiting factor on spatial resolution. In the first experiment, the radius of a band-pass filter used to recover a sub-exposure image is about one-ninth of the available bandwidth, as seen in Figure \ref{fig:Sim}; that is, the spatial frequency is reduced to one-ninth of the maximum available resolution as a result of the FDMI process. In the second experiment, the radius of the band-pass filters is about one-fourth of the available bandwidth; that is, the spatial resolution is reduced to one-quarter of the maximum spatial resolution. It should, however, be noted that the bandwidth of a natural image is typically small; therefore, the effective reduction in spatial resolution is not as much in practice. 

\begin{figure*}
\centering\includegraphics[width=0.48\textwidth]{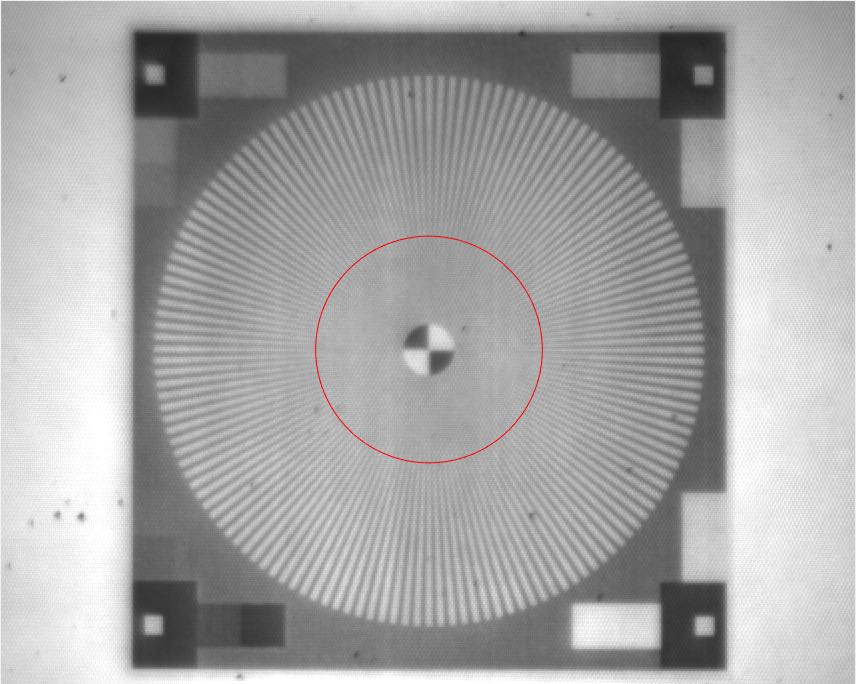}
\centering\includegraphics[width=0.48\textwidth]{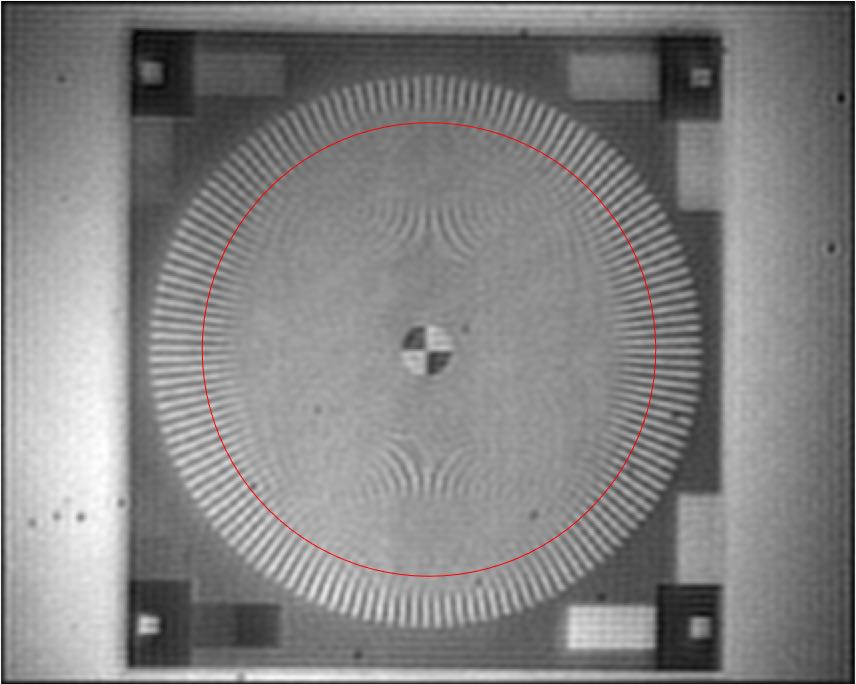}\\
(a) \hspace{6cm} (b) \\
\vspace{0.08cm}
\caption{Spatial resolution of the optical system. (a) Image without any optical modulation of the SLM, (b) Extracted sub-exposure image when the SLM pattern is vertical square wave with one-pixel-on one-pixel-off SLM pixels.}
\label{fig: expResolution}
\end{figure*}

To quantify the overall spatial resolution of our system, we took two photos of a {\it Siemens Star}, one without any optical modulation of the SLM (that is, full reflection), and the other with an SLM pattern of maximum spatial frequency (one-pixel-on one-pixel-off SLM pixels). The images are shown in Figure \ref{fig: expResolution}. The radii beyond which the radial lines can be distinguished are marked as {\it red circles} in both images. According to \cite{loebich2007}, the spatial resolution is {\it (Number of cycles for the Siemens Star) $\times$ (Image height in pixels) / (2$\pi$(Radius of the circle))}. It turns out that the resolution of the optical system (without any SLM modulation) is  138 line width per picture height; while it is 69 line width per picture height for the sub-exposure image. The reduction in spatial resolution is $50\%$, which is expected considering the spatial frequency of the SLM pattern.

\begin{figure*}
\centering\includegraphics[width=0.35\textwidth]{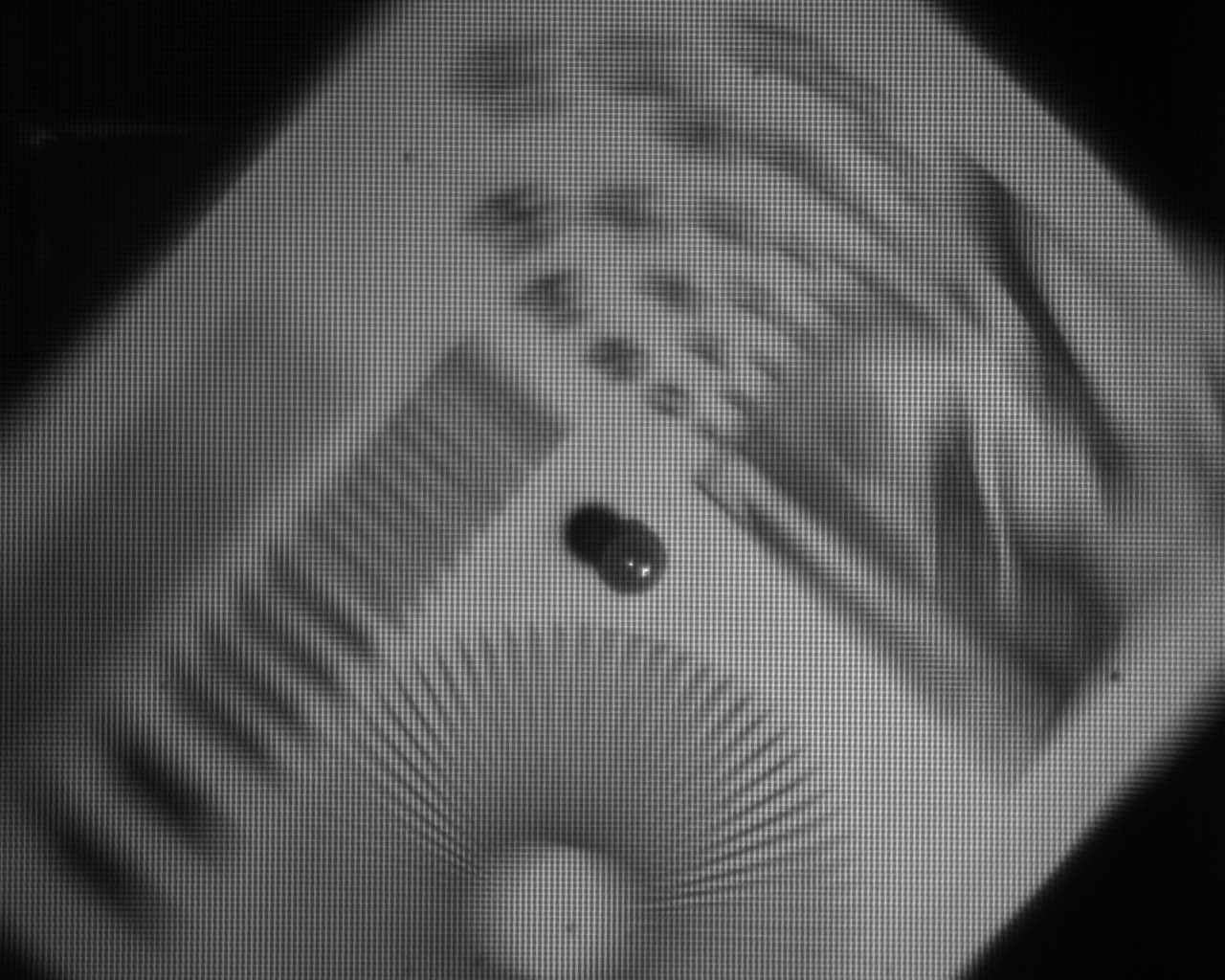}
\centering\includegraphics[trim= 590 690 480 140, clip=true, width=0.3\textwidth]{MFO}
\hspace{0.1cm}
\centering\includegraphics[width=0.28\textwidth]{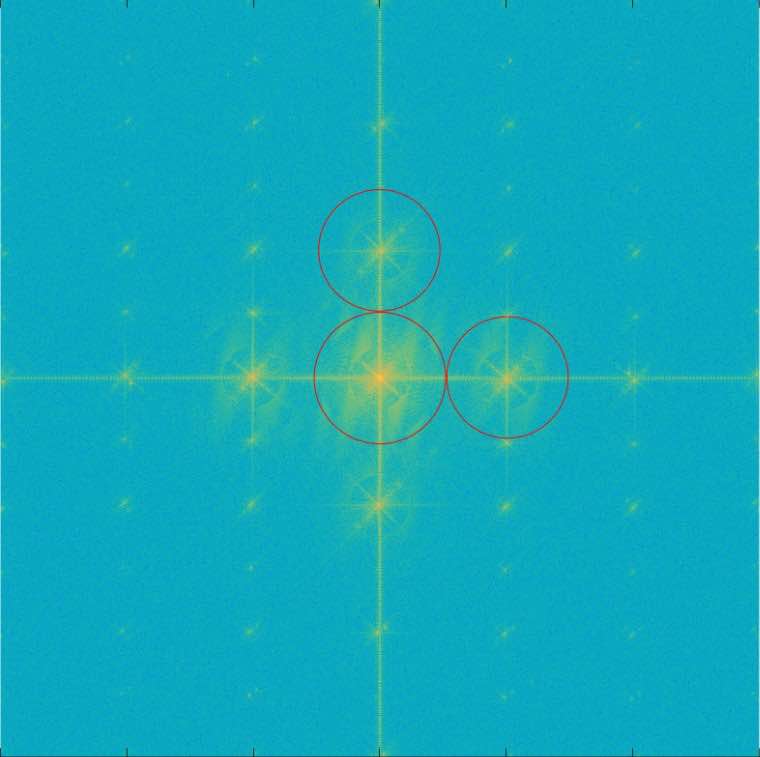}\\
(a) \hspace{3.5cm} (b) \hspace{3.5cm} (c) \\
\vspace{0.08cm}
\centering\includegraphics[width=0.35\textwidth]{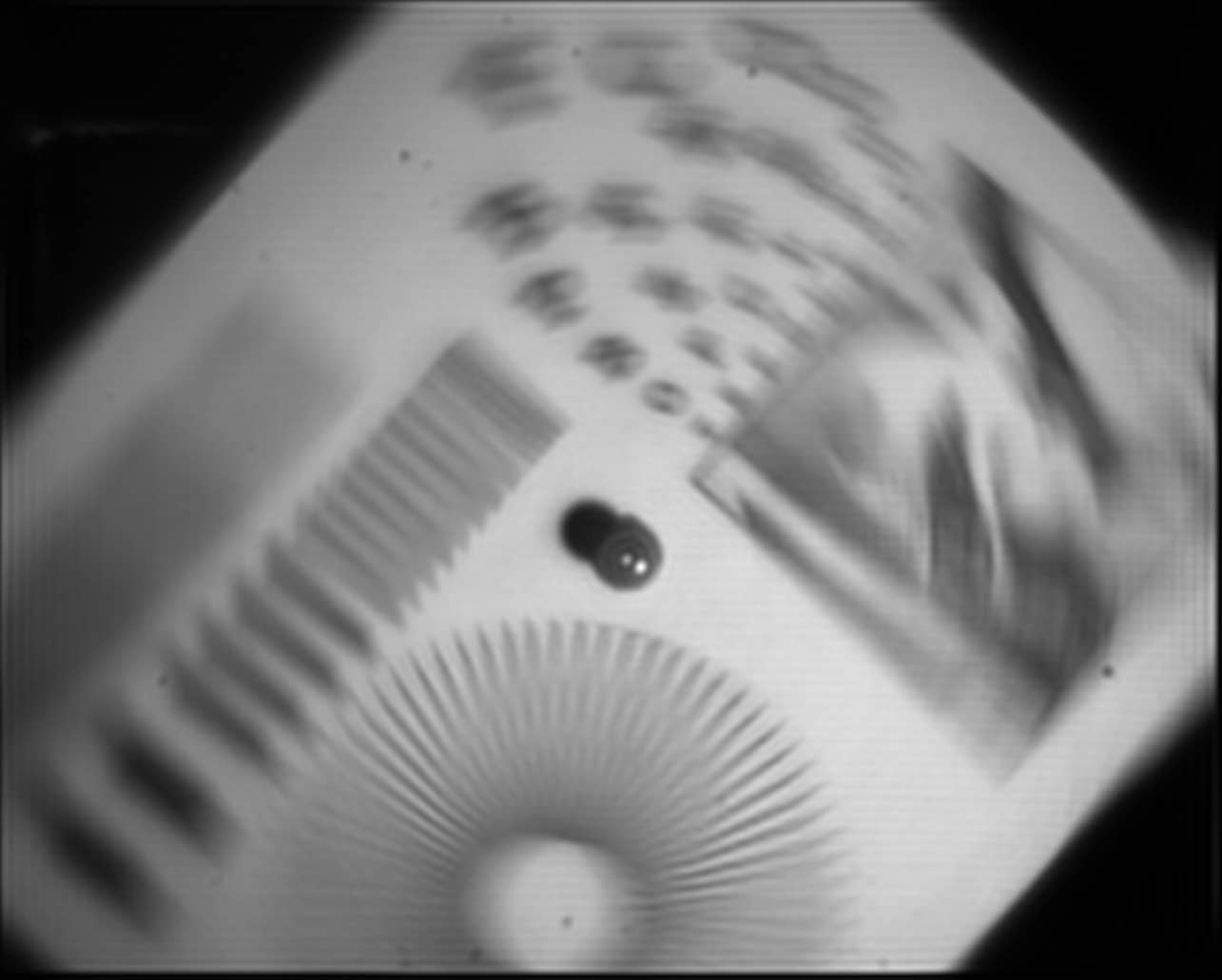}
\centering\includegraphics[trim= 450 550 420 100, clip=true, width=0.3\textwidth]{MFC}
\centering\includegraphics[trim= 300 220 520 380, clip=true, width=0.3\textwidth]{MFC}\\
(d) \hspace{8cm} ~ \\
\vspace{0.08cm}
\centering\includegraphics[width=0.35\textwidth]{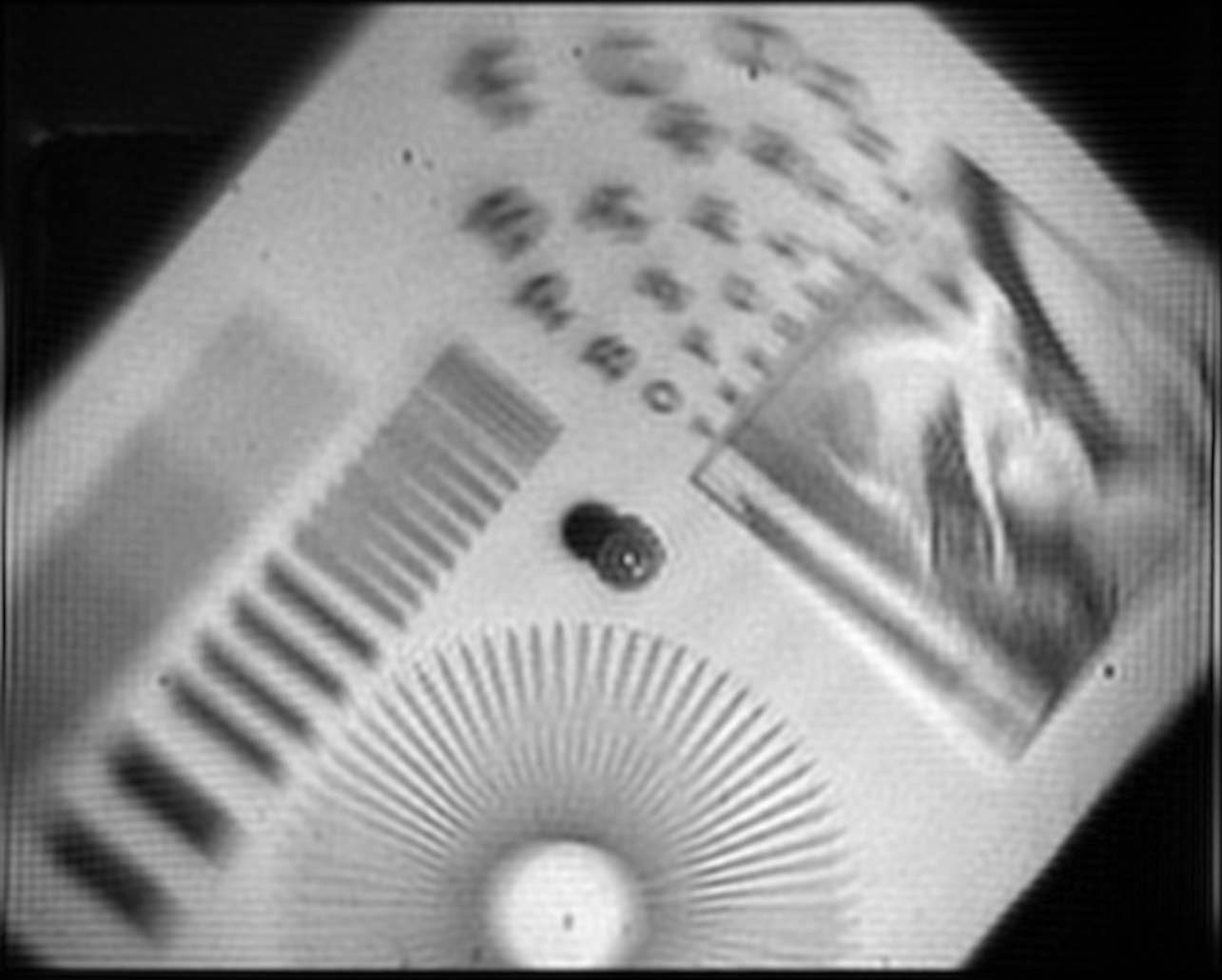}
\centering\includegraphics[trim= 450 550 370 050, clip=true, width=0.3\textwidth]{MFV}
\centering\includegraphics[trim= 300 220 520 380, clip=true, width=0.3\textwidth]{MFV}\\
%\centering\includegraphics[trim= 580 250 150 280, clip=true, width=0.2\textwidth]{MFV}\\
(e) \hspace{8cm} ~ \\
\vspace{0.08cm}
\centering\includegraphics[width=0.35\textwidth]{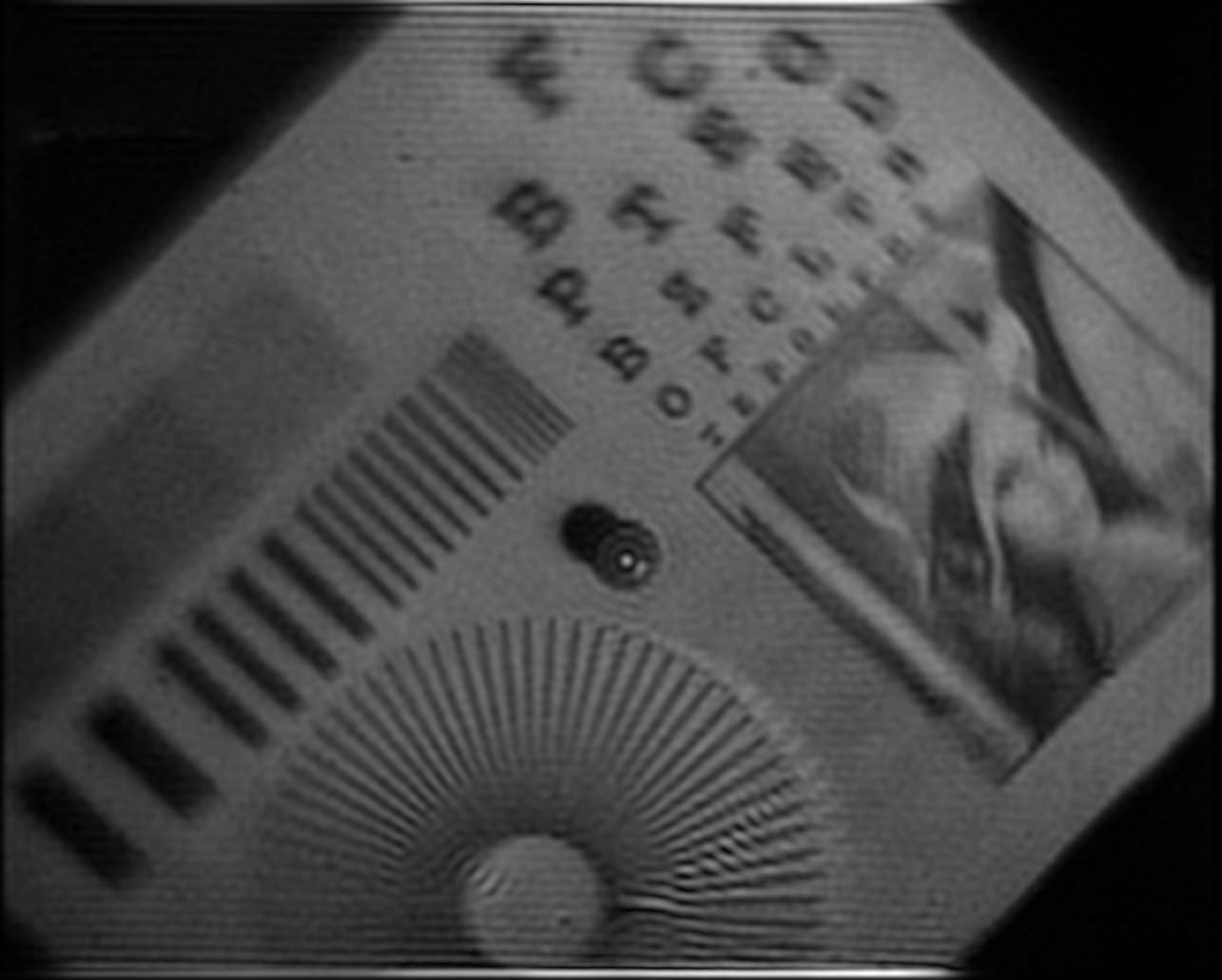}
\centering\includegraphics[trim= 450 550 370 050, clip=true, width=0.3\textwidth]{MFH}
\centering\includegraphics[trim= 300 220 520 380, clip=true, width=0.3\textwidth]{MFH}\\
%\centering\includegraphics[trim= 580 250 150 280, clip=true, width=0.2\textwidth]{MFH}\\
(f) \hspace{8cm} ~ \\
\vspace{0.08cm}
\caption{Sub-exposure image extraction with FDMI. (a) Recorded exposure-coded image, (b) A zoomed-in region from (a), (c) Fourier transform (magnitude) of the recorded image, (d) Image extracted from the baseband, corresponding to 45 millisecond exposure period, (e) Image extracted from the horizontal sideband, corresponding to 30 millisecond exposure period, (f) Image extracted from the vertical sideband, corresponding to 15 millisecond exposure period. Zoomed-in regions for (d), (e), and (f) are also shown.}
\label{fig:exp1bFourier}
\end{figure*}

In the third experiment, shown in Figure \ref{fig:exp1bFourier}, the target object is a printout, which is rotated as the camera captures an image. The exposure time of the camera is set to 45 milliseconds. In the first 30 millisecond period, the SLM pattern is a square wave in horizontal direction; in the second 15 millisecond period, the SLM pattern is a square wave in vertical direction.  The captured image is shown in Figure \ref{fig:exp1bFourier}(a). A zoomed-in region is shown in Figure \ref{fig:exp1bFourier}(b), where horizontal and vertical lines can be clearly seen. The Fourier transform of the image is shown in Figure \ref{fig:exp1bFourier}(c). The band-pass filters to be applied on the sidebands and the low-pass filter to be applied on the baseband are also marked on the Fourier transform. By applying the band-pass filter to the horizontal sideband, the 30 millisecond sub-exposure image is extracted. By applying the band-pass filter to the vertical sideband, the 15 millisecond sub-exposure image is extracted. The baseband gives the sum of the sub-exposure images, as if there were no optical modulation.

In Figure \ref{fig:exp1bFourier}(d), the baseband image is shown. The motion blur prevents seeing the details in the image, which has a full-exposure period of 45 milliseconds. The 30 millisecond sub-exposure image is shown in Figure \ref{fig:exp1bFourier}(e); and the 15 millisecond sub-exposure image is shown in Figure \ref{fig:exp1bFourier}(f). The readability of the text and the visibility of details are much improved in the 15 millisecond sub-exposure image.

\begin{figure*}
\centering\includegraphics[width=0.48\textwidth]{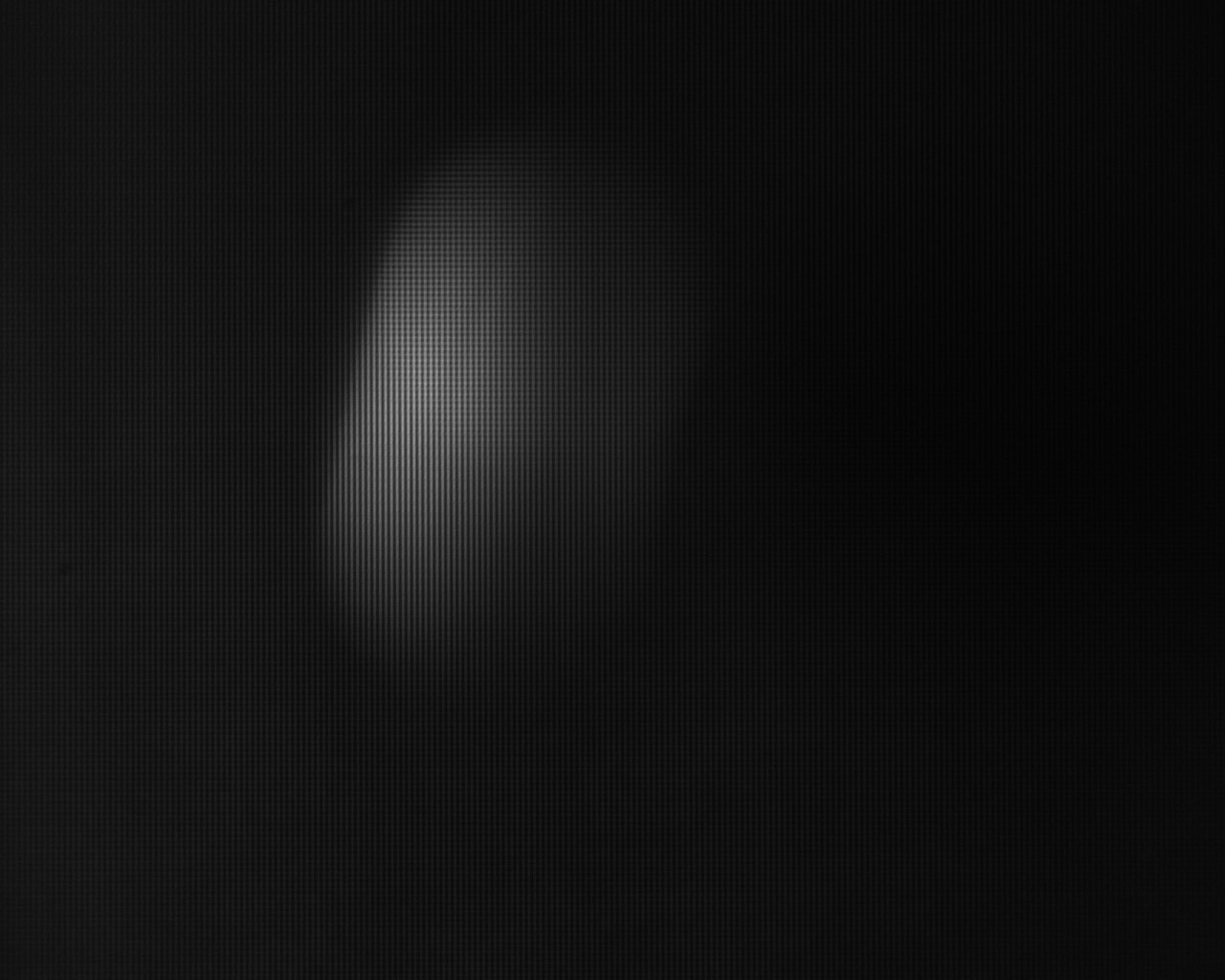}
\centering\includegraphics[width=0.48\textwidth]{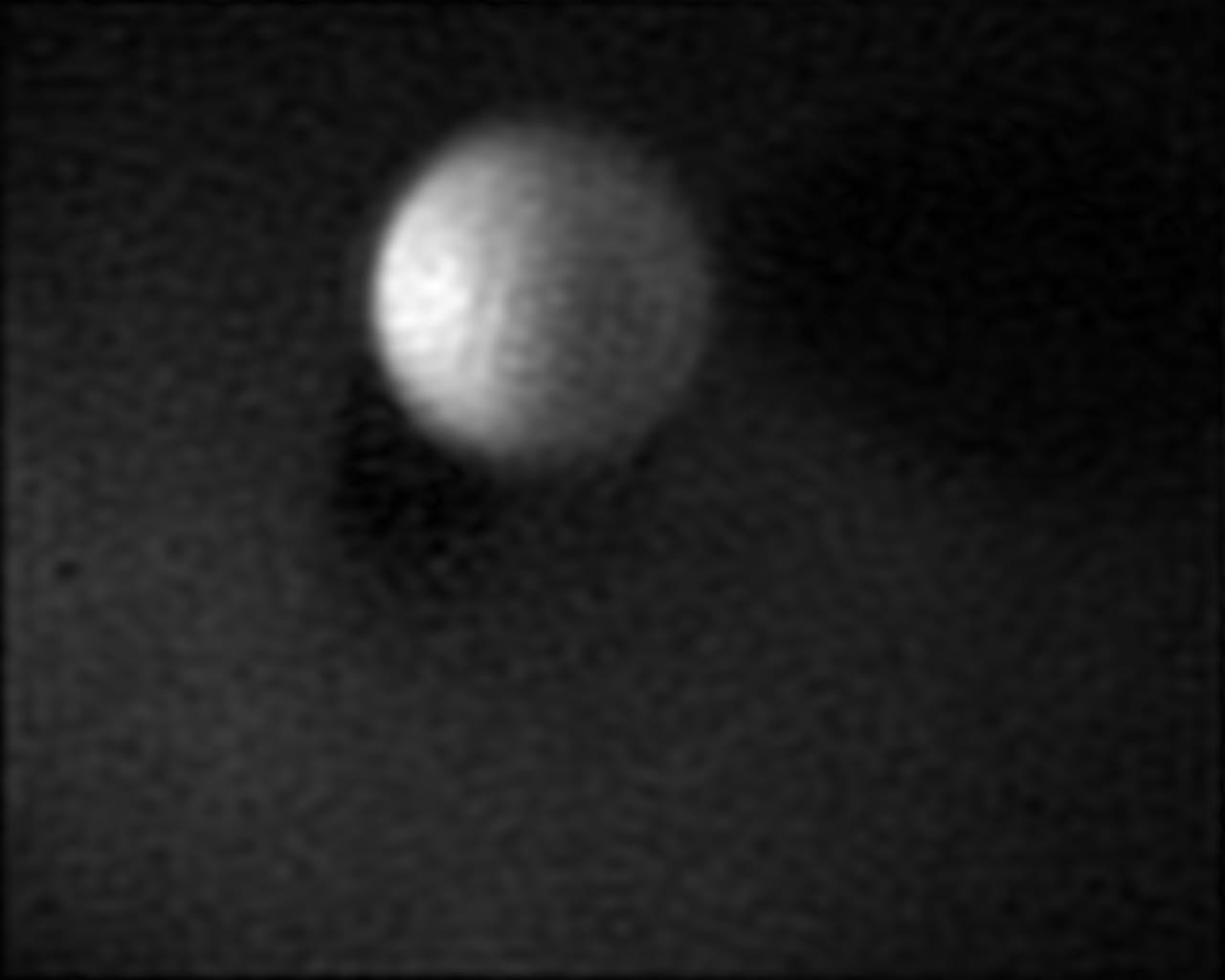}\\
(a) \hspace{6cm} (b) \\
\vspace{0.08cm}
\caption{Sub-exposure image extraction with FDMI. (a) Recorded exposure-coded image, (b) Extracted sub-exposure image.}
\label{fig: exp2bFourier}
\end{figure*}

In the fourth experiment, a ball is thrown in front of the camera. The recorded image and a 15 millisecond sub-exposure image are shown in Figure \ref{fig: exp2bFourier}. The sub-exposure image is able to capture the ball shape with much less motion blur.

\begin{figure*}
\centering
\begin{subfigure}[]{1\textwidth}
\centering
\includegraphics[width=0.45\textwidth]{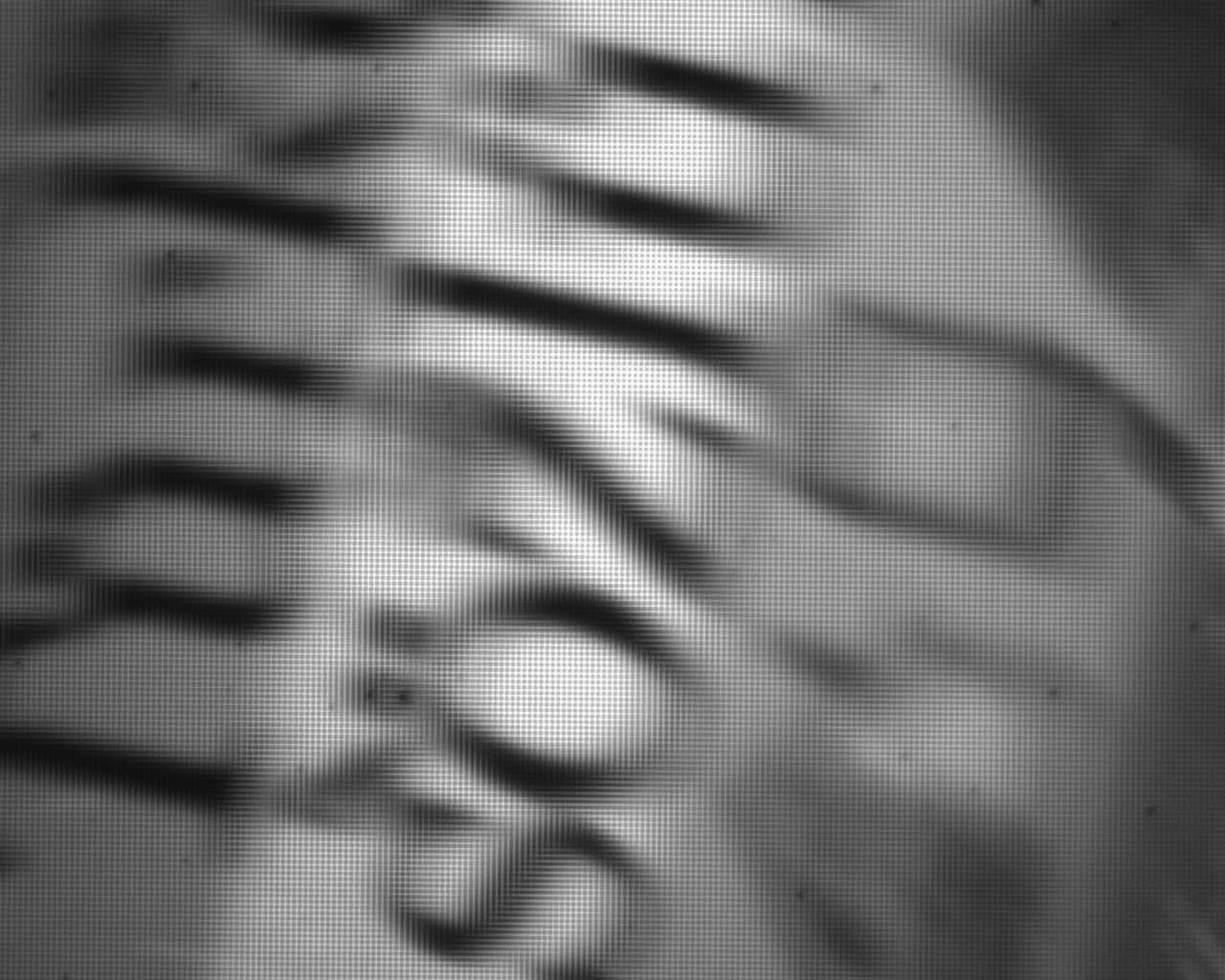}
\includegraphics[trim= 500 340 470 390, clip=true, width=0.45\textwidth]{Exp3O}
\caption{}
\end{subfigure}

\vspace{0.2cm}

\begin{subfigure}[]{1\textwidth}
\centering
\includegraphics[width=0.45\textwidth]{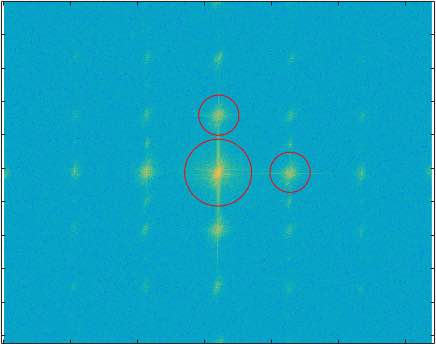}
\includegraphics[width=0.45\textwidth]{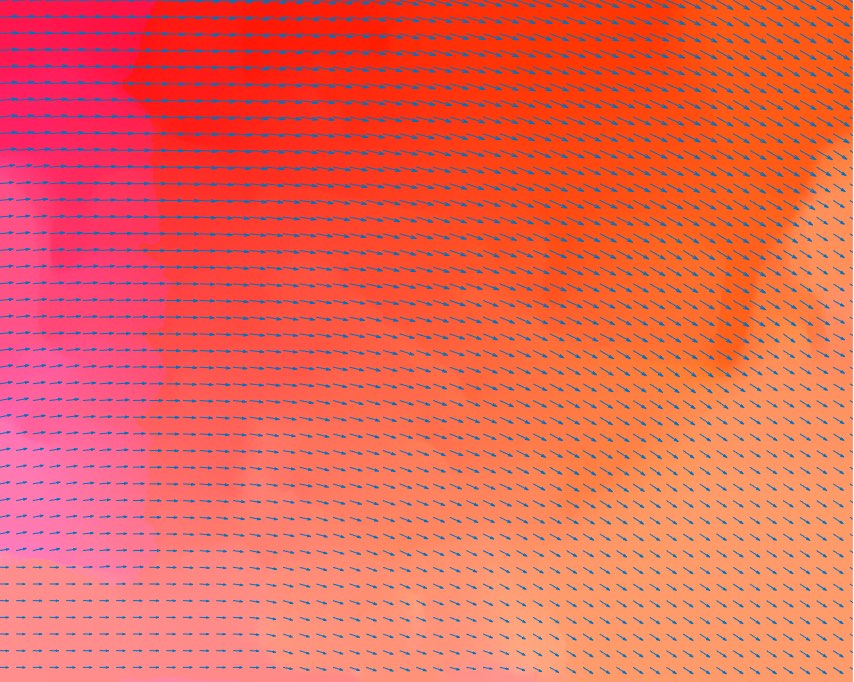}
\caption{}
\end{subfigure}

\vspace{0.2cm}

\begin{subfigure}[]{1\textwidth}
\centering
\includegraphics[width=0.18\textwidth]{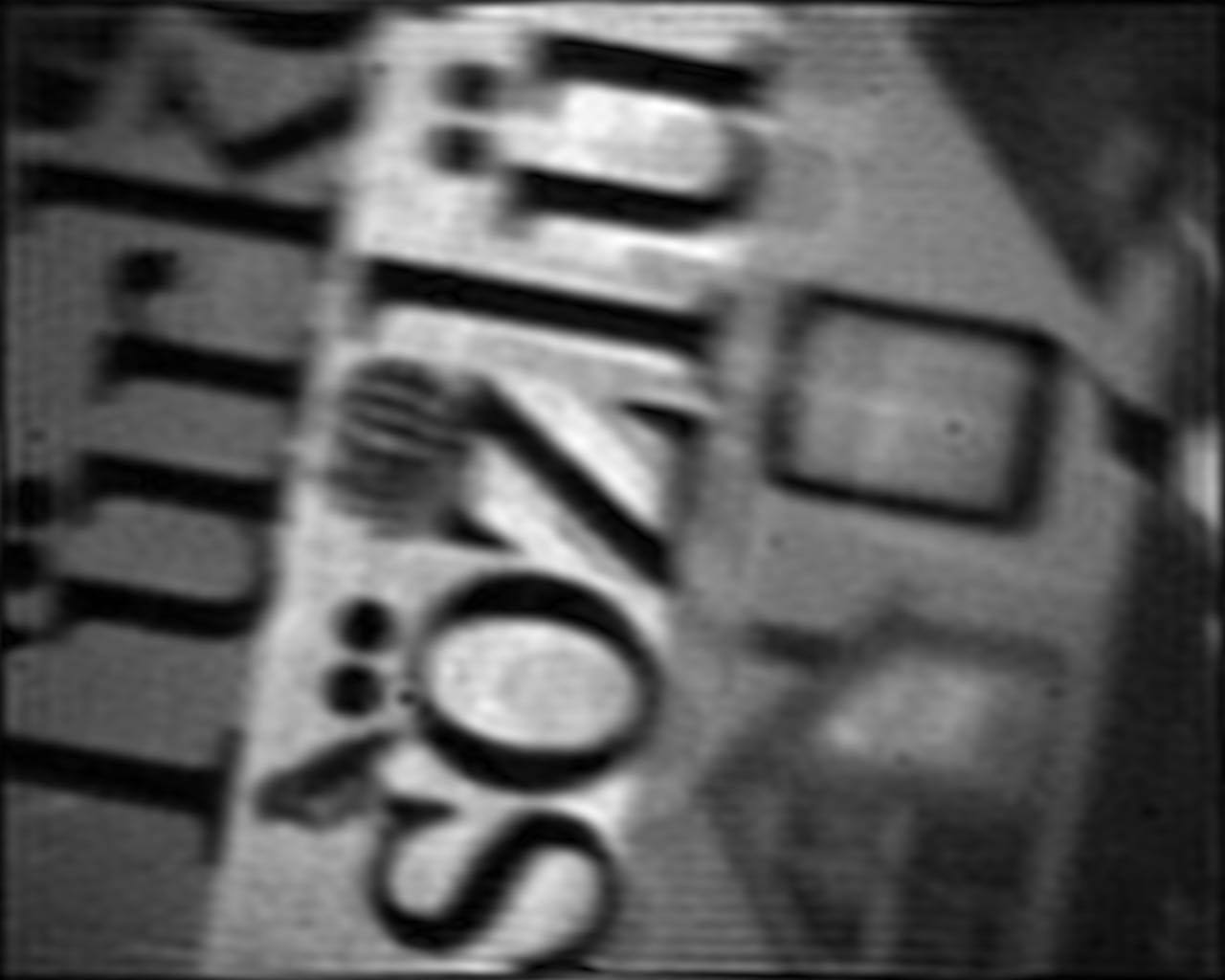}
\includegraphics[width=0.18\textwidth]{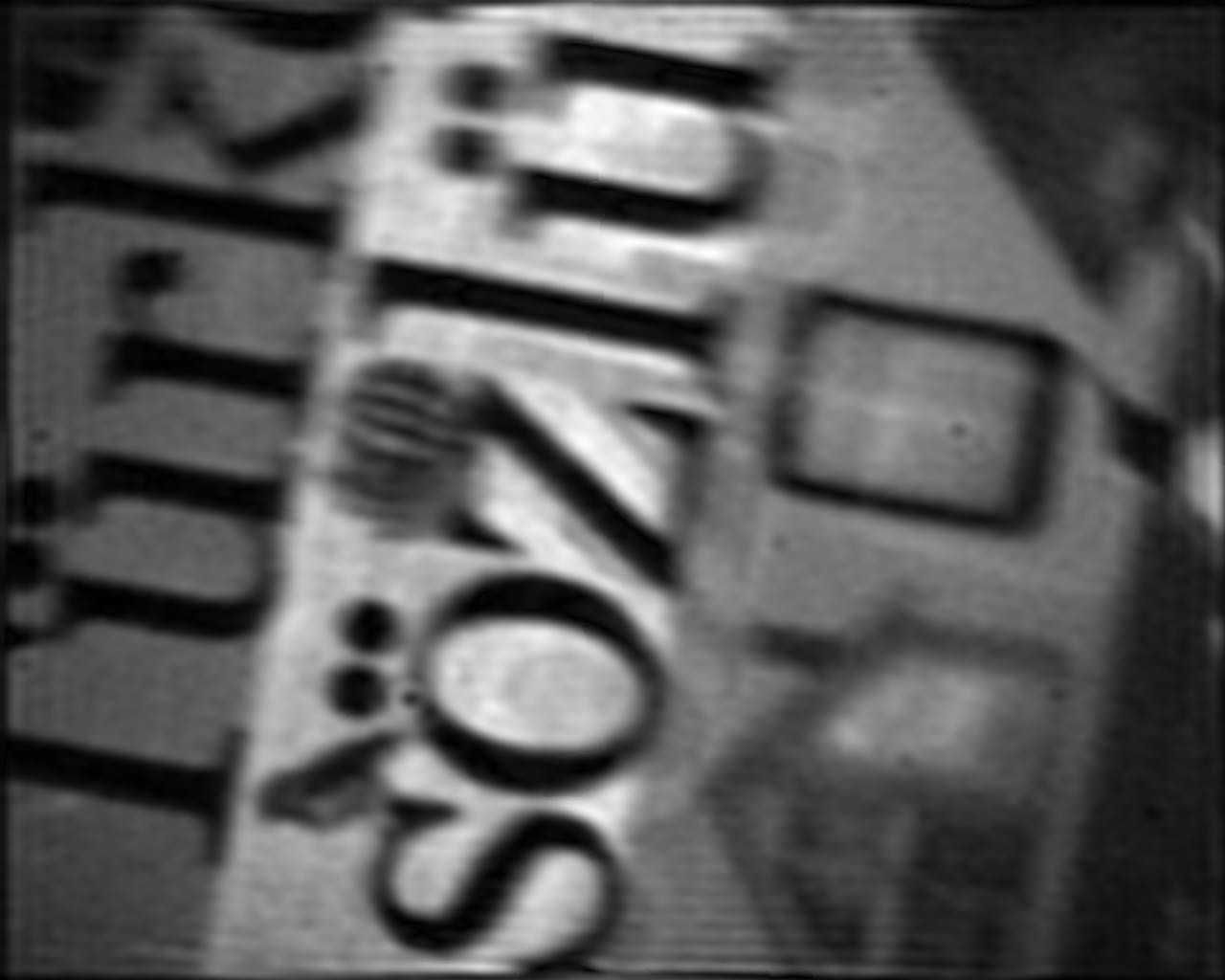}
\includegraphics[width=0.18\textwidth]{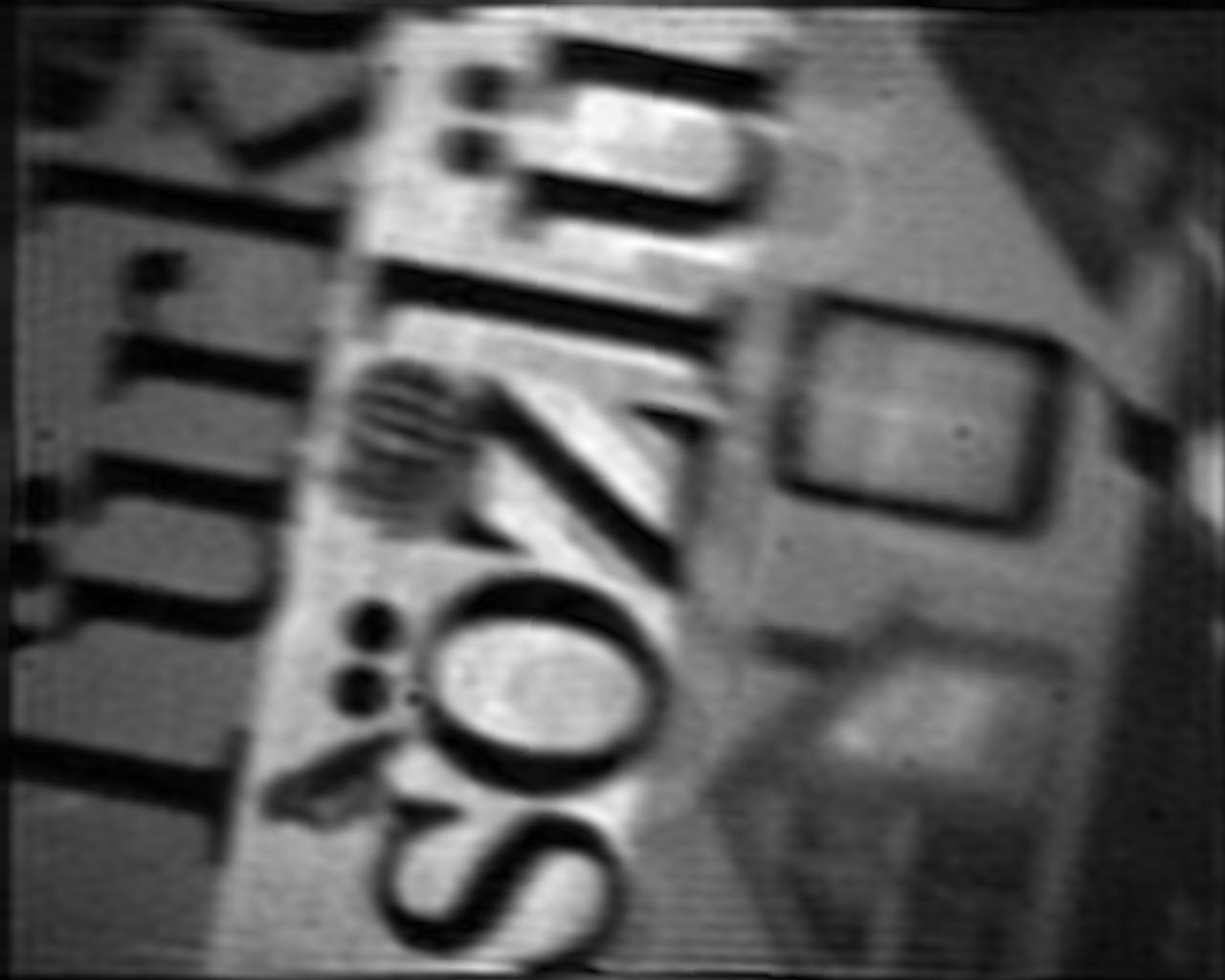}
\includegraphics[width=0.18\textwidth]{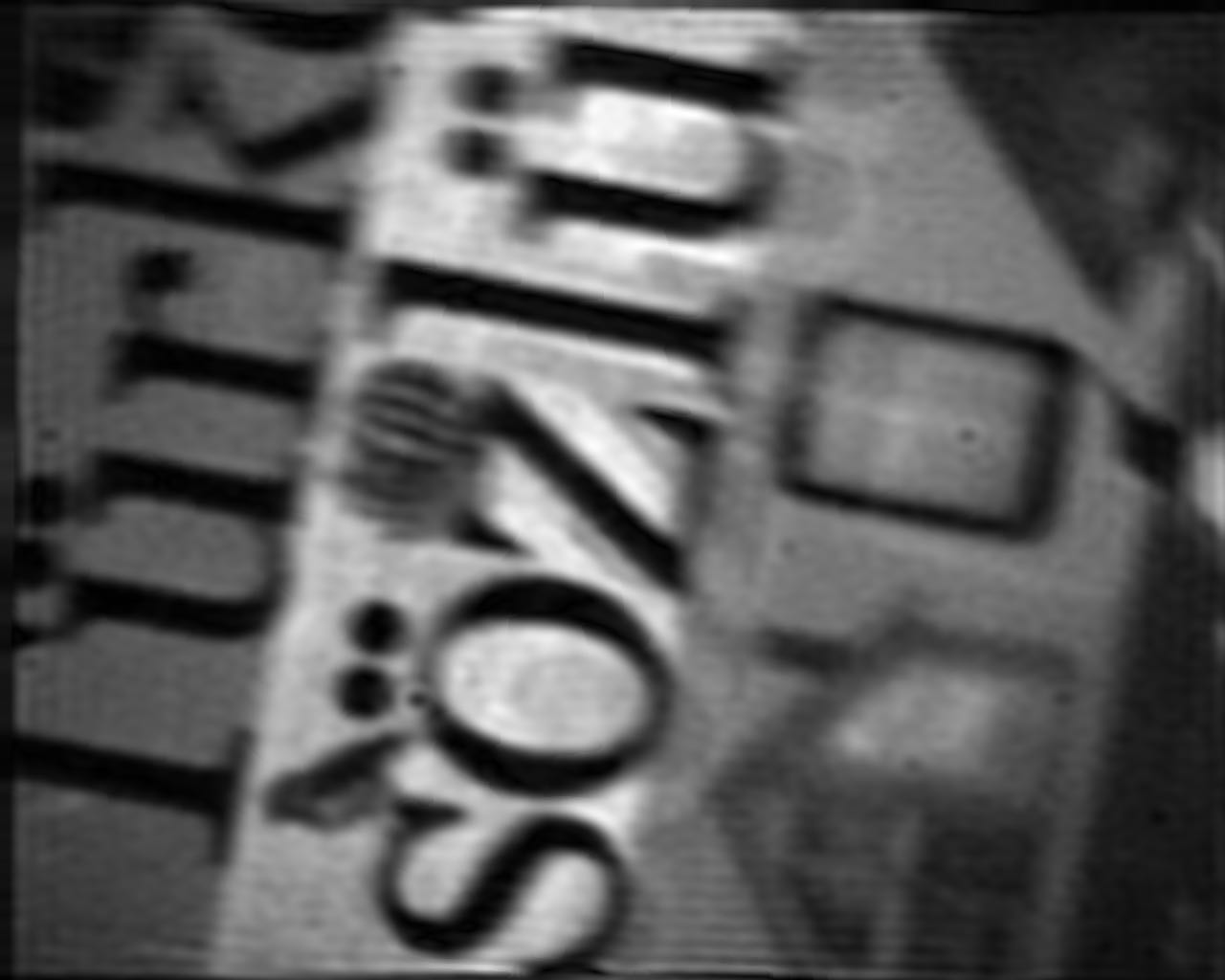}\\
\includegraphics[width=0.18\textwidth]{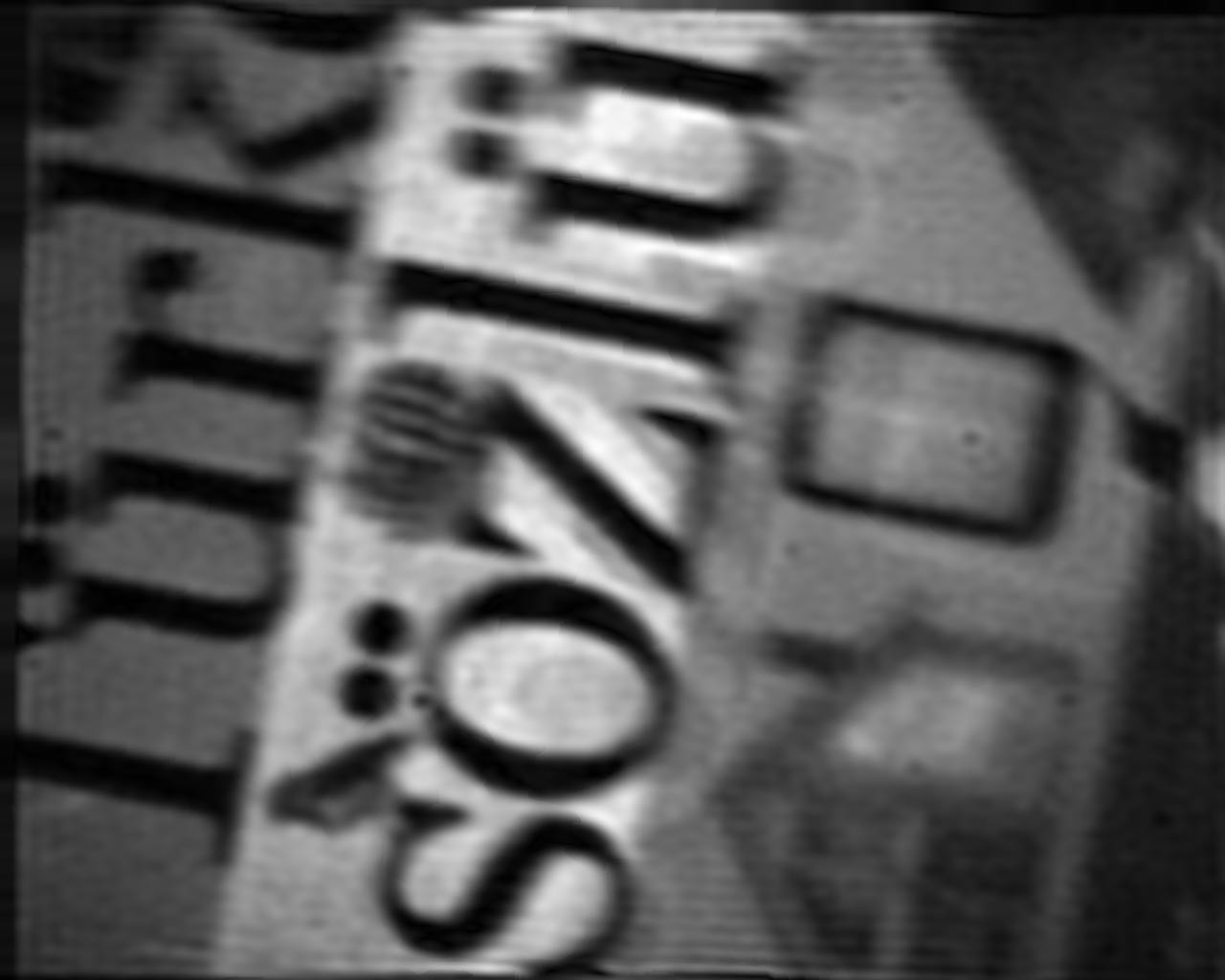}
\includegraphics[width=0.18\textwidth]{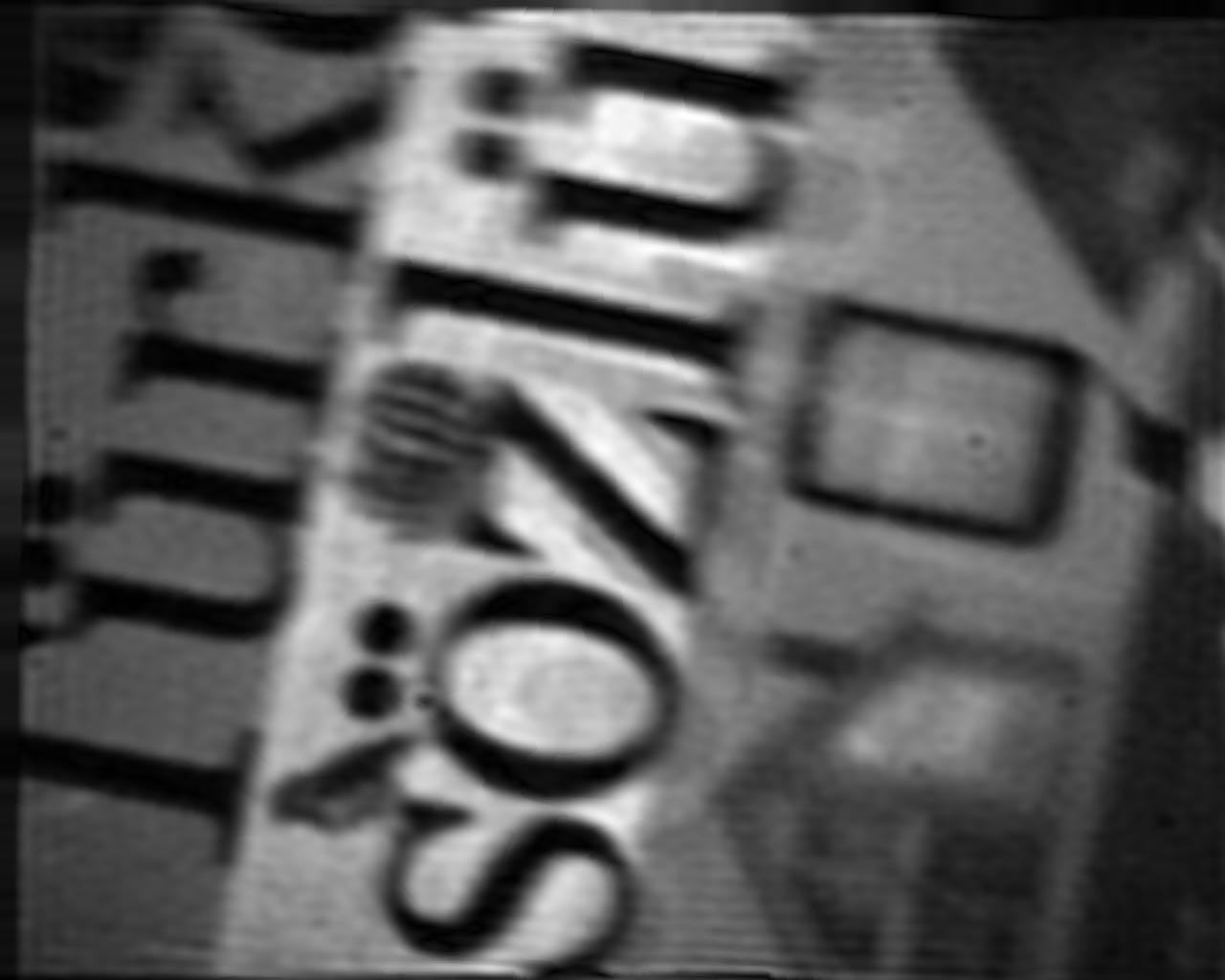}
\includegraphics[width=0.18\textwidth]{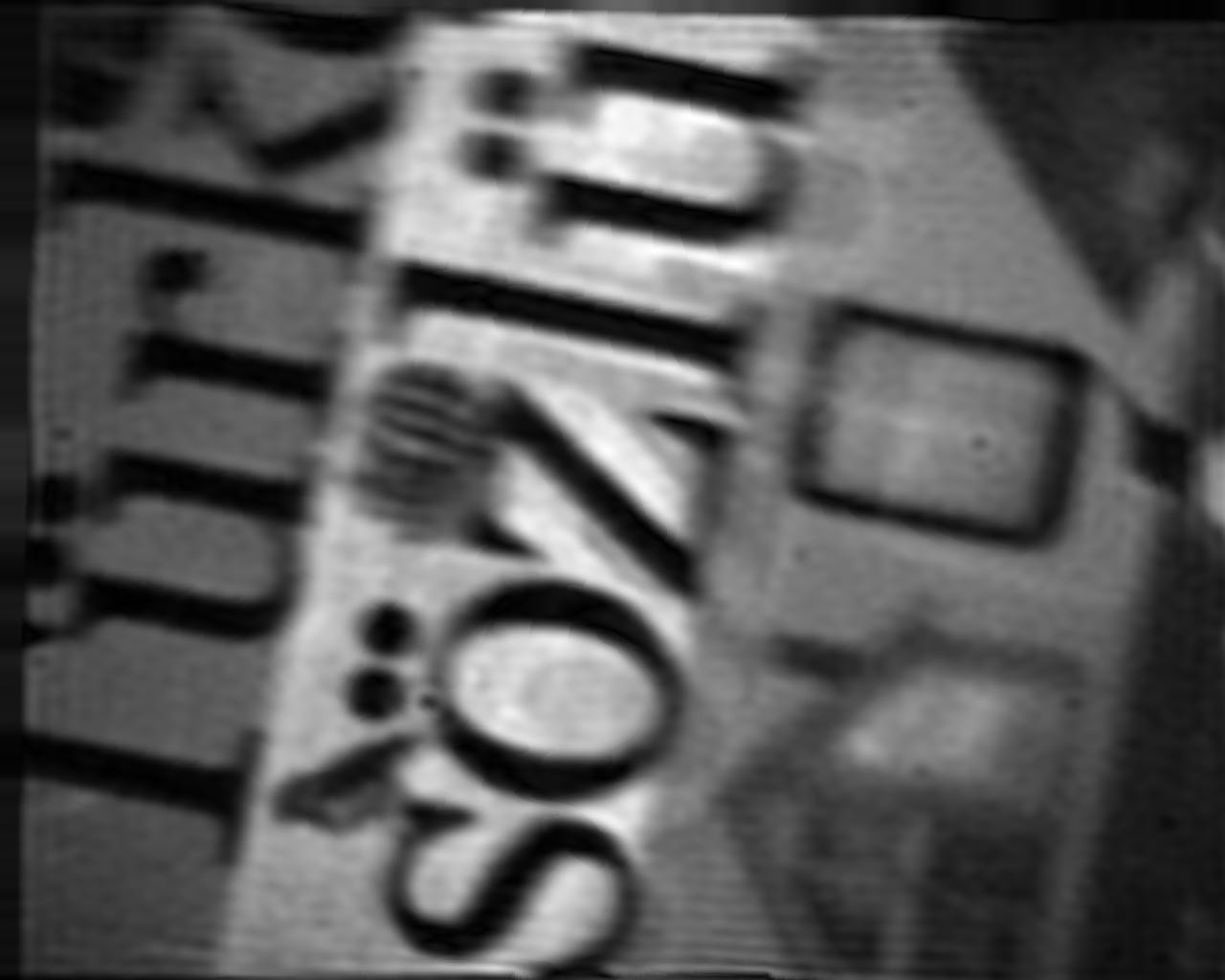}
\includegraphics[width=0.18\textwidth]{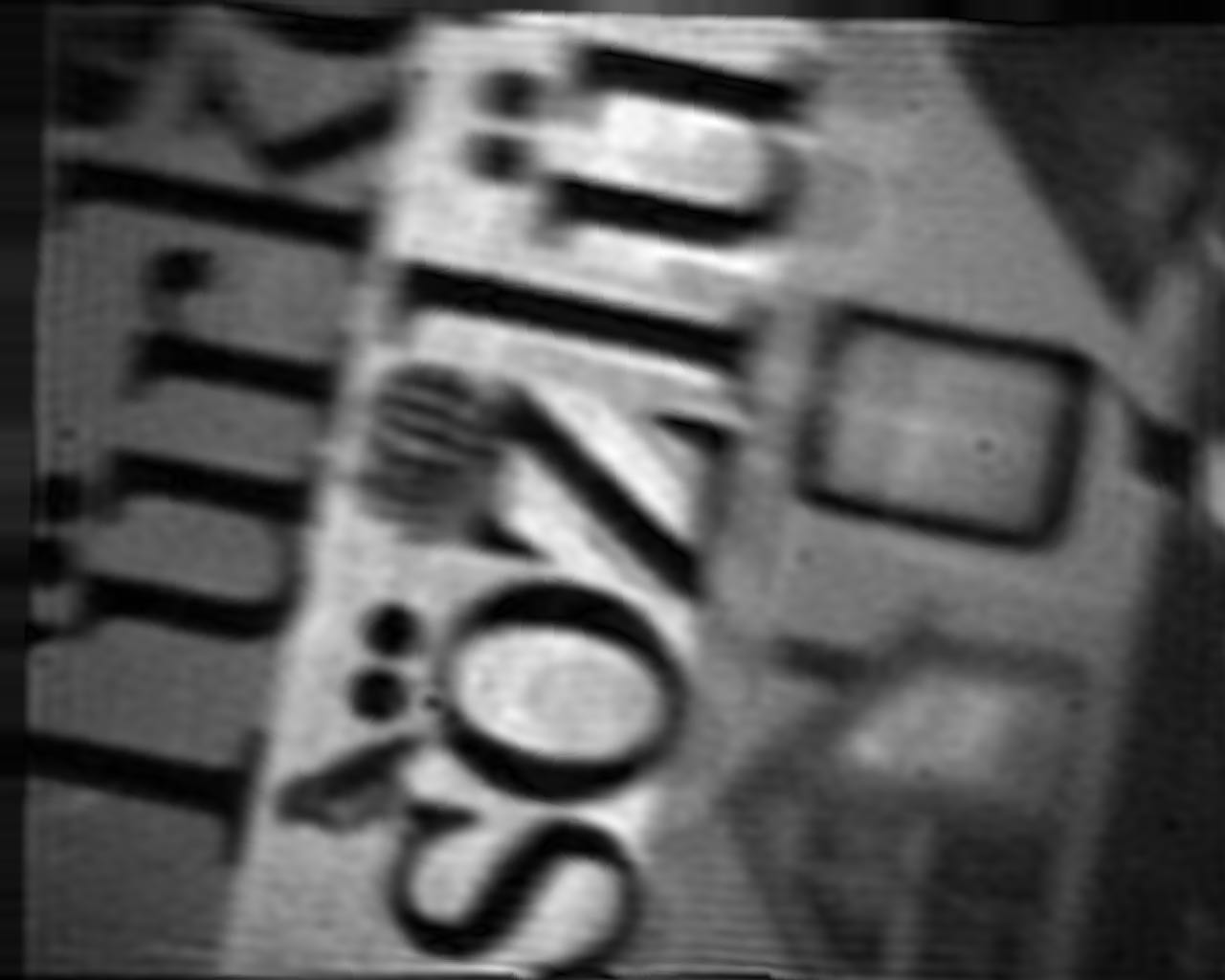}\\ 
\includegraphics[width=0.18\textwidth]{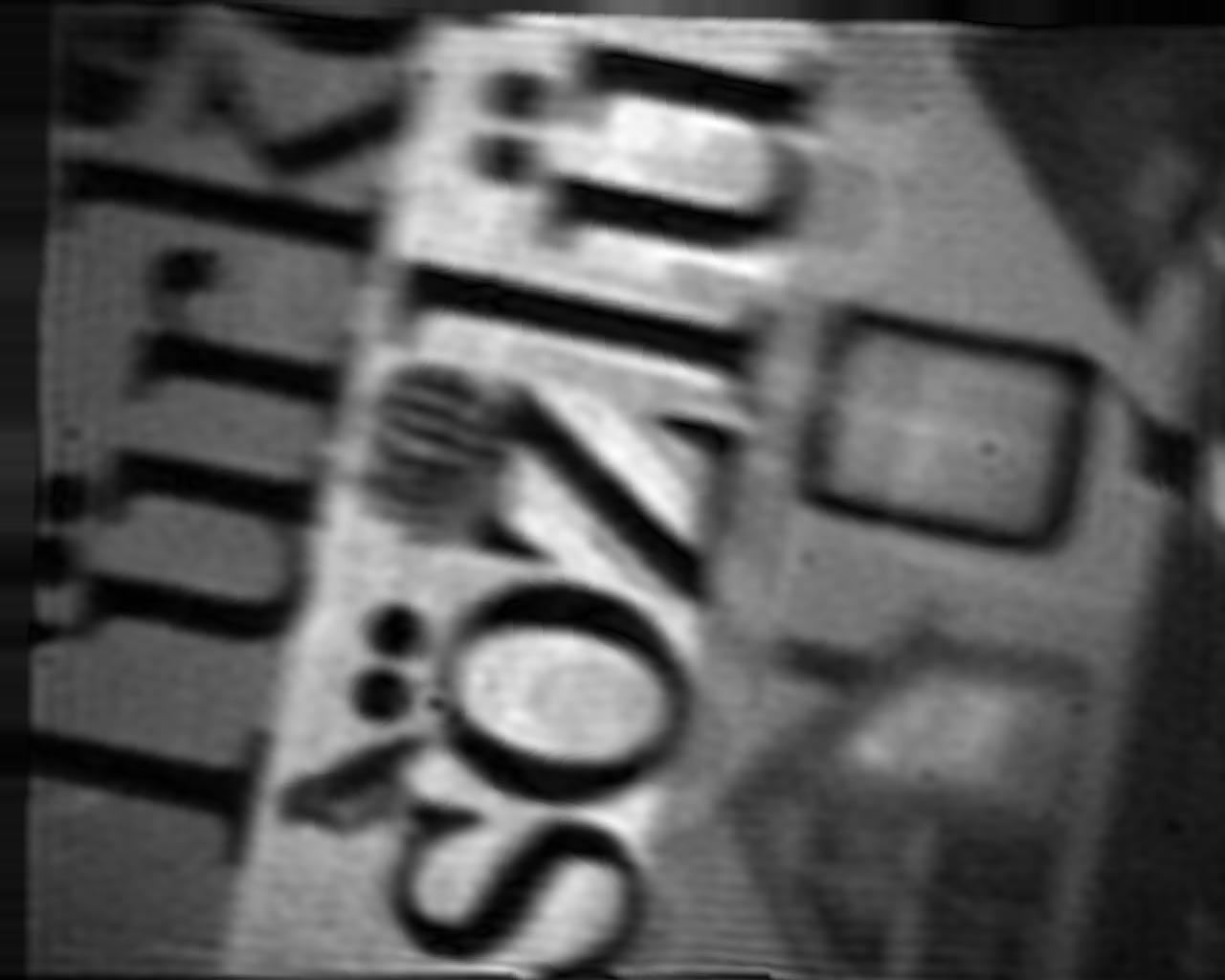}
\includegraphics[width=0.18\textwidth]{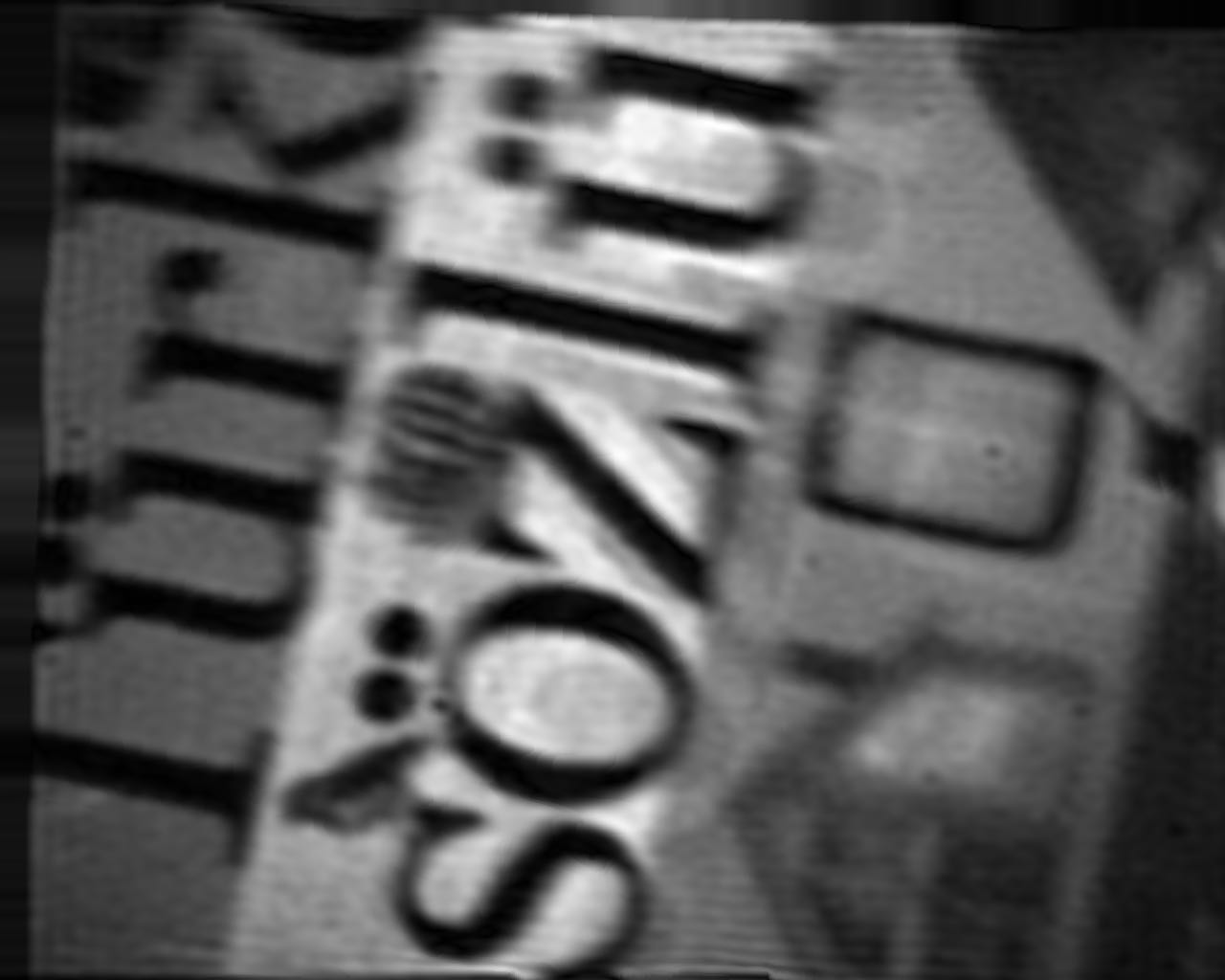}
\includegraphics[width=0.18\textwidth]{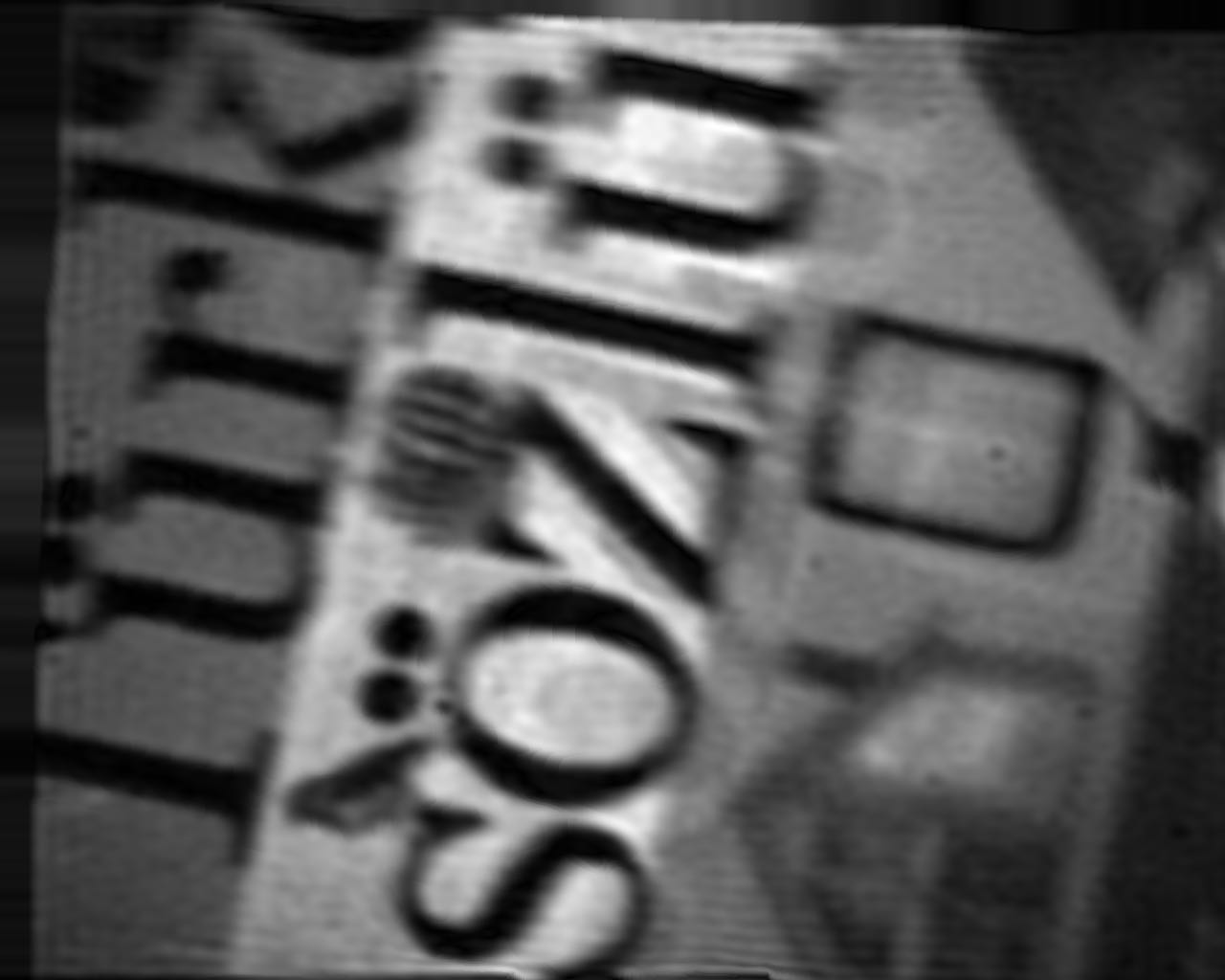}
\includegraphics[width=0.18\textwidth]{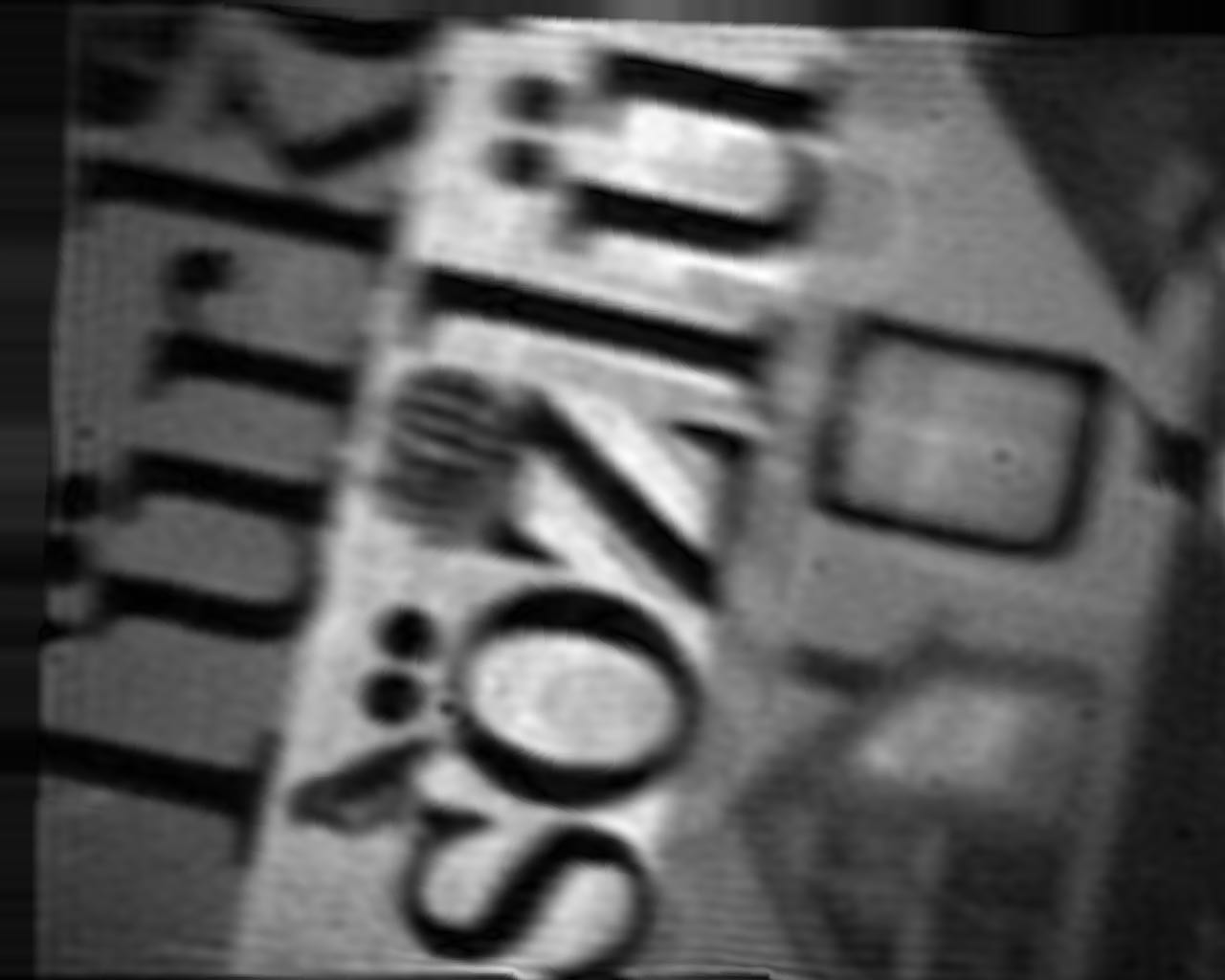}\\
\includegraphics[width=0.18\textwidth]{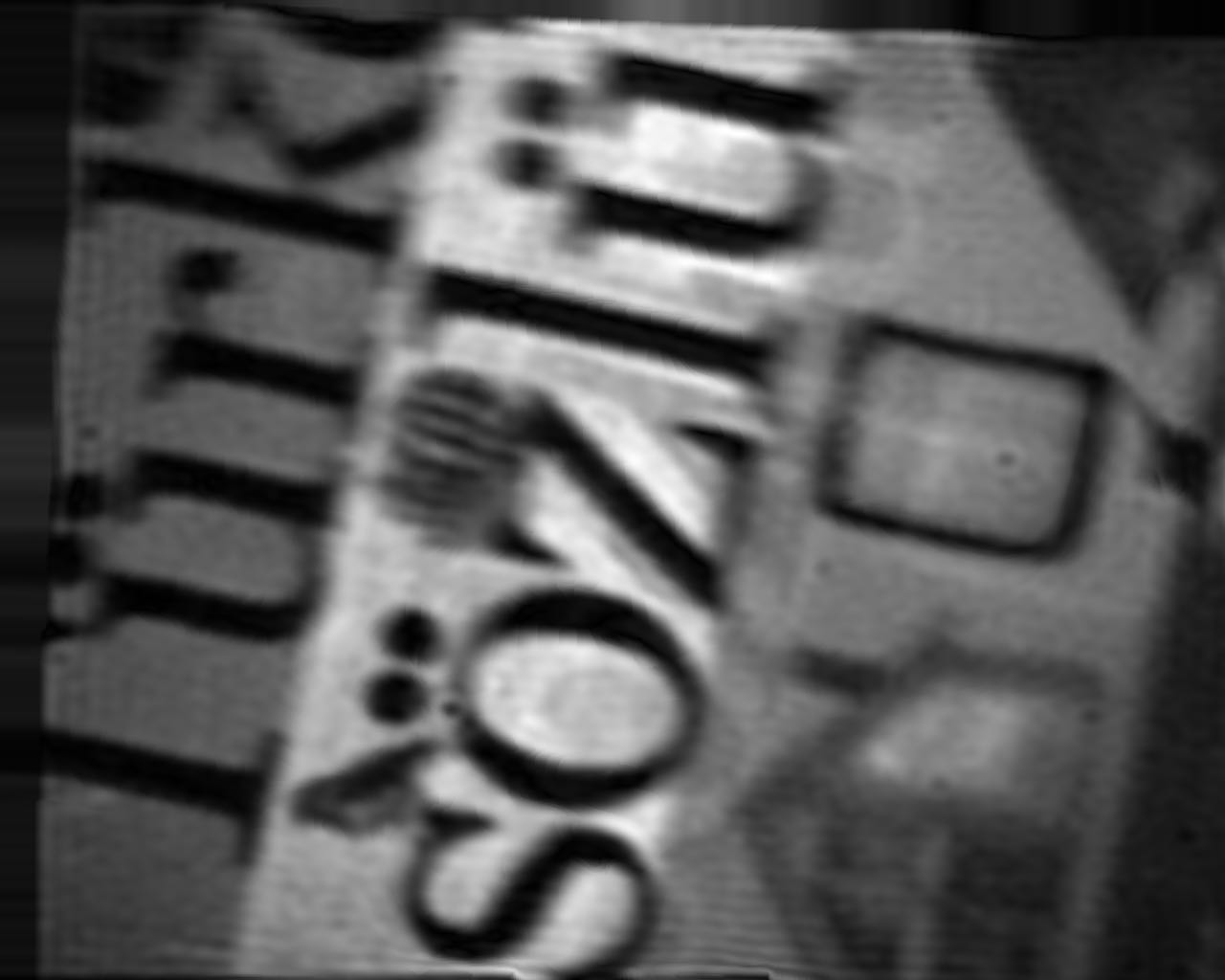}
\includegraphics[width=0.18\textwidth]{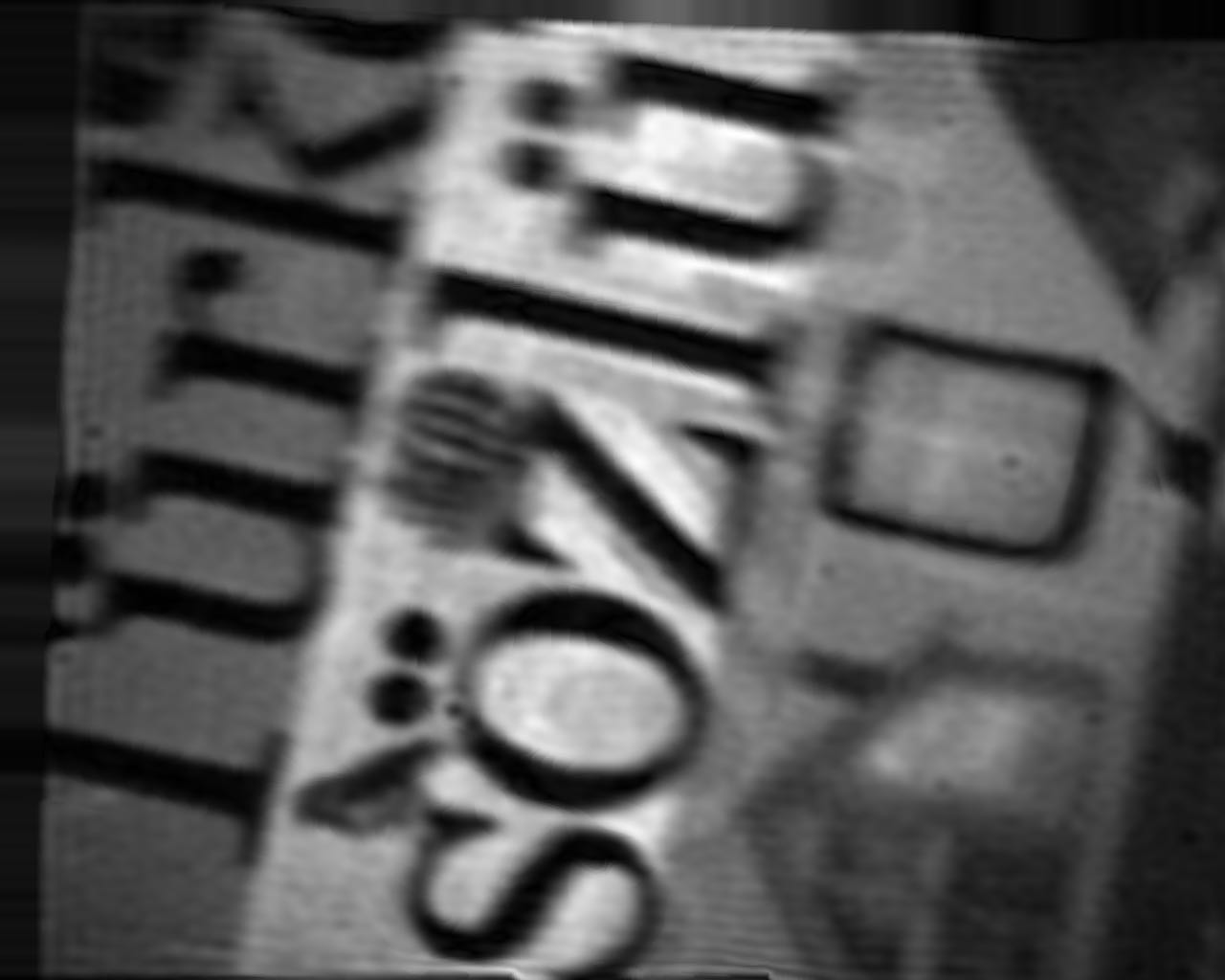}
\includegraphics[width=0.18\textwidth]{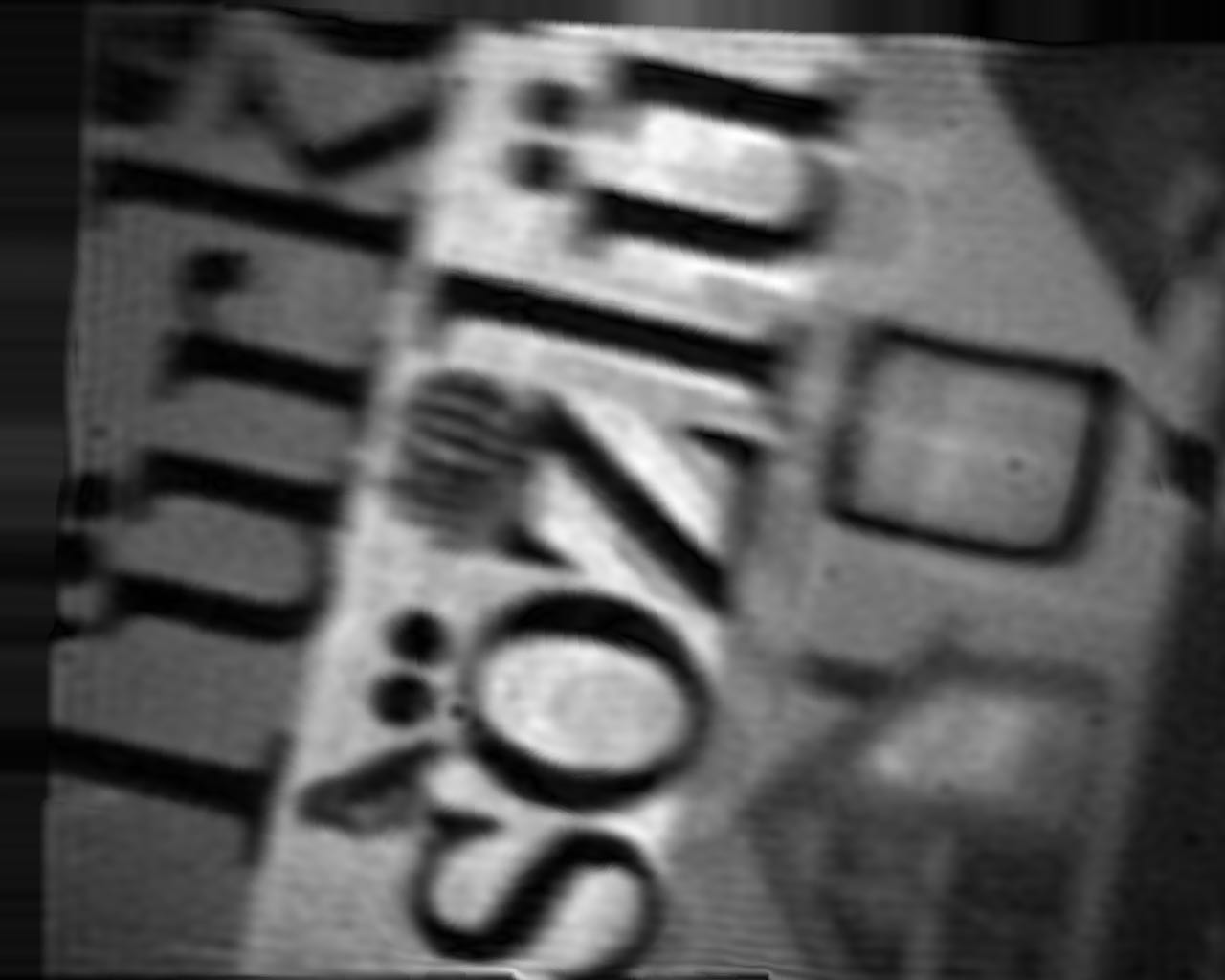}
\includegraphics[width=0.18\textwidth]{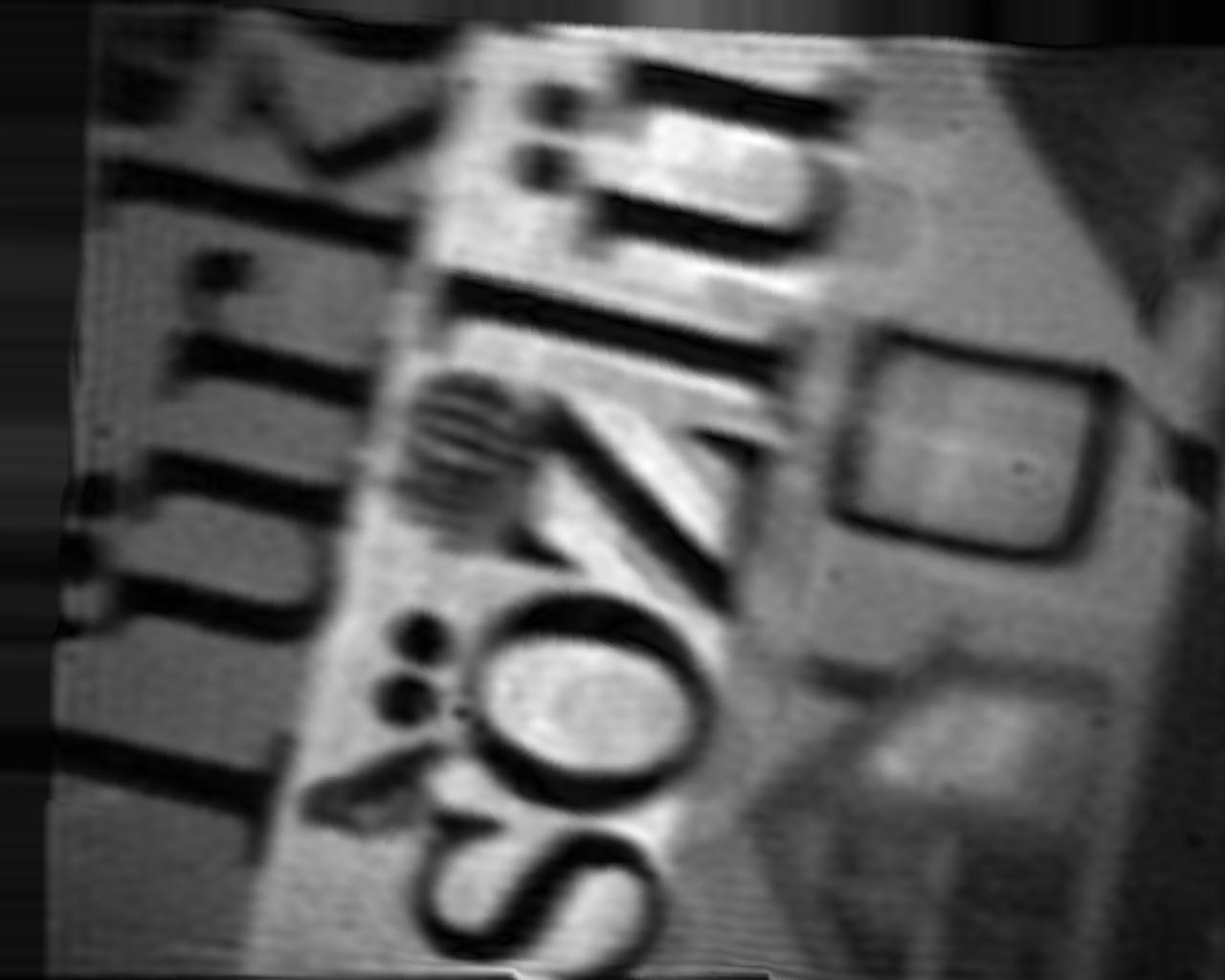}\\ 
\caption{}
\end{subfigure}

\caption{Extracting intermediate sub-exposure frames through optical flow estimation. (a) Recorded exposure-coded image, and a zoomed-in region. (b) Fourier transform (magnitude) of the image, and the estimated motion field between the sub-exposure images. (c) Sixteen sub-exposure frames generated using the estimated motion field.}
\label{fig:Exp4}
\end{figure*}

Finally, we would like to demonstrate that more than two frames can be extracted through motion estimation. The results are shown in Figure \ref{fig:Exp4}. We applied horizontal grating in the first 15 milliseconds, followed by a 15 milliseconds of no-reflection period, and finally vertical grating in the last 15 milliseconds. Two sub-exposure images corresponding to the first part and the last part of the exposure period are extracted, which are then used to estimate the motion field using the optical flow estimation algorithm given in \cite{Liu}. The estimated flow vectors are divided to estimate intermediate frames through image warping, resulting in 16 frames.

\section{Conclusions}

In this paper, we demonstrate a sub-exposure image extraction method, which is based on allocating different regions in Fourier domain to different sub-exposure images. While this theoretical idea is the main goal of the paper, we demonstrate the feasibility of the idea with real life experiments using a prototype optical system. The optical system has physical limitations due to low spatial resolution of the SLM and low light efficiency, which is about $25\%$. Better results would be achieved in the future with the development of high-resolution sensors allowing pixel-wise exposure coding. 

An advantage of the FDMI approach is the low computational complexity, which involves Fourier transform and band-pass filtering. The sub-exposure images from the sidebands and the full exposed image from the baseband are easily extracted by taking Fourier transform, band-pass filtering, and inverse Fourier transform.

Extraction of sub-exposure images can be used for different purposes. In addition to high-speed imaging, one may try to estimate space-varying motion blur, which could be used in scene understanding, segmentation, and space-varying deblurring applications.

\newpage

%\bibliographystyle{elsarticle-num}
%\bibliography{References}

\end{document}